\documentclass[10pt,twocolumn,letterpaper]{article}

\usepackage{cvpr}              

\usepackage[dvipsnames]{xcolor}

\definecolor{cvprblue}{rgb}{0.21,0.49,0.74}
\usepackage[pagebackref,breaklinks,colorlinks,citecolor=cvprblue]{hyperref}

\usepackage{multirow}
\usepackage{placeins}

\def\paperID{302} 
\def\confName{3DV\xspace}
\def\confYear{2025\xspace}

\title{SphereFusion: Efficient Panorama Depth Estimation via Gated Fusion}

\author{Qingsong Yan\textsuperscript{1}, 
    Qiang Wang \textsuperscript{2}\thanks{Corresponding author}, 
    Kaiyong Zhao \textsuperscript{3},
    Jie Chen \textsuperscript{4}, \\
    Bo Li \textsuperscript{5}, 
    Xiaoweo Chu \textsuperscript{6}, 
    Fei Deng \textsuperscript{1}\\
    \textsuperscript{1}Wuhan University, Wuhan, China \ 
    \textsuperscript{2}HIT (Shenzhen), Shenzhen, China \ 
    \textsuperscript{3}XGRIDS, Shenzhen, China \\
    \textsuperscript{4}HKBU, Hong Kong, China \ 
    \textsuperscript{5}HKUST, Hong Kong, China \ 
    \textsuperscript{6}HKUST (Guangzhou), Guangzhou, China\\
    {\tt\small  yanqs\_whu@whu.edu.cn, qiang.wang@hit.edu.cn, kyzhao@xgrids.com, chenjie@comp.hkbu.edu.cn } \\
    {\tt\small bli@cse.ust.hk, xwchu@ust.hk, fdeng@sgg.whu.edu.cn }
}

\begin{document}
\maketitle
\begin{abstract}

Due to the rapid development of panorama cameras, the task of estimating panorama depth has attracted significant attention from the computer vision community, especially in applications such as robot sensing and autonomous driving. 
However, existing methods relying on different projection formats often encounter challenges, either struggling with distortion and discontinuity in the case of equirectangular, cubemap, and tangent projections, or experiencing a loss of texture details with the spherical projection. 
To tackle these concerns, we present SphereFusion, an end-to-end framework that combines the strengths of various projection methods. 
Specifically, SphereFusion initially employs 2D image convolution and mesh operations to extract two distinct types of features from the panorama image in both equirectangular and spherical projection domains. These features are then projected onto the spherical domain, where a gate fusion module selects the most reliable features for fusion. Finally, SphereFusion estimates panorama depth within the spherical domain.
Meanwhile, SphereFusion employs a cache strategy to improve the efficiency of mesh operation.
Extensive experiments on three public panorama datasets demonstrate that SphereFusion achieves competitive results with other state-of-the-art methods, while presenting the fastest inference speed at only 17 ms on a 512$\times$1024 panorama image.

\vspace{-1em}

\end{abstract}    
\section{Introduction}
\label{sec:intro}

Depth estimation is an important task in computer vision that helps to understand the 3D environment. In particular, the panorama image has a 360$^\circ$ field of view (FOV) and can reconstruct the entire surrounding environment in one shot \cite{zioulis2018omnidepth,wang2020bifuse}. With the development of consumer-level panorama cameras, such as Ricoh Theta, Samsung Gear360, and Insta360 ONE, it becomes an intriguing task to estimate the depth map from the panorama image \cite{pintore2021slicenet,sun2021hohonet,yan2022spheredepth,shen2022panoformer,li2022omnifusion}, and many related datasets have been generated to facilitate the research of panorama depth estimation \cite{zioulis2018omnidepth,armeni2017joint,chang2017matterport3d,zheng2020structured3d,albanis2021pano3d}.

\begin{figure}
	\begin{center}
		\includegraphics[width=0.7\linewidth]{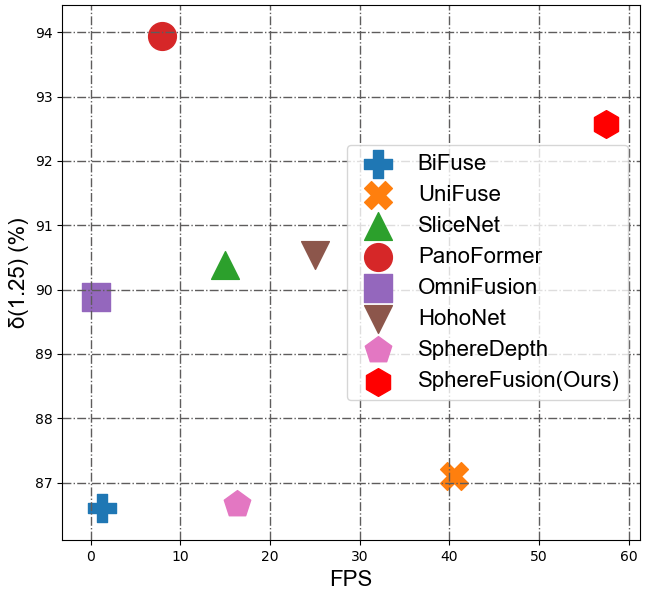}
	\end{center}

    \vspace{-1.5em}
 
	\caption{
        Comparison with BiFuse \cite{wang2020bifuse}, UniFuse \cite{jiang2021unifuse}, SliceNet \cite{pintore2021slicenet}, PanoFormer \cite{shen2022panoformer}, OminiFusion \cite{li2022omnifusion}, SphereDepth \cite{yan2022spheredepth}, HohoNet \cite{sun2021hohonet} on Stanford2D3D \cite{armeni2017joint} with resolution of $512 \times 1024$. The horizontal axis is the FPS, and the vertical axis is $ \delta(1.25) (\%)$, which counts the percentage of the absolute relative difference between the prediction and the ground truth that is less than 1.25. The higher FPS and higher $ \delta(1.25) (\%)$ mean better.
	}
	\label{fig:time_viz}

    \vspace{-2.0em}
 
\end{figure}

However, it is challenging to find a proper way to represent the panorama image. 
The most popular equirectangular projection faces huge distortion around the poles and poor discontinuity near the borders. The cubemap projection \cite{skupin2017standardization} and the tangent projection \cite{eder2020tangent,shen2022panoformer,li2022omnifusion} project the panorama image to several planes to avoid distortion but introduce serious discontinuity problems and have to rely on a well-designed fusion strategy to merge them. The spherical projection \cite{yan2022spheredepth} can deal with distortion and discontinuity by approximating the sphere, but it is hard to capture details of the panorama image and cannot directly handle high-resolution panorama images.

Based on the characteristics of these projections, different strategies are used to extract features from panorama images. 
The equirectangular projection represents the panorama image in one image plane and can directly utilize 2D image convolution to extract features. Nevertheless, the equirectangular projection faces huge distortion and needs specially designed 2D convolution kernels to extract reliable features \cite{zioulis2018omnidepth,tateno2018distortion,chen2021distortion}.
Like the equirectangular projection, the cubemap projection and tangent projection employ multiple planes to represent the panorama image and use 2D image convolution to extract features. However, they require a suitable mechanism to fuse features from different planes and maintain global consistency \cite{cheng2018cube,li2022omnifusion,shen2022panoformer,peng2022high}. 
\cite{yan2022spheredepth,hu2021subdivision,feng2019meshnet,hanocka2019meshcnn}.
On the other hand, the spherical projection is an ideal way to represent the panorama image, and utilizes mesh operation to extract features. Nonetheless, the mesh operation struggles to capture details and exhibits inefficiency \cite{yan2022spheredepth,hu2021subdivision,feng2019meshnet,hanocka2019meshcnn}.
Meanwhile, BiFuse \cite{wang2020bifuse} and UniFuse \cite{jiang2021unifuse} try to combine the strengths of different projections. Although they achieve better depth estimation results, they still suffer from distortion around the poles.

To this end, this paper proposes SphereFusion, which unites equirectangular and spherical projections.
SphereFusion first represents the panorama image by equirectangular and spherical projections, then extracts two types of features, and finally fuses them through a fusion module to estimate the panorama depth in the spherical projection.
To balance the accuracy and efficiency, SphereFusion implements a lightweight encoder to extract features \cite{he2016deep} and utilizes a cache strategy to reduce the computation complexity of the mesh operation.
Fig. \ref{fig:time_viz} compares the efficiency and quality of the depth map with state-of-the-art methods \cite{wang2020bifuse,jiang2021unifuse,pintore2021slicenet,shen2022panoformer,li2022omnifusion,yan2022spheredepth,sun2021hohonet}.  
SphereFusion obtains high-quality depth maps and achieves around 60 FPS during inference on the panorama image with a resolution of $512 \times 1024$.
Our contributions are summarized as follows.

\begin{enumerate}

\item We propose a panorama depth estimation method, SphereFusion, to estimate the depth map in the spherical projection, which uses the features from the equirectangular projection to improve the details.
	
\item We design a feature fusion module, GateFuse, which selects reliable features from two projections to improve the quality of the depth map.
	
\item Experiments demonstrate that SphereFusion achieves competitive results on three public panorama datasets and can produce point clouds with less noise and higher completeness. Besides, SphereFusion achieves 60 FPS of inference speed on an NVIDIA RTX 3090, outperforming existing methods.

\end{enumerate}
\section{Related Work}
\label{sec:related}


\subsection{Perspective Depth Estimation}
Over the past few decades, the perspective 2D image from the classical pinhole camera model has attracted much interest in inferring depth estimation for 3D sensing applications. 
The traditional methods, such as Make3D \cite{saxena2008make3d} and Pop-up \cite{hoiem2005automatic}, leverage the prior depth distribution and regress the pixel-wise depth via a Markov random field (MRF). 
Due to the success of deep learning, researchers have also utilized its capability of multi-level feature extraction to improve the depth quality. As the first trial, Eigen \cite{eigen2014depth} and Laina \cite{laina2016deeper} design end-to-end convolutional neural networks to regress the depth map from an RGB image. 
Furthermore, several recent studies \cite{bhat2021adabins,yuan2022new} take advantage of the visual transformer (ViT) \cite{vit2021}, which captures the relationship between each pixel patch to regress the depth map with a global context. 
To improve the model generalization and depth consistency, Midas \cite{ranftl2020towards} focuses on the training strategy of mixing different datasets and tuning the loss function. 
Other studies, such as \cite{eigen2015predicting,alhashim2018high,yin2021virtual,patil2022p3depth}, introduce some constraints like the surface normal or the semantic information to co-optimize the depth map.
However, a key challenge in training models for depth estimation is the lack of high-quality depth labels, which motivates the studies of unsupervised learning \cite{zhou2017unsupervised,godard2019digging,bian2021unsupervised,zhao2020towards,watson2021temporal}. SfMLearner \cite{zhou2017unsupervised} leverages photometric consistency to estimate the relative pose and depth simultaneously. MonoDepth2 \cite{godard2019digging} filters out the outliers that violate photometric consistency to deal with occlusions. In addition, SfMLearner-SC \cite{bian2021unsupervised} explores the depth consistency between video frames to restrict the depth of each pixel.
Although depth estimation for 2D perspective images has achieved considerable performance, they cannot be directly applied to the panorama image. 
The main reason is that the panorama image needs a particular type of projection, which usually causes distortion or discontinuity, where network for perspective vision are hard to capture reliable features.

\begin{figure*} [t]
	\begin{center}
		\includegraphics[width=0.8\linewidth]{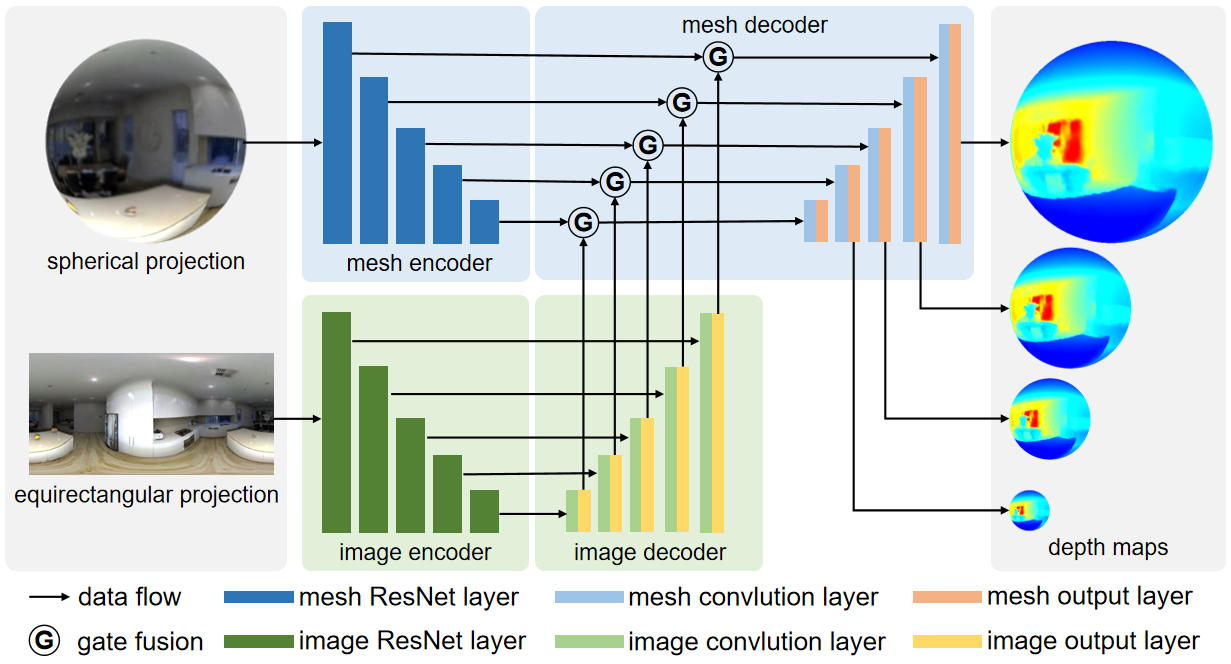}
	\end{center}
    \vspace{-1.5em}
	\caption{
        Given a panorama image in the equirectangular projection and the spherical projection, SphereFusion simultaneously extracts features by a 2D image encoder and a mesh encoder, which follows the ResNet structure \cite{he2016deep}, then fuses these features by the Gate Fusion module in the spherical projection, and finally estimates the depth map through the mesh decoder.
	}
	\label{fig:pipeline}
    \vspace{-1.5em}
\end{figure*}

\subsection{Panorama Depth Estimation}

Unlike the perspective image, which only provides a limited FOV, the panorama image has a 360$^\circ$ FOV. 
Compared to the development of perspective depth estimation, panorama depth estimation is still in its infancy stage. 
Most of the recent studies \cite{zioulis2018omnidepth,wang2020bifuse,sun2021hohonet,shen2022panoformer,junayed2022himode,shen2022neural} rely on special projection methods to transform the panorama image to the 2D perspective image and estimate the depth using the perspective neural network structures.
The equirectangular projection is the most common projection, allowing all the surrounding information to be observed from a single 2D image. 
OmniDepth \cite{zioulis2018omnidepth} proposes the first end-to-end network based on the equirectangular projection for panorama depth estimation along with a large synthetic dataset. 
SliceNet \cite{pintore2021slicenet} uses sliced feature maps to estimate the depth map through LSTM. 
The cubemap projection transforms the panorama image into six perspective images to estimate depth maps and then fuse them through a specially designed module \cite{cheng2018cube,wang2020bifuse}.
The tangent projection approximates the sphere through more perspective images.
OmniFusion \cite{li2022omnifusion}, PanoFormer \cite{shen2022panoformer}, and 360MonoDepth \cite{rey2022360monodepth} samples perspective images from the spherical surface \cite{eder2020tangent} and applies the existing convolutional or ViT models on each tangent patch. 
Despite the success of these methods, the main problem of those projections is introducing distortion and discontinuity in certain areas of the panorama image. 
Although several subsequent studies have attempted to improve the depth quality of these regions, such as designing special convolutional kernels \cite{yu2017flat2sphere, de2018eliminating,su2019kernel,eder2019mapped,tateno2018distortion,wang2020360sd,cheng2020omnidirectional}, applying the deformable convolution \cite{chen2021distortion,fernandez2020corners}, and multi-task learning \cite{eder2019pano,zeng2020joint,jin2020geometric}, we argue that the influence of distortion and discontinuity cannot be removed entirely.
In addition to changing the convolution kernel, BiFuse \cite{wang2020bifuse} and UniFuse \cite{jiang2021unifuse} combine the equirectangular projection and the cubemap projection.

To eliminate the drawbacks of the above two projection methods, some recent studies attempt to process the panorama image in the spherical domain.
SpherePHD \cite{lee2020spherephd} uses the icosahedral spherical mesh to represent the panorama image and extract semantic maps. SphereDepth \cite{yan2022spheredepth} modifies the mesh operation from the SubdivNet \cite{hu2021subdivision}, which is more efficient than MeshCNN \cite{hanocka2019meshcnn} and MeshNet \cite{feng2019meshnet}.
In addition, S2CNN \cite{cohen2018spherical} uses the spheric harmonics function to build networks but is unsuitable for dense estimation. US2CNN \cite{jiang2019spherical} builds grids and manually assigns weight to build the network.

\section{Method}
\label{sec:method}

In this section, we describe the details of SphereFusion, as shown in Fig. \ref{fig:pipeline}.
We first describe how to represent the panorama image in Section \ref{sec:represent} and extract features in the spherical projection in Section \ref{sec:mesh}. 
Then, we show the pipeline of our SphereFusion in Section \ref{sec:pipeline}.
Finally, we present the loss function in Section \ref{sec:loss}.

\begin{figure*} [t]
	\begin{center}

        \subfloat[Panorama Image]{\includegraphics[width=0.22\linewidth]{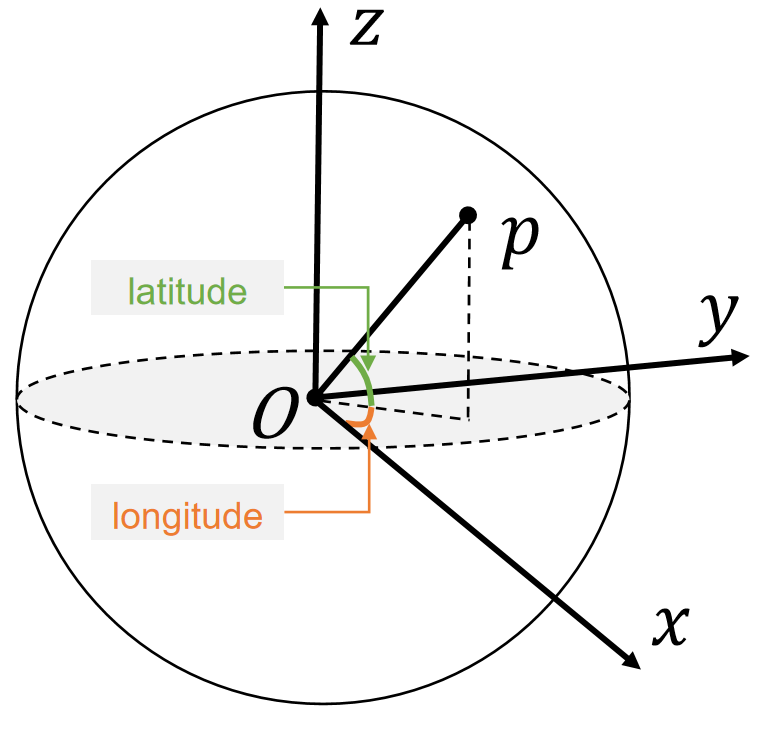}}
		\hspace{0.1em}
		\subfloat[Equirectangular Projection]{\includegraphics[width=0.30\linewidth]{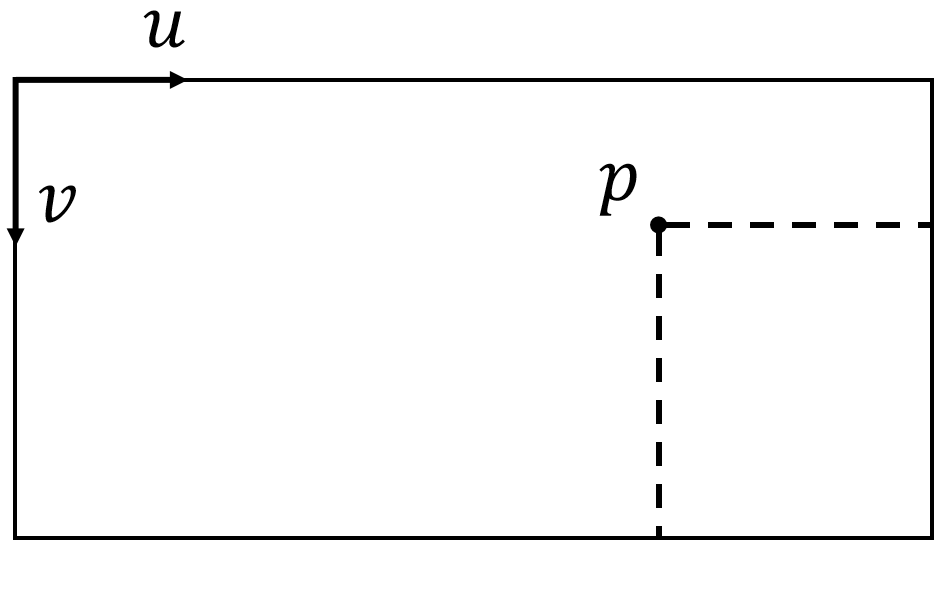}}	
        \hspace{0.1em}
		\subfloat[Spherical Projection]{\includegraphics[width=0.2\linewidth]{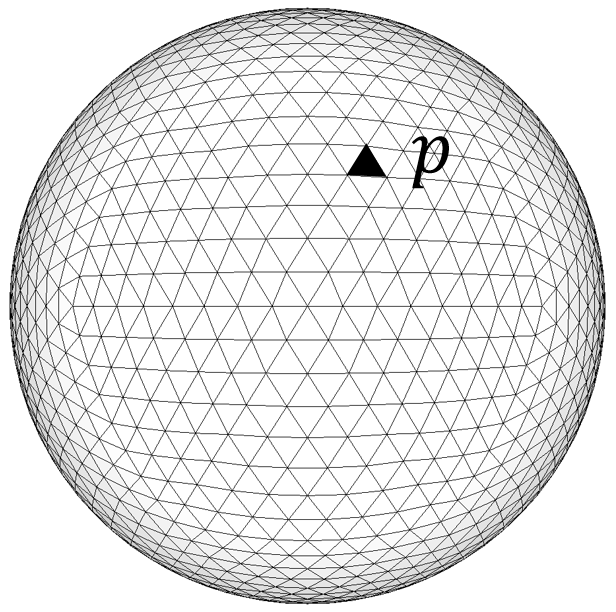}}	
  
	\end{center}
    \vspace{-1.5em}
	\caption{
        The ideal representation of a panorama image is the sphere, but it is impractical. The equirectangular projection is the most popular method, but it suffers from distortion at the poles and discontinuity at the borders. The spherical mesh can approximate the sphere, and their difference becomes smaller with higher MR.
	}
    \vspace{-2.0em}
	\label{fig:projections}
\end{figure*}

\begin{figure} [t]
	\begin{center}
		\subfloat[Mesh Convolution]{\includegraphics[width=0.8\linewidth]{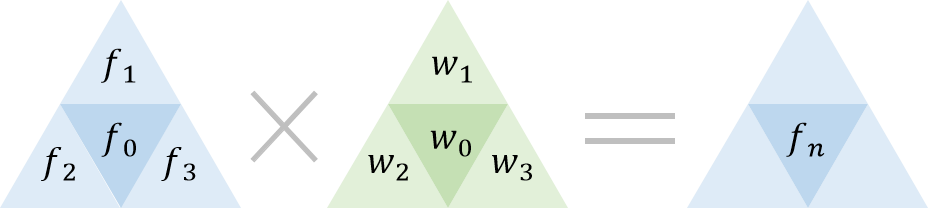}}
		\vspace{0.1em}
		\subfloat[Mesh Pooling]{\includegraphics[width=0.8\linewidth]{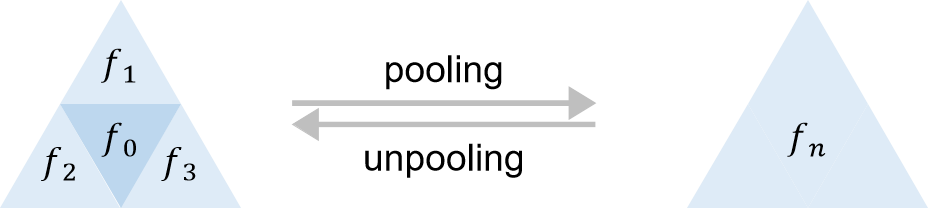}}	
  
	\end{center}
    \vspace{-2.0em}
	\caption{
		Mesh Operations includes Mesh Convolution and the Mesh Pooling/Unpooling \cite{yan2022spheredepth}, which relies on the relationship between triangles of the spherical mesh. 
	}
    \vspace{-1.5em}
	\label{fig:mesh_op}
\end{figure}

\subsection{Panorama Representation}
\label{sec:represent}

Finding a suitable way to represent the panorama image is the key to high-quality panorama depth estimation. 
The equirectangular projection \cite{zioulis2018omnidepth} suffers from distortion on the poles and discontinuity at the borders, but it can directly use 2D convolution to extract features from images. 
The cubemap and tangent projection have small distortion but need a special mechanism to fuse different patches \cite{li2022omnifusion,shen2022panoformer,peng2022high}. 
The spherical projection \cite{yan2022spheredepth} is an ideal way to represent the panorama image, but mesh convolution is hard to extract texture features compared with 2D image convolution.
In light of these, we simultaneously employ the equirectangular and spherical projections. 

The equirectangular projection uses a 2D image plane to represent the panorama image, as Fig. \ref{fig:projections}(b) shows, where the image plane is built on the latitude and longitude of the sphere surface. Given a pixel $p=(u,v)$ on the image plane, we can calculate its position on the sphere surface by Eq. \ref{eq:to_equi}, where $W$ is the image width and $H$ is the image height. 

\begin{align}
	\label{eq:to_equi}
	\begin{cases}
		longitude &= (2u/W - 1 ) \times \pi \\
		latitude &=( v/H - 0.5 ) \times \pi
	\end{cases}
\end{align}

The spherical projection is based on the icosahedron spherical mesh  \cite{eder2020tangent}, which can approximate the sphere by a higher mesh resolution (MR), where MR represents the times of loop subdivision \cite{hu2021subdivision} is applied on the icosahedron spherical mesh and determines the number of triangles in the spherical mesh by $20 \times 4^{MR}$. 
One triangle in the spherical mesh represents one pixel in equirectangular projection, as Fig. \ref{fig:projections}(c) shows, Tangent \cite{eder2020tangent} points out that a panorama image in the equirectangular projection with higher image resolution needs a spherical mesh with higher MR.
In our implementation, we use the triangle center $(x,y,z)$ to represent the whole triangle and can calculate its position on the sphere surface by Eq. \ref{eq:to_mesh}.

\begin{align}
	\label{eq:to_mesh}
	\begin{cases}
		longitude &= atan(y,x) \\
		latitude &= atan(z,\sqrt{x^2+y^2})
	\end{cases}
\end{align}

According to Eq. \ref{eq:to_equi} and Eq. \ref{eq:to_mesh}, we define the E2S ( equirectangular projection to spherical projection ) and S2E ( spherical projection to equirectangular projection ).

\subsection{Mesh Operations}
\label{sec:mesh}

As we utilize the spherical projection to represent the panorama image, we need mesh convolution and the mesh pooling/unpooling to extract features, which is inspired by SubdivNet \cite{hu2021subdivision} and SphereDepth \cite{yan2022spheredepth}.

Mesh convolution relies on the FAF ( face adjacent face ), which describes the topological relationship between triangles of the spherical mesh, as Fig. \ref{fig:mesh_op}(a) shows. Each triangle in the mesh has three neighbors and can extract features by linear interpolation by Eq. \ref{eq:mesh_conv}, where $w_i (i=0,1,2,3)$ are the weight parameters, $b_0$ is the bias parameter, $f_0$ is the feature of the center triangle, $f_1$, $f_2$, $f_3$ are the feature of adjacent triangles, $f_n$ is the extracted features.

\begin{align}
	\label{eq:mesh_conv}
	f_{n}= \sum_{i=0}^{3}w_if_{i}+b_0
\end{align}

The mesh pooling and unpooling are fundamental components of constructing an encoder-decoder structure. 
Similar to the image pooling, mesh pooling merges four triangles into one, as shown in Fig. \ref{fig:mesh_op}(b) and calculates the feature of the new $f_{n}$ triangle through features of $f_0$, $f_1$,$f_2$, and $f_3$ by the mesh max pooling.
The mesh unpooling is the opposite of the mesh pooling, which splits one triangle into four triangles by loop subdivision.

\begin{figure*} [t]
	\begin{center}
		\includegraphics[width=0.8\linewidth]{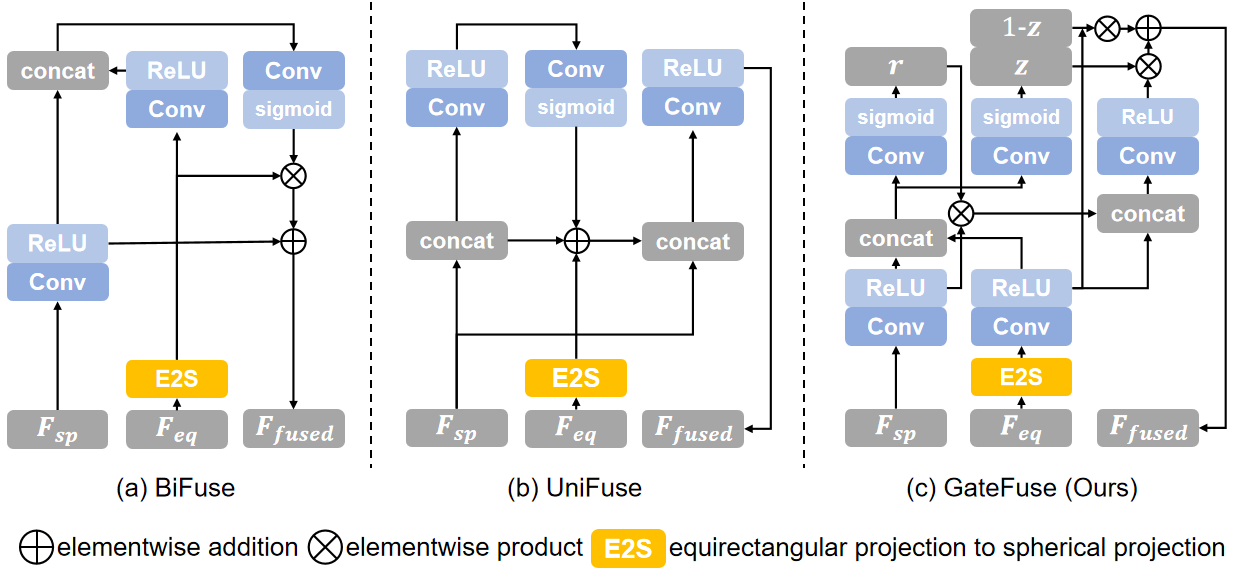}
	\end{center}
    \vspace{-1.5em}
	\caption{
		We implement BiFuse \cite{wang2020bifuse}, UniFuse \cite{jiang2021unifuse}, and our GateFuse to fuse features from spherical projection $F_{sp}$ and equirectangular projection $F_{eq}$. Unlike BiFuse and UniFuse select features from $F_{eq}$ and fuse them to $F_{sp}$, GateFuse selects features from $F_{sp}$ and $F_{eq}$.
	}
	\label{fig:fuse}
    \vspace{-1.5em}
\end{figure*}

\subsection{Our Framework}
\label{sec:pipeline}

\subsubsection{The Network Encoder}
Given a panorama image, we represent it by the spherical and equirectangular projection and employ the mesh encoder and image encoder to extract features, respectively.
The network structure of the mesh encoder follows the ResNet \cite{he2016deep}. Considering the extremely high computational complexity of mesh operations \cite{hu2021subdivision,yan2022spheredepth}, the mesh encoder uses the simplest ResNet18, and generates five scales of spherical features $F_{sp}$, of which the channels are 64, 64, 128, 256, and 512. SphereFusion randomly initializes the mesh encoder and trains it from scratch.
The image encoder directly uses the ResNet50 to extract image features $F_{eq}$, of which the channels are 64, 256, 512, 1024, and 2048. 
As 2D image convolution can not capture reliable features on distorted regions \cite{zioulis2018omnidepth}, we add a lightweight image encoder to extend the receptive field. Meanwhile, the image decoder aligns the number of channels of $F_{eq}$ to $F_{sp}$ to facilitate the subsequent feature fusion. SphereFusion initializes the image encoder by a pre-trained weight and randomly initializes the image decoder.

\subsubsection{The Fusion Module}

After extracting features from the spherical projection $F_{sp}$ and the equirectangular projection $F_{eq}$, SphereFusion uses the GateFuse module to fuse these features in the spherical domain. 
The core idea of the GateFuse module is to enhance the $F_{sp}$ by $F_{eq}$ through a reset gate and a forget gate inspired by GRU \cite{cho2014learning}.
For extracted features in each scale, GateFuse first transforms $F_{eq}$ to the spherical projection through the E2S module, then concatenates these features to estimate a reset gate value $r$ and a forget gate value $z$, where $r$ selects features from $F_{sp}$ and $z$ selects features from $F_{eq}$. Finally, GateFuse adds these features to get fused features $F_{fused}$.
Compared with BiFuse \cite{wang2020bifuse} and UniFuse \cite{jiang2021unifuse}, which use concatenate features to estimate a mask to select reliable features from $F_{eq}$, GateFuse simultaneously selects reliable features from $F_{sp}$ and $F_{eq}$, instead of simply trusting one type of features.
To better compare different fusion modules, we implement BiFuse, UniFuse, and our GateFuse and visualize these modules in Fig. \ref{fig:fuse}.

\subsubsection{The Network Decoder}
With fused features, we construct a mesh decoder to estimate the panorama depth. 
Following the network structure of UNet \cite{ronneberger2015u}, SphereFusion uses skip-connection to concatenate fused features $F_{fused}$ with features from the mesh encoder $F_{sp}$, and uses the mesh unpooling to reconstruct a high-resolution panorama depth map. 
Specifically, the mesh decoder has six layers, and the number of channels in each layer are 1024, 512, 64, 32, 32, and 32, respectively. 
Meanwhile, each layer contains one mesh unpooling layer to gradually increase the MR of the spherical mesh from 2 to 7 for 360D and 3 to 8 for Matterport3D and Stanford2D3D. The mesh decoder outputs multi-resolution panorama depth maps to speed up the training procedure.

\subsubsection{Inference Efficiency}
Unlike 2D image convolution, which can directly find adjacent pixels by coordinate, the mesh operation needs to compute the FAF of each triangle to identify nearby triangles in each layer, which will become more complex with higher MR. 
However, as the mesh pooling/unpooling changes the spherical mesh, mesh convolution layers between two mesh pooling/unpooling layers use a spherical mesh with the same MR. 
Meanwhile, the mesh decoder and the mesh encoder at the same scale use a spherical mesh with the same MR. 
Based on these findings, SphereFusion first identifies all active spherical meshes, then merges these spherical meshes with the same MR, and then calculates the corresponding FAF once.
During training and testing, SphereFusion stores these connectivity information in cache memory without recalculating FAF for each layer, which can significantly improve inference efficiency.

\subsection{Loss Function}
\label{sec:loss}

Following recent works \cite{pintore2021slicenet,wang2020bifuse}, we use the BerHu loss \cite{laina2016deeper} during training as Eq. \ref{eq:loss} shows, where $y$ is the ground truth depth and $\hat{y}$ is the predicted depth, and the threshold $T$ is set to 0.2 in all our experiments.

\begin{align}
	\label{eq:loss}
	\mathcal{L}(y,\hat{y}) = 
	\begin{cases}
		|y-\hat{y}|, &|y-\hat{y}|<T \\
		\frac{(y-\hat{y})^2+T^2}{2T}, &|y-\hat{y}|\ge T
	\end{cases}
\end{align}

To accelerate the training, SphereFusion predicts depth maps with multiple scales and extendsEq. \ref{eq:loss} to multi-scale, as Eq. \ref{eq:multi_loss} shows, where $s$ is the scale, $s_i$ is the weight, $V_i$ is the valid pixel, $N_i$ is the number of the valid pixel.

\begin{align}
	\label{eq:multi_loss}
	Loss = \sum_{i<s}{s_i \frac{\sum_{p \in V_i}{\mathcal{L}(y,\hat{y})}}{N_i}  }
\end{align}

\section{Experiments}
\label{sec:experiment}

\subsection{Datasets}


\textbf{360D} \cite{zioulis2018omnidepth} is a synthetic panorama dataset and contains 35977 panorama images with a resolution of $256 \times 512$. 

\noindent \textbf{Matterport3D} \cite{chang2017matterport3d} is a large real-world dataset, which has 10800 panorama images. We resize all panorama images and depth maps into $512 \times 1024$ during training and testing.

\noindent \textbf{Stanford2D3D} \cite{armeni2017joint} is a real-world indoor dataset that contains 1413 panoramas. We resize all panorama images and depth maps into $512 \times 1024$.

\subsection{Implementation}

We implement our method by Jittor \cite{hu2020jittor}.
On 360D, we train the network with 30 epochs, setting the batch size to 4 and the initial learning rate to 0.0002. 
On Matterport3D and Stanford2D3D, we train the network with 30 epochs, setting the batch size to 2 and the initial learning rate to 0.0001. 
On all datasets, we train on one Nvidia Tesla V100 and halve the learning rate after every ten epochs. 
Since Matterport3D and Stanford2D3D utilize the same sensors to capture panorama images, and the size of Stanford2D3D is relatively small, we combine their training data but test each dataset separately.

\subsection{Quantitative Evaluation}

During the quantitative evaluation,  we follow common evaluation metrics \cite{wang2020bifuse} and use MAE, MRE, RMSE, RMSE (log), and $\delta$ and compare with state-of-the-art panorama depth estimation methods \cite{wang2020bifuse,jiang2021unifuse,pintore2021slicenet,shen2022panoformer,li2022omnifusion,yan2022spheredepth,sun2021hohonet}.
To deal with outliers, we ignore pixels whose depth is outside of the range $0.1 \sim 10$ meters for 360D \cite{zioulis2018omnidepth} and $0.1 \sim 16$ for Stanford2D3D \cite{armeni2017joint} and Matterport3D \cite{chang2017matterport3d}. 
Table \ref{tab:sota_results} shows evaluation results on three datasets, including the quality of the depth map and the inference efficiency.

\begin{table*} [ht]
	\begin{center}
		\caption{Quantitative evaluation results of depth maps, where `S2D3D' is the short for Standard2D3D and `M3D' is the short for Matterport3D. `---' means no data is available from the original paper. 
		We mark out top three methods for better comparison. 
		}
		\label{tab:sota_results}

        \vspace{-0.50em}
  
		\resizebox{\linewidth}{!}{\begin{tabular}{l|lcccc|ccc|c}
			\hline\noalign{\smallskip}
			Dataset & Method & MRE$\downarrow$ & MAE$\downarrow$ & RMSE$\downarrow$ & RMSE(log)$\downarrow$ &$\delta_1\uparrow$ &$\delta_2\uparrow$ &$\delta_3\uparrow$ & Time(s)$\downarrow$ \\
			
			\noalign{\smallskip}
			\hline
			\noalign{\smallskip}
			
			\multirow{11}{*}{S2D3D}
			& FCRN \cite{laina2016deeper} & 0.1837 & 0.3428 & 0.5774 & 0.1100 & 0.7230 & 0.9207 & 0.9731 & ---\\ 
			& OmniDepth \cite{zioulis2018omnidepth} & 0.1996 & 0.3743 & 0.6152 & 0.1212 & 0.6877 & 0.8891 & 0.9578 & ---\\
			& BiFuse  \cite{wang2020bifuse}  & 0.1209 & 0.2343 & 0.4142 & 0.0787 & 0.8660 & 0.9580 & 0.9860 & 0.7825\\ 
			& SliceNet$^*$ \cite{pintore2021slicenet} & $\textbf{0.0998}^3$ & $\textbf{0.1737}^2$ & 0.3728 & 0.0765 & 0.9038 & 0.9623 & 0.9843 & 0.0668 \\
			& UniFuse \cite{jiang2021unifuse}       & 0.1114 & 0.2082 & 0.3691 & $\textbf{0.0721}^3$   & 0.8711     & 0.9664     & 0.9882 & $\textbf{0.0247}^2$    \\
			& HohoNet \cite{sun2021hohonet} & 0.1014 & $\textbf{0.2027}^3$ & 0.3834 & $\textbf{0.0668}^2$ & $\textbf{0.9054}^3$ & 0.9693 & 0.9771 & $\textbf{0.0400}^3$ \\ 
			& PanoFormer \cite{shen2022panoformer} & --- & --- & $\textbf{0.3083}^1$ &   ---  & $\textbf{0.9394}^1$     & $\textbf{0.9838}^1$     & $\textbf{0.9941}^1$  & 0.1253   \\
			& OmniFusion \cite{li2022omnifusion} & $\textbf{0.0950}^2$ &  ---  & $\textbf{0.3474}^3$ & 0.1599   & 0.8988     & $\textbf{0.9769}^2$     & $\textbf{0.9924}^2$  & 1.5885   \\
			& SphereDepth \cite{yan2022spheredepth} & 0.1158 & 0.2323 & 0.4512 & 0.0754 & 0.8666 & 0.9642 & 0.9863  & 0.0612 \\
			& \textbf{SphereFusion}     & $\textbf{0.0899}^1$ & $\textbf{0.1654}^1$ & $\textbf{0.3194}^2$ & $\textbf{0.0611}^1$   & $\textbf{0.9257}^2$     & $\textbf{0.9755}^3$     & $\textbf{0.9904}^3$ & $\textbf{0.0174}^1$  \\
			
			\noalign{\smallskip}
			\hline
			\noalign{\smallskip}
			
			\multirow{11}{*}{M3D} 
			& FCRN  \cite{laina2016deeper}    & 0.2409 & 0.4008 & 0.6704 & 0.1244 & 0.7703 & 0.9174 & 0.9617 & ---\\ 
			& OmniDepth \cite{zioulis2018omnidepth} & 0.2901 & 0.4838 & 0.7643 & 0.1450 & 0.6830 & 0.8794 & 0.9429 & ---\\ 
			& BiFuse \cite{wang2020bifuse}  & 0.2048 & 0.3470 & 0.6259 & 0.1134 & 0.8452 & 0.9319 & 0.9632 & 0.7825\\ 
			& SliceNet \cite{pintore2021slicenet} & 0.1764 & 0.3296 & 0.6133 & 0.1045 & 0.8716 & 0.9483 & 0.9716 & 0.0668 \\
			& UniFuse \cite{jiang2021unifuse} & $\textbf{0.1063}^2$ & \textbf{0.2814} & 0.4941 & \textbf{0.0701}   & $\textbf{0.8897}^3$     & $\textbf{0.9623}^3$     & 0.9831  & $\textbf{0.0247}^2$    \\
			& HohoNet \cite{sun2021hohonet} & 0.1488 & $\textbf{0.2862}^3$ & 0.5138 & 0.0871 & 0.8786 & 0.9519 & 0.9771 & $\textbf{0.0400}^3$ \\ 
			& PanoFormer \cite{shen2022panoformer} & --- & --- & $\textbf{0.3635}^1$ &  ---  & $\textbf{0.9184}^2$     & $\textbf{0.9804}^1$  &  $\textbf{0.9916}^2$   & 0.1253   \\
			& OmniFusion \cite{li2022omnifusion} & $\textbf{0.0900}^1$ & --- & $\textbf{0.4261}^2$ & 0.1483   & $\textbf{0.9189}^1$     & $\textbf{0.9797}^2$     & $\textbf{0.9931}^1$  & 1.5885   \\
			& SphereDepth \cite{yan2022spheredepth} & 0.1205 & 0.3311 & 0.5922 & $\textbf{0.0806}^3$ & 0.8620 & 0.9519 & 0.9770  & 0.0612 \\
			& \textbf{SphereFusion} & $\textbf{0.1145}^3$ & $\textbf{0.2852}^2$ & $\textbf{0.4885}^3$ & $\textbf{0.0733}^2$   & 0.8701     & 0.9613     & $\textbf{0.9838}^3$ & $\textbf{0.0174}^1$  \\
			
			\noalign{\smallskip}
			\hline
			\noalign{\smallskip}
			
			\multirow{9}{*}{360D}
			& FCRN \cite{laina2016deeper} & 0.0699 & 0.1381 & 0.2833 & 0.0473 & 0.9532 & 0.9905 & 0.9966 & --- \\
			& OmniDepth \cite{zioulis2018omnidepth} & 0.0931 & 0.1706 & 0.3171 & 0.0725 & 0.9092 & 0.9702 & 0.9851 & --- \\
			& BiFuse \cite{wang2020bifuse}  & 0.0615 & 0.1143 & 0.2440 & 0.0428 & 0.9699 & 0.9927 & 0.9969 & 0.6971\\ 
			& SliceNet \cite{pintore2021slicenet} & 0.0467 & $\textbf{0.1134}^3$ & $\textbf{0.1323}^1$ & $\textbf{0.0212}^1$ & 0.9788 & 0.9952 & 0.9969& 0.0595 \\
			& UniFuse \cite{jiang2021unifuse} & $\textbf{0.0466}^3$ & $\textbf{0.0996}^2$ & 0.1968 & $\textbf{0.0315}^3$ & 0.9835 & 0.9965 & 0.9987 & $\textbf{0.0221}^2$    \\
			& PanoFormer \cite{shen2022panoformer} & --- & --- & $\textbf{0.1429}^2$ & --- & $\textbf{0.9876}^1$ & $\textbf{0.9975}^1$ & $\textbf{0.9991}^1$ & 0.1116   \\
			& OmniFusion \cite{li2022omnifusion} & $\textbf{0.0430}^2$ & --- & $\textbf{0.1808}^3$ & 0.0735 & $\textbf{0.9859}^3$ & $\textbf{0.9969}^3$ & $\textbf{0.9989}^3$ & 1.4151   \\
			& SphereDepth \cite{yan2022spheredepth} & 0.0550 & 0.1145 & 0.2364 & 0.0369 & 0.9743 & 0.9944 & 0.9978 & $\textbf{0.0545}^3$ \\
			& \textbf{SphereFusion} & $\textbf{0.0417}^1$ & $\textbf{0.0894}^1$ & 0.1813 & $\textbf{0.0286}^2$ & $\textbf{0.9869}^2$ & $\textbf{0.9970}^2$ & $\textbf{0.9989}^2$ & $\textbf{0.0155}^1$  \\
			
			\hline
			
		\end{tabular}}
	\end{center}
	\scriptsize{$^*$We recalculate all metrics using open-source models.}

    \vspace{-2.0em}
 
\end{table*}


On 360D \cite{zioulis2018omnidepth}, our method achieves the best results on MRE and MAE and ranks second on RMSE(log), $\delta_1$, $\delta_2$, and $\delta_3$.
On Matterport3D \cite{chang2017matterport3d}, our method ranks second on MRE, MAE, and RMSE(log). 
As for Stanford2D3D \cite{armeni2017joint}, our method achieves the lowest MRE, MRE, and RMSE(log) values while ranking second in RMSE and $\delta_1$. 
Overall, our method achieves competitive performance with the state-of-the-art methods. 
Notably, compared to SphereDepth \cite{yan2022spheredepth}, which only utilizes the mesh operation, SphereFusion significantly improves the quality of the depth map by the fusion strategy.
Moreover, SphereFusion achieves comparable results with only a simple ResNet structure compared to OmniFusion \cite{li2022omnifusion} and PanoFormer \cite{shen2022panoformer}, which use the ViT as the encoder, demonstrating the importance of choosing the proper projection.

\begin{figure} [!ht]
	\centering
	\captionsetup[subfigure]{labelformat=empty}

	\subfloat[]{\includegraphics[width=.18\linewidth]{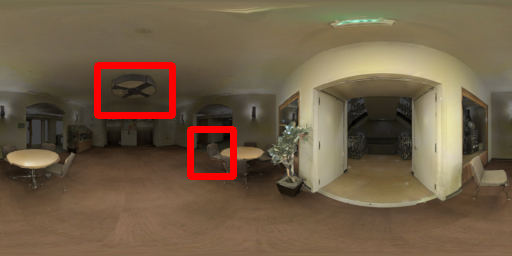}}
	\hspace{0.1em}
	\subfloat[]{\includegraphics[width=.18\linewidth]{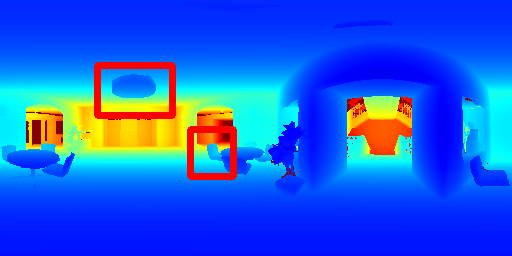}}
	\hspace{0.1em}
	\subfloat[]{\includegraphics[width=.18\linewidth]{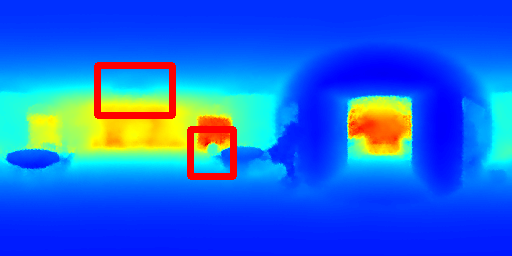}}
	\hspace{0.1em}
	\subfloat[]{\includegraphics[width=.18\linewidth]{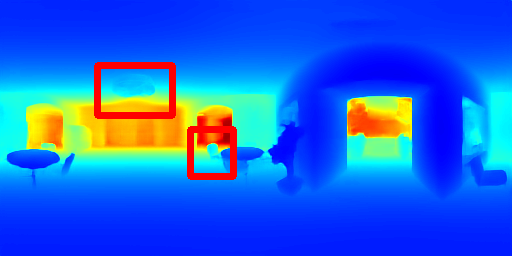}}
	\hspace{0.1em}
	\subfloat[]{\includegraphics[width=.18\linewidth]{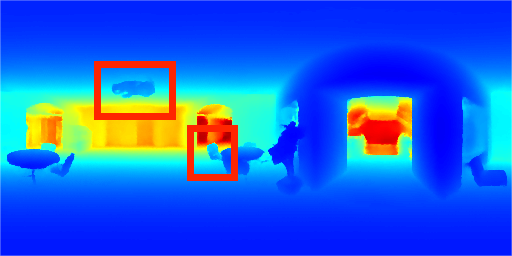}}
	
    \vspace{-1.0em}
		
	\subfloat[RGB]{\includegraphics[width=.18\linewidth]{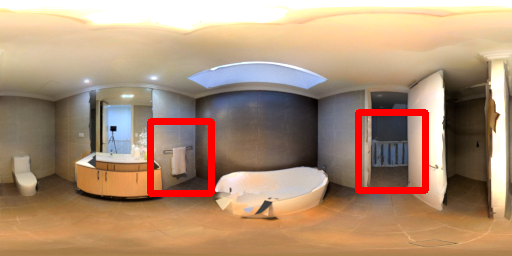}}
	\hspace{0.1em}
	\subfloat[GT]{\includegraphics[width=.18\linewidth]{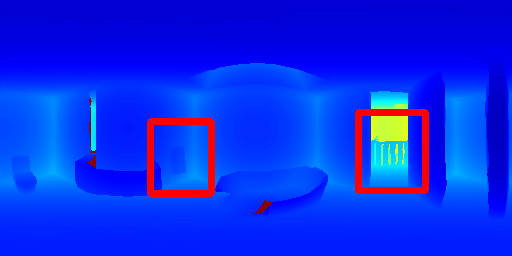}}
	\hspace{0.1em}
	\subfloat[SphereDepth]{\includegraphics[width=.18\linewidth]{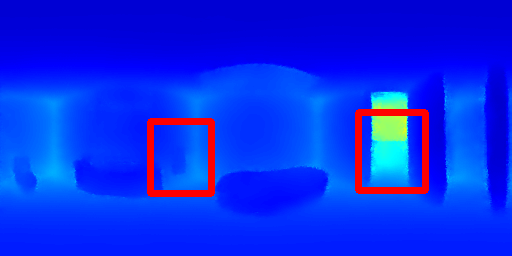}}
	\hspace{0.1em}
	\subfloat[UniFuse]{\includegraphics[width=.18\linewidth]{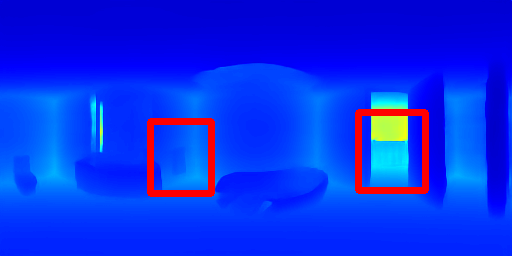}}
	\hspace{0.1em}
	\subfloat[ours]{\includegraphics[width=.18\linewidth]{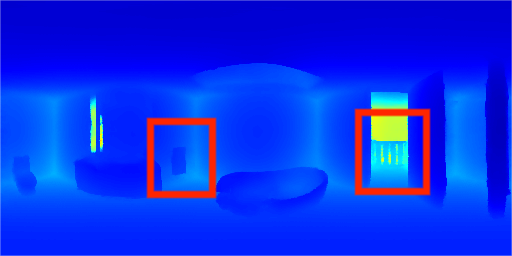}}
	
	\subfloat[]{\includegraphics[width=.23\linewidth]{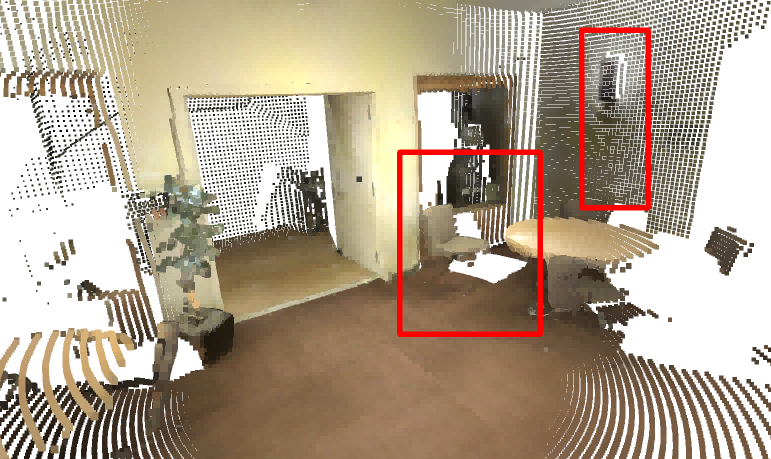}}
	\hspace{0.1em}
	\subfloat[]{\includegraphics[width=.23\linewidth]{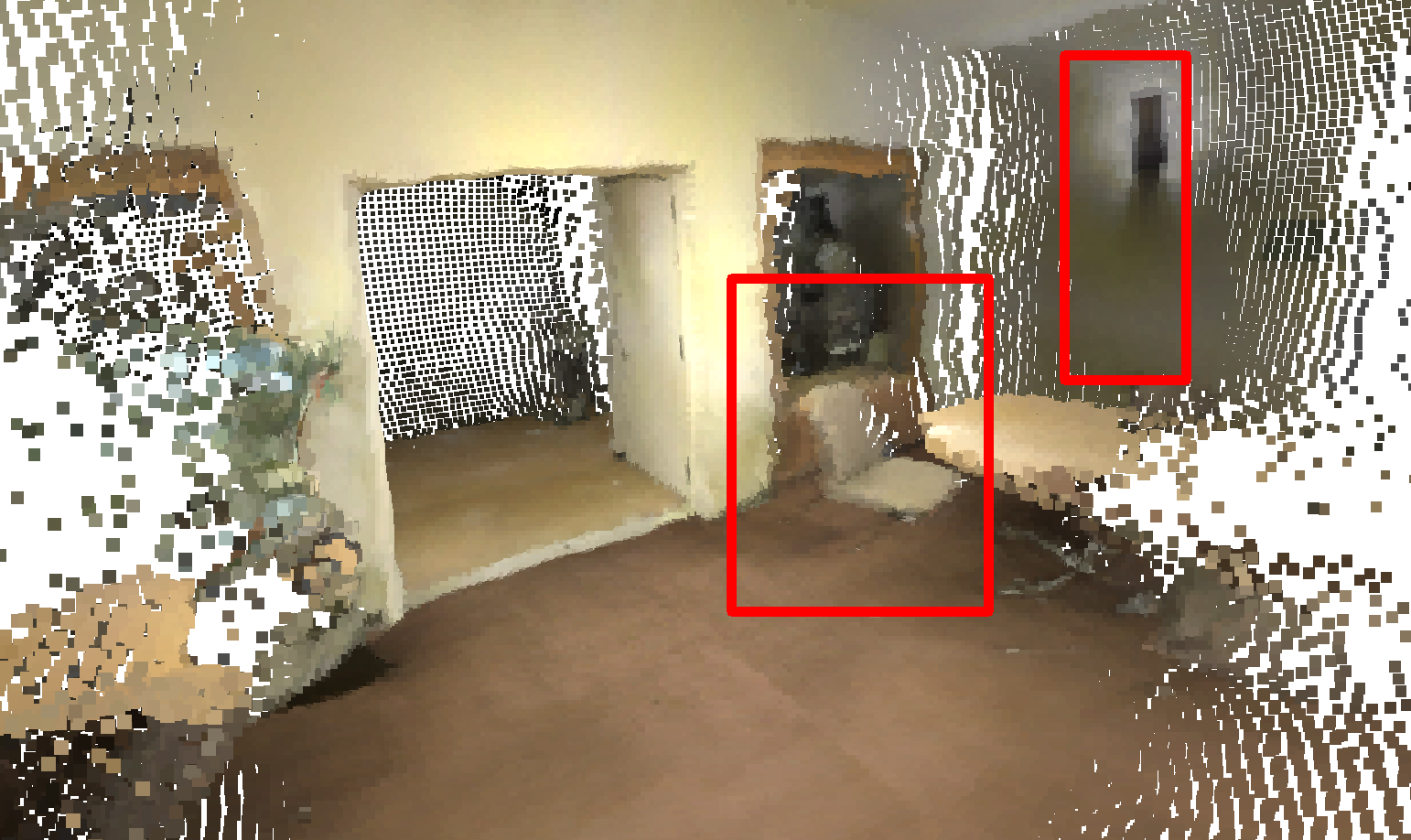}}
	\hspace{0.1em}
	\subfloat[]{\includegraphics[width=.23\linewidth]{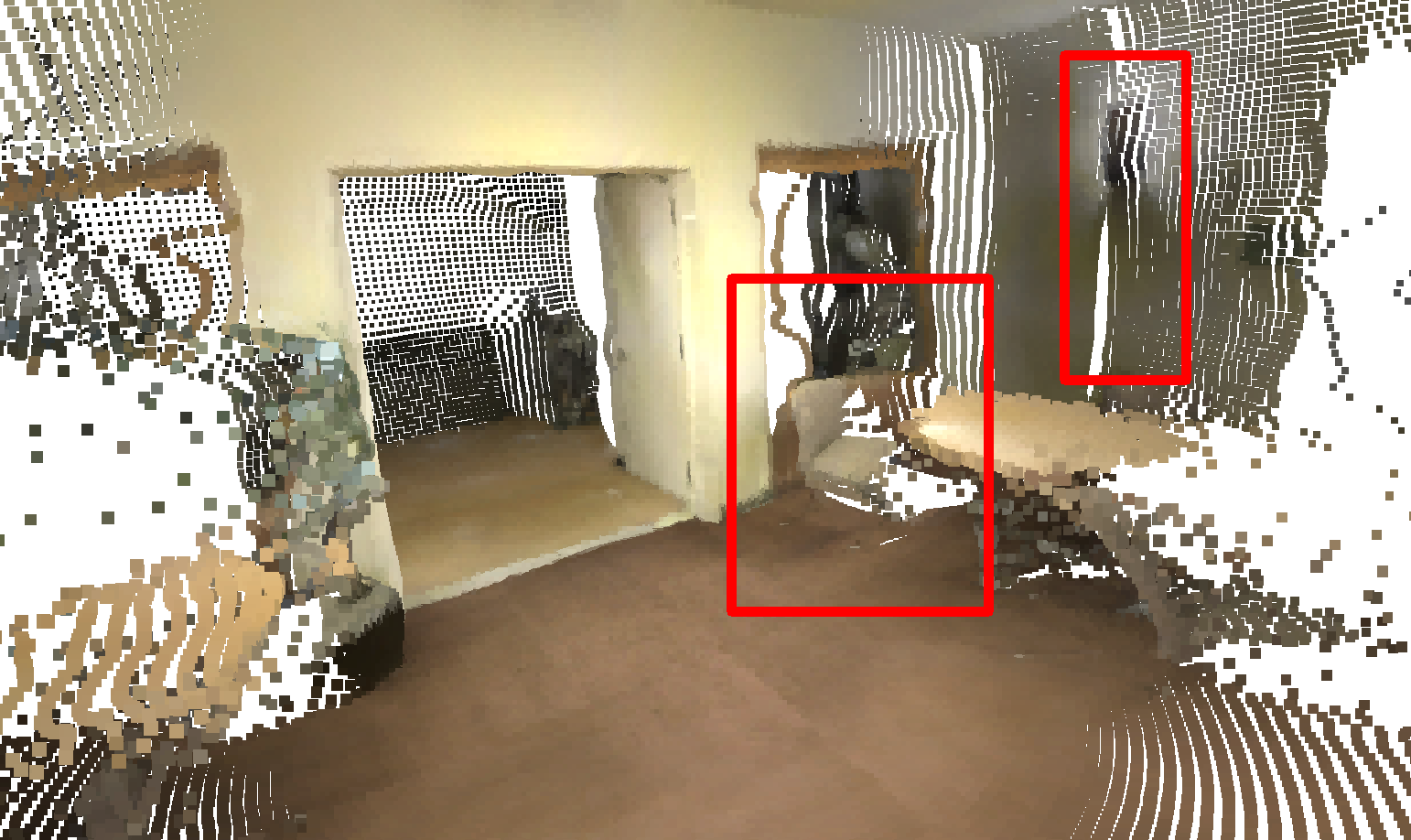}}
	\hspace{0.1em}
	\subfloat[]{\includegraphics[width=.23\linewidth]{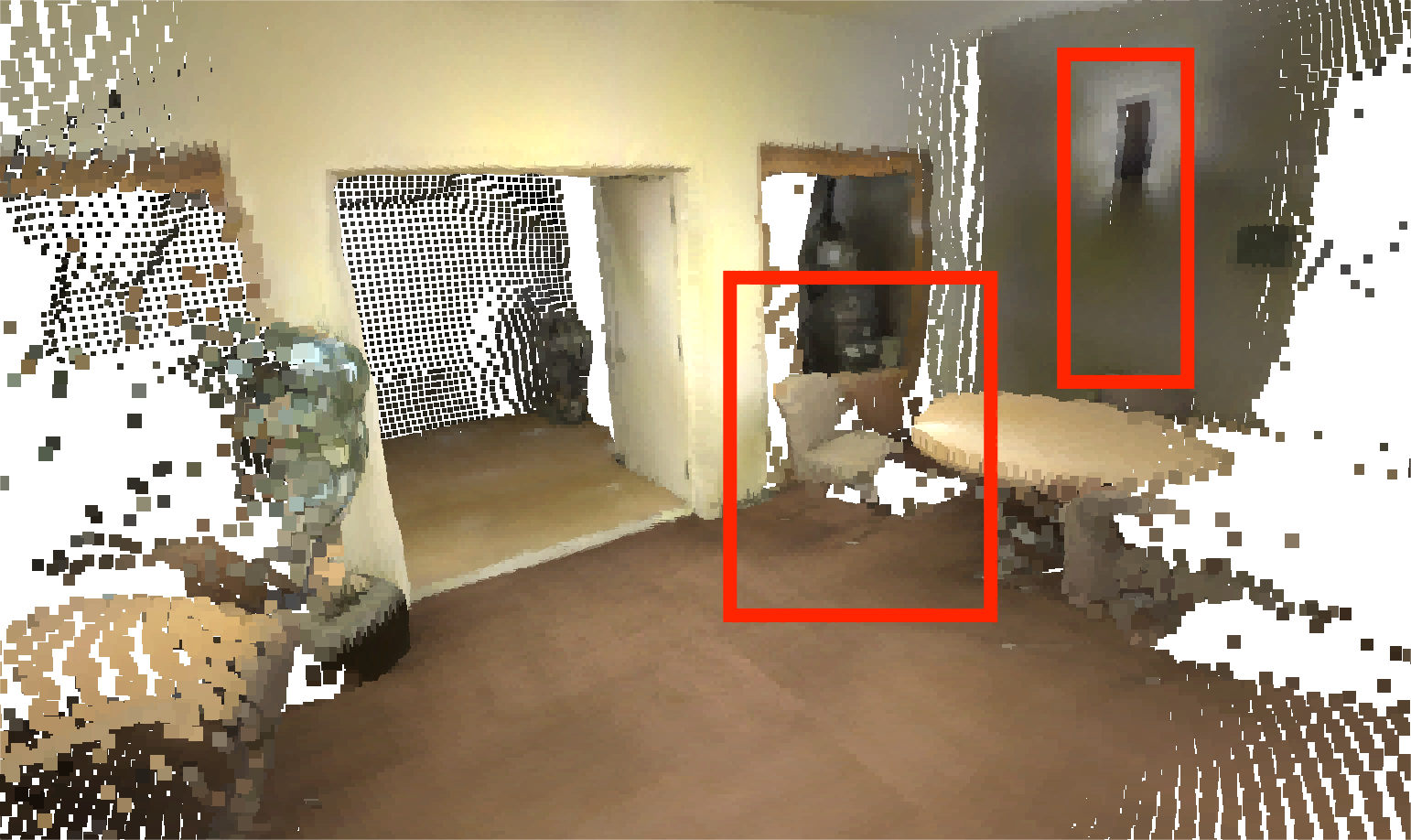}}
	
    \vspace{-1.0em}

	\subfloat[GT]{\includegraphics[width=.23\linewidth]{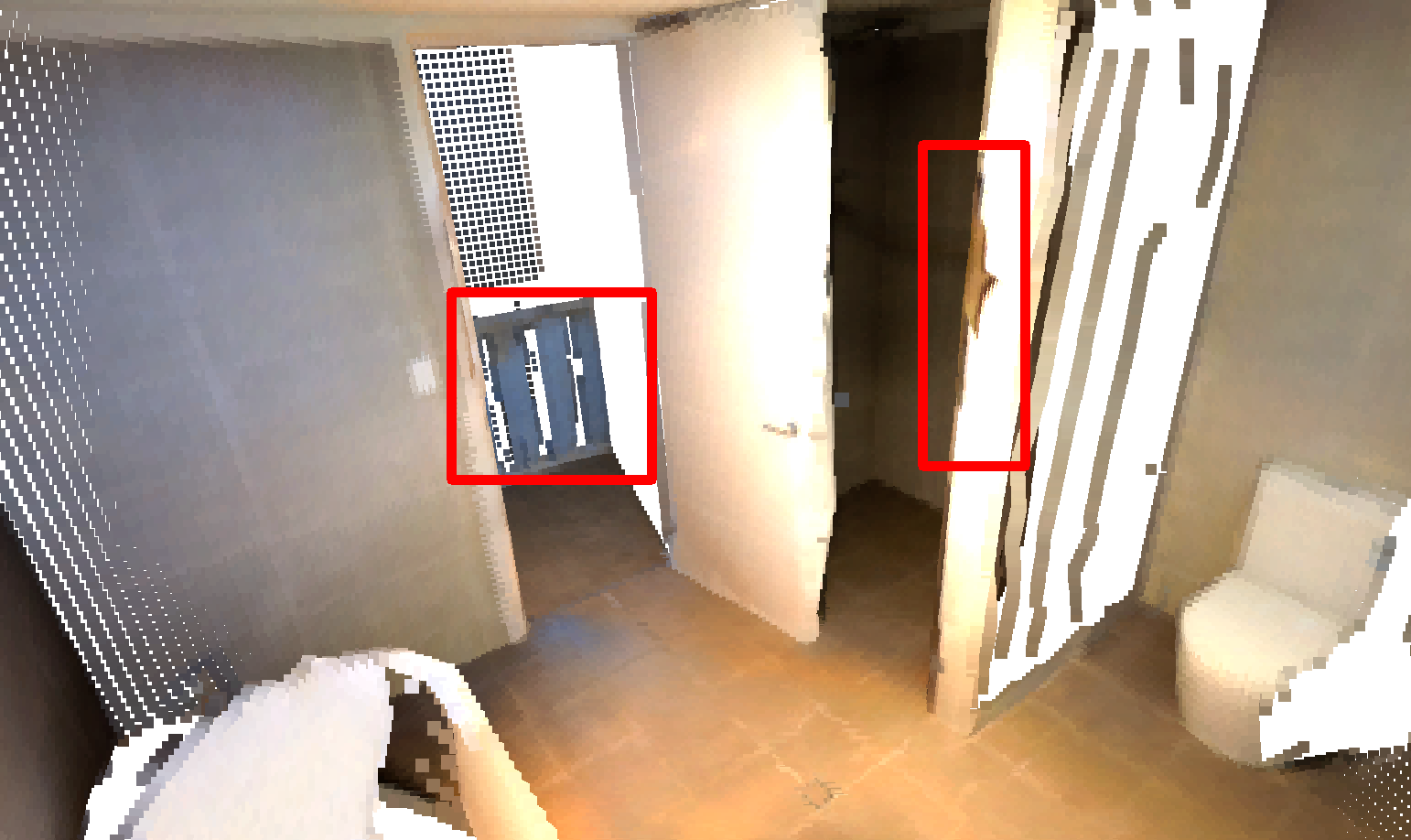}}
	\hspace{0.1em}
	\subfloat[SphereDepth]{\includegraphics[width=.23\linewidth]{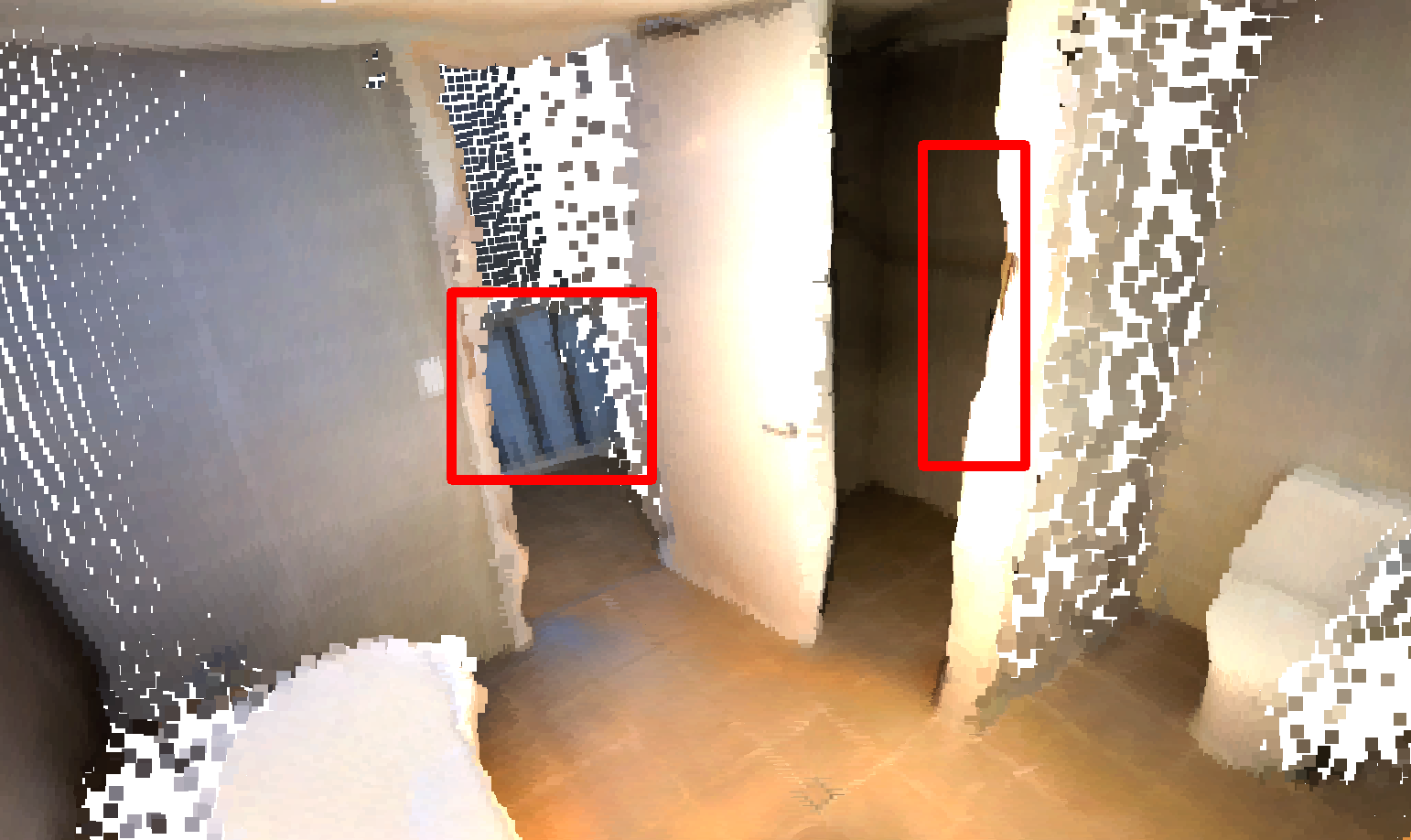}}
	\hspace{0.1em}
	\subfloat[UniFuse]{\includegraphics[width=.23\linewidth]{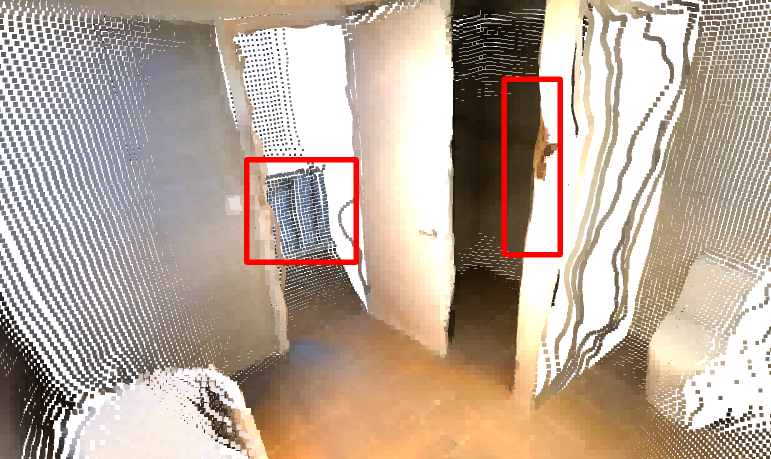}}
	\hspace{0.1em}
	\subfloat[ours]{\includegraphics[width=.23\linewidth]{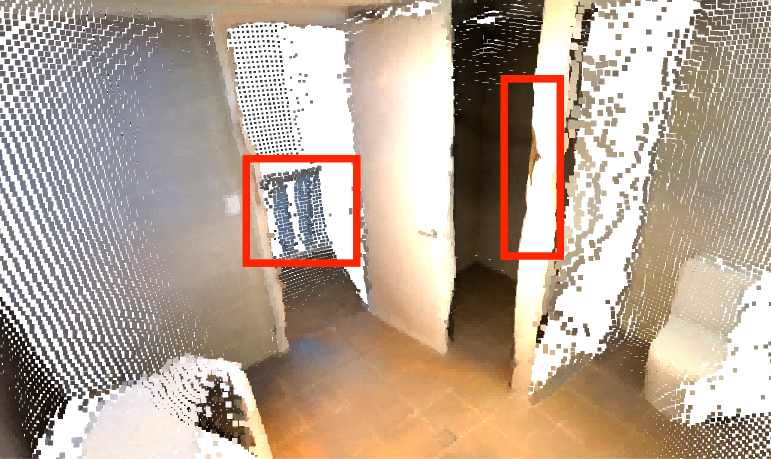}}	

    \vspace{-1.0em}
    
	\caption{
        We select two scenes from 360D \cite{zioulis2018omnidepth} and visualize depth maps and  point clouds.
	}
	\label{fig:3d60_depth_cloud}

    \vspace{-1.3em}
 
\end{figure}

\begin{figure}  [!h]
	\centering
	\captionsetup[subfigure]{labelformat=empty}
	
	\subfloat[]{\includegraphics[width=.18\linewidth]{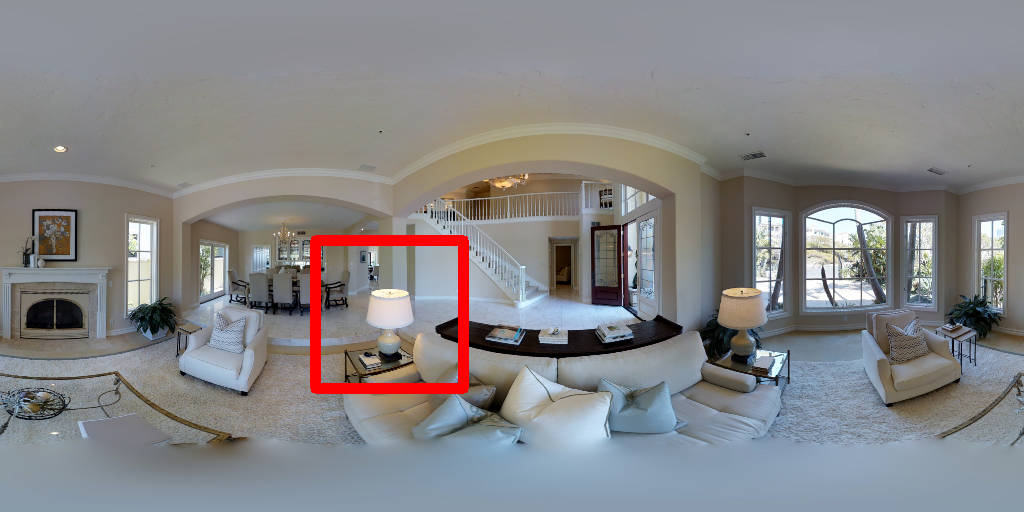}}
	\hspace{0.1em}
	\subfloat[]{\includegraphics[width=.18\linewidth]{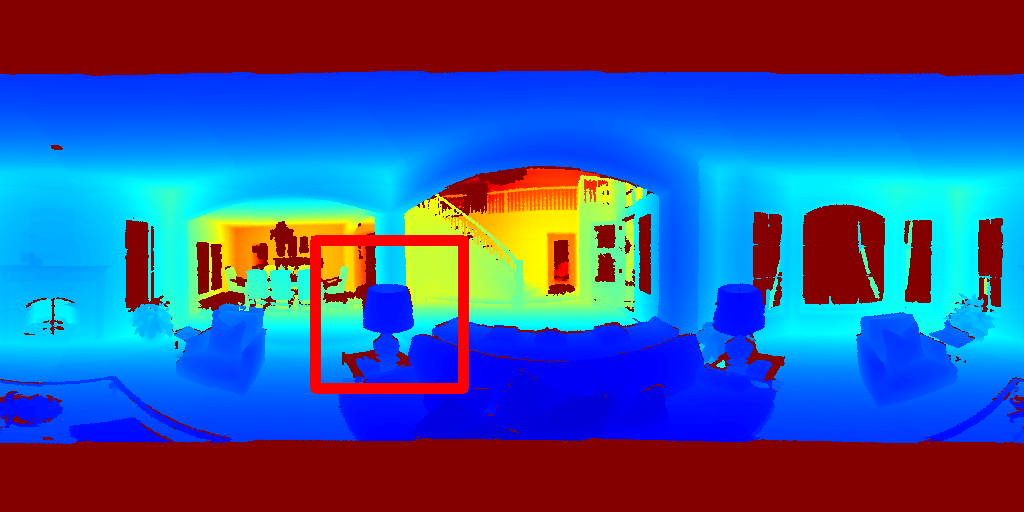}}
	\hspace{0.1em}
	\subfloat[]{\includegraphics[width=.18\linewidth]{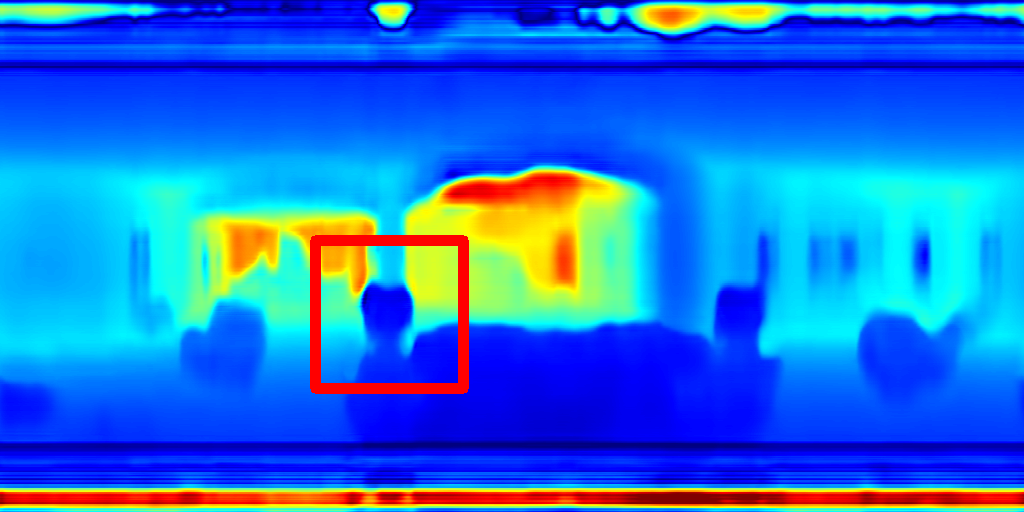}}
	\hspace{0.1em}
	\subfloat[]{\includegraphics[width=.18\linewidth]{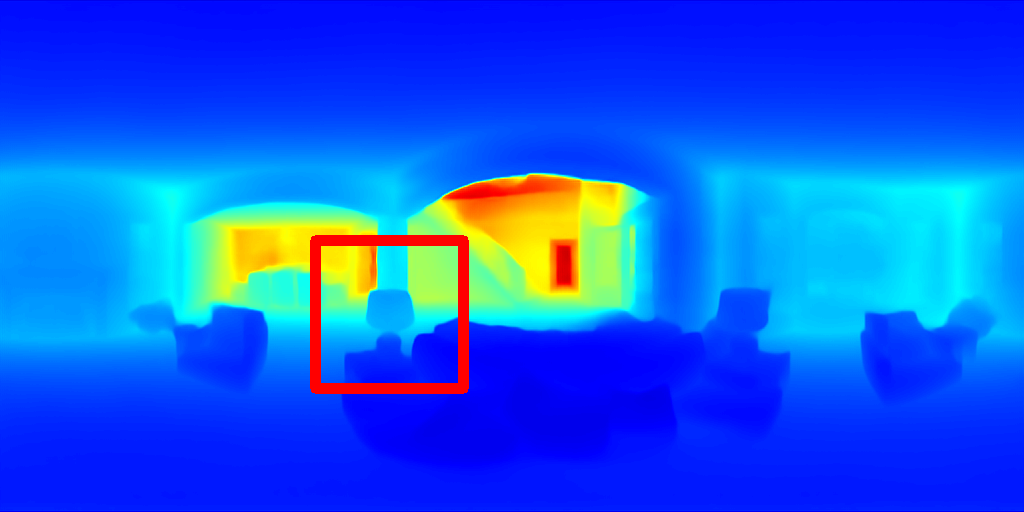}}
	\hspace{0.1em}
	\subfloat[]{\includegraphics[width=.18\linewidth]{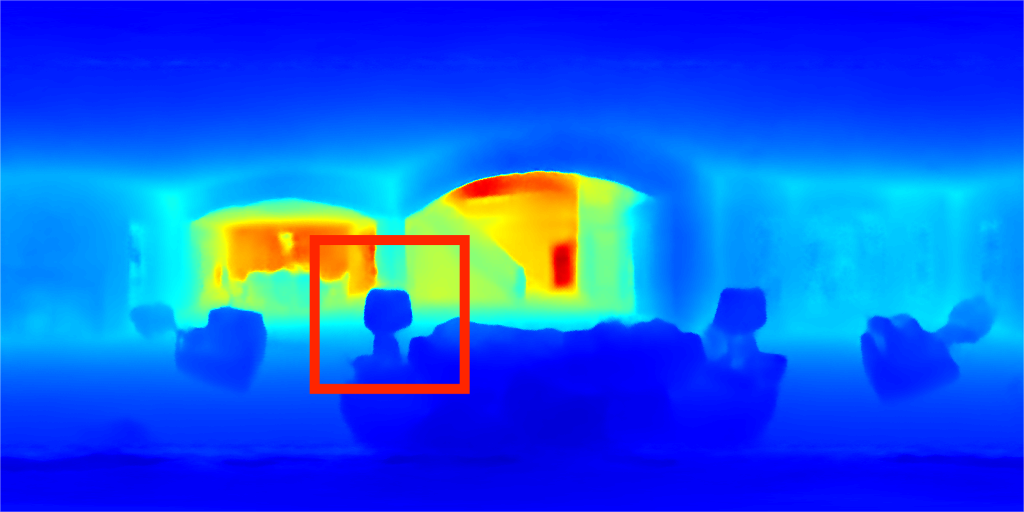}}
    
    \vspace{-1.0em}

	\subfloat[RGB]{\includegraphics[width=.18\linewidth]{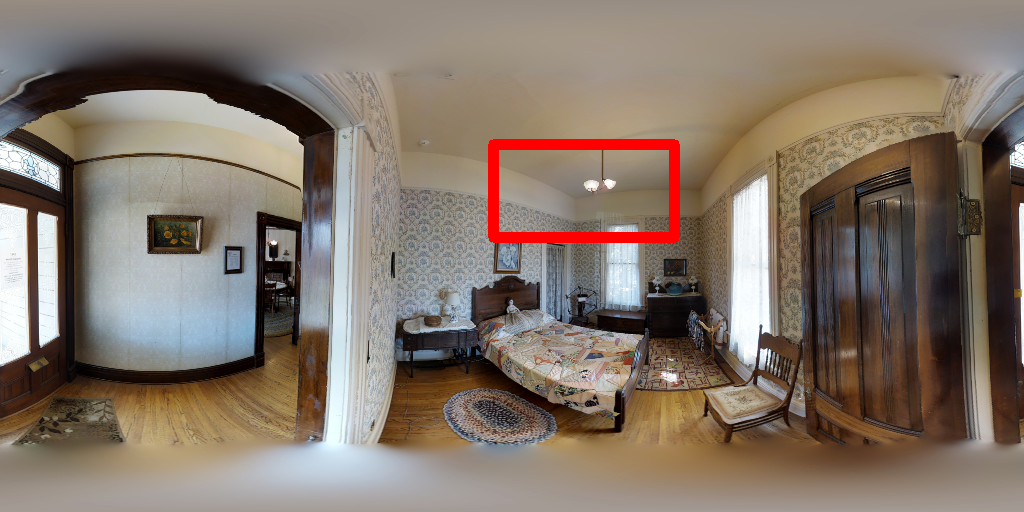}}
	\hspace{0.1em}
	\subfloat[GT]{\includegraphics[width=.18\linewidth]{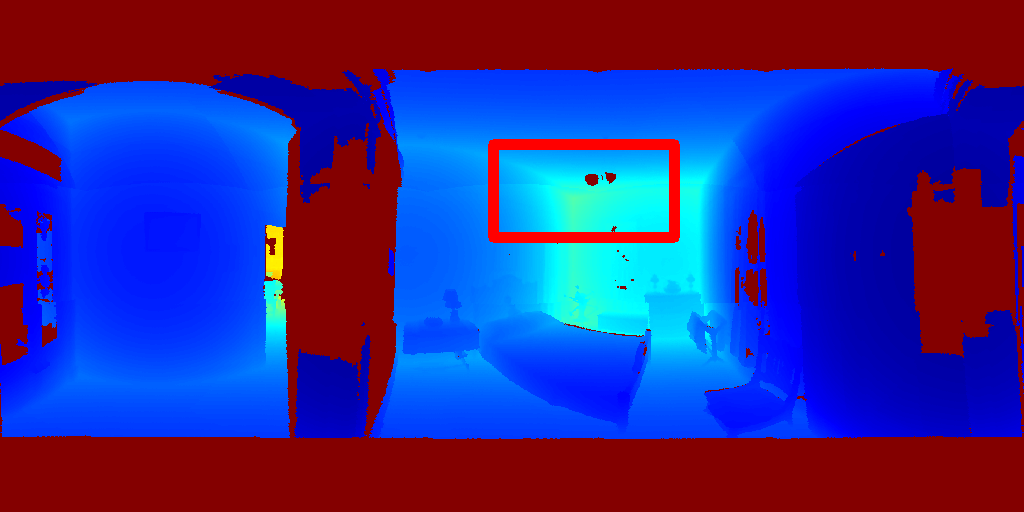}}
	\hspace{0.1em}
	\subfloat[SliceNet]{\includegraphics[width=.18\linewidth]{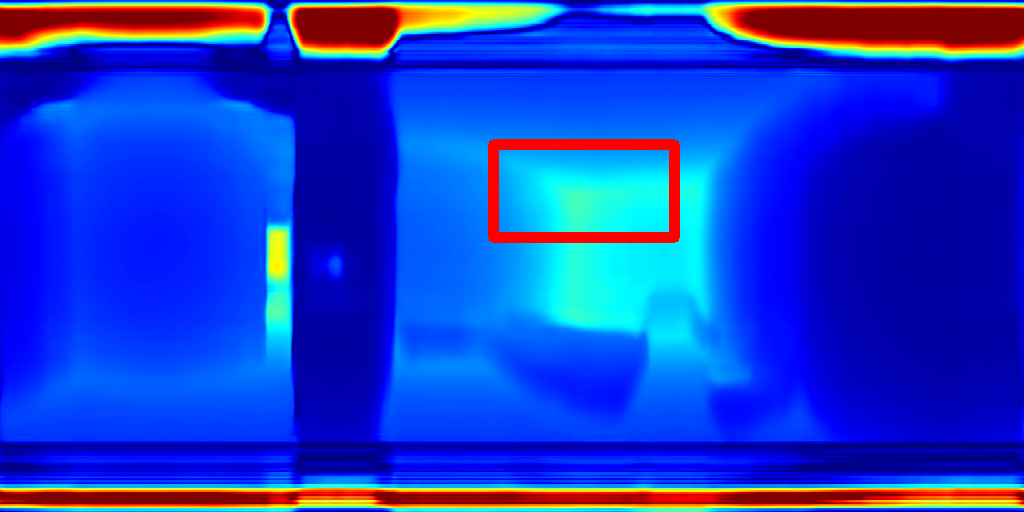}}
	\hspace{0.1em}
	\subfloat[PanoFormer]{\includegraphics[width=.18\linewidth]{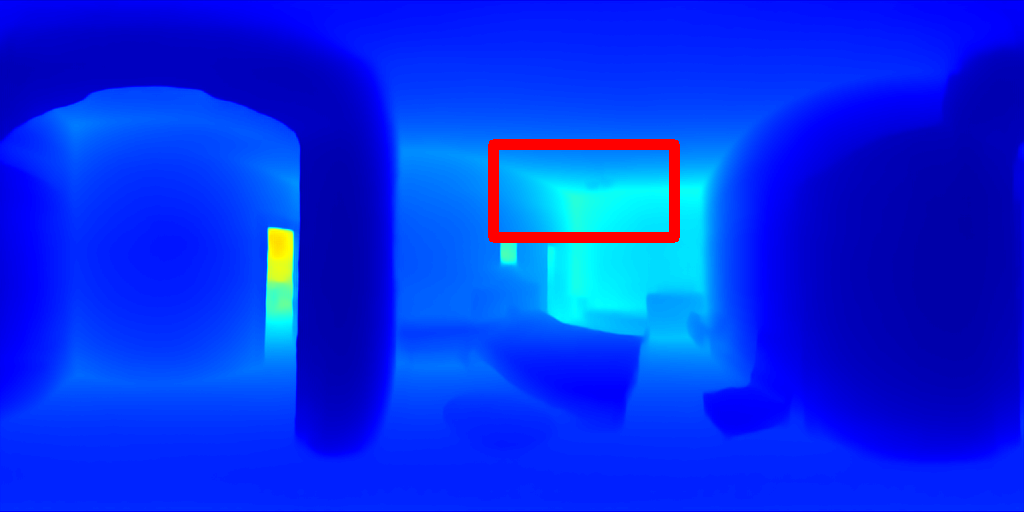}}
	\hspace{0.1em}
	\subfloat[ours]{\includegraphics[width=.18\linewidth]{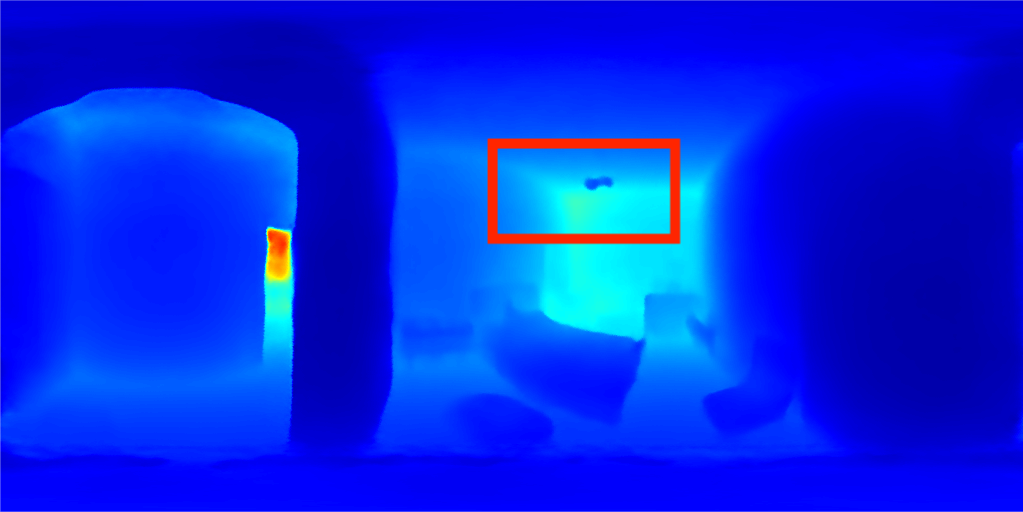}}

	\subfloat[]{\includegraphics[width=.23\linewidth]{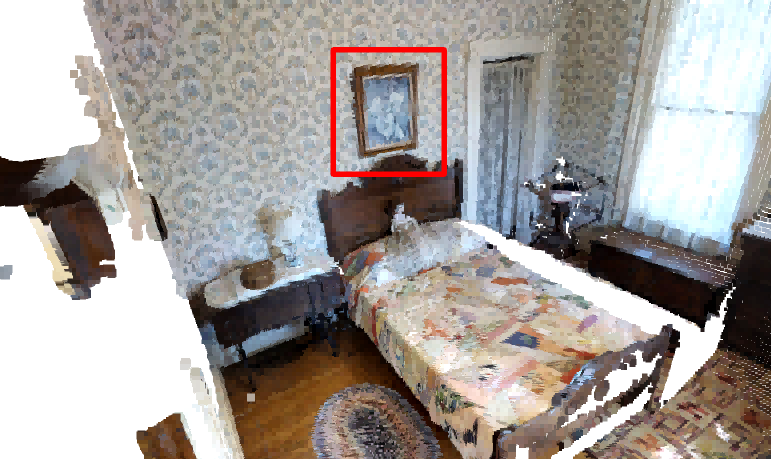}}
	\hspace{0.1em}
	\subfloat[]{\includegraphics[width=.23\linewidth]{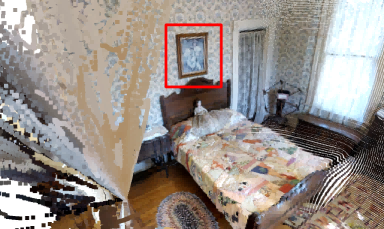}}
	\hspace{0.1em}
	\subfloat[]{\includegraphics[width=.23\linewidth]{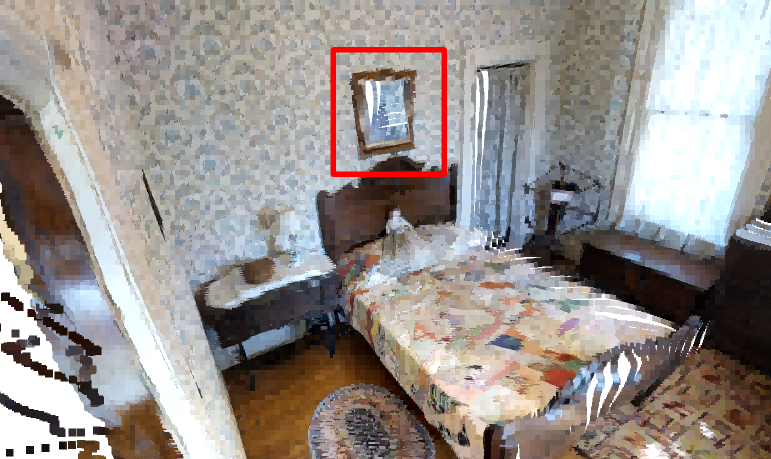}}
	\hspace{0.1em}
	\subfloat[]{\includegraphics[width=.23\linewidth]{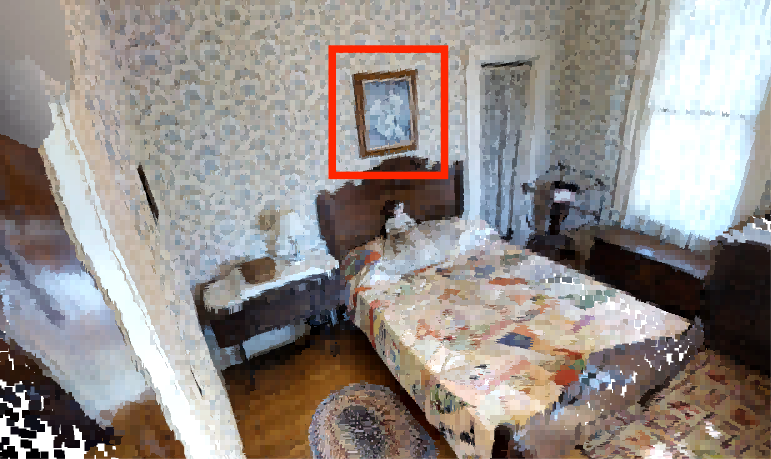}}
	
    \vspace{-1.0em}

	\subfloat[GT]{\includegraphics[width=.23\linewidth]{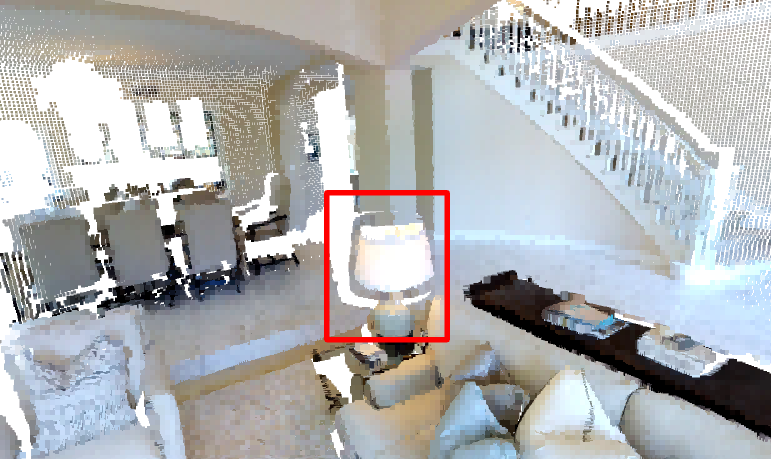}}
	\hspace{0.1em}
	\subfloat[SliceNet]{\includegraphics[width=.23\linewidth]{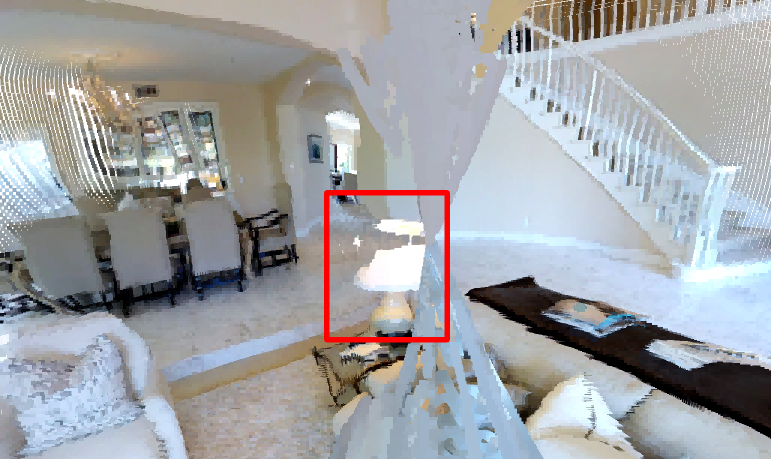}}
	\hspace{0.1em}
	\subfloat[PanoFormer]{\includegraphics[width=.23\linewidth]{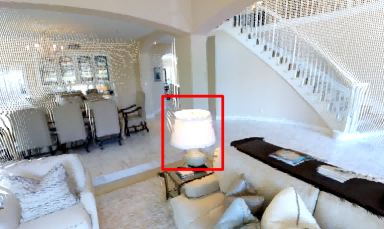}}
	\hspace{0.1em}
	\subfloat[ours]{\includegraphics[width=.23\linewidth]{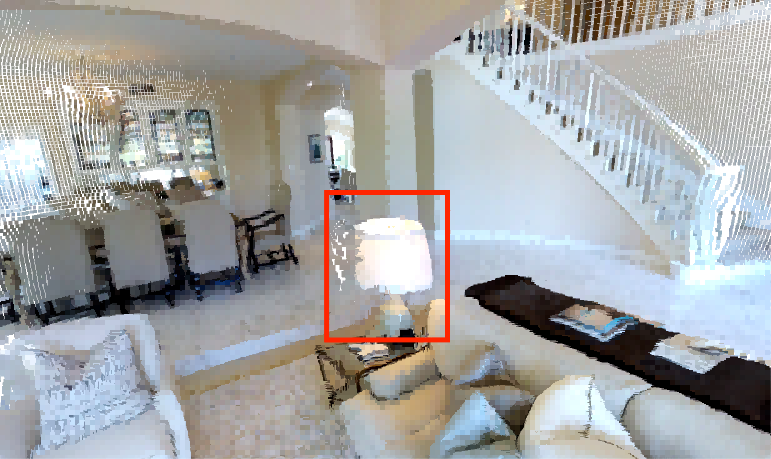}}

    \vspace{-1.0em}
	\caption{
        We select two scenes from Matterport3D \cite{chang2017matterport3d} and estimate depth maps by SliceNet \cite{pintore2021slicenet}, PanoFormer \cite{shen2022panoformer}, and our method SphereFusion. 
	}
	\label{fig:mat3d_depth_cloud}
    \vspace{-1.3em}
    
\end{figure}

\begin{figure} [t]
	\centering
	\captionsetup[subfigure]{labelformat=empty}
	
	\subfloat[]{\includegraphics[width=.18\linewidth]{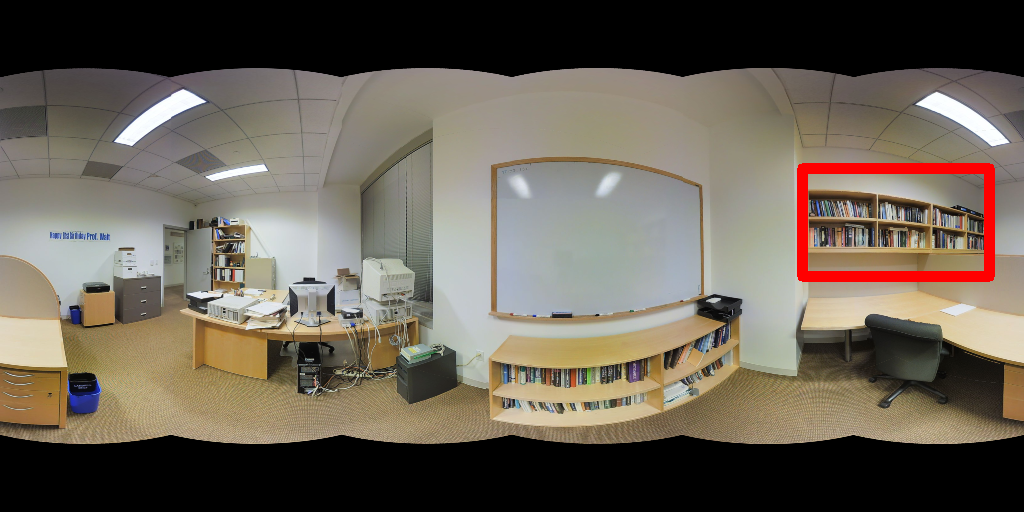}}
	\hspace{0.1em}
	\subfloat[]{\includegraphics[width=.18\linewidth]{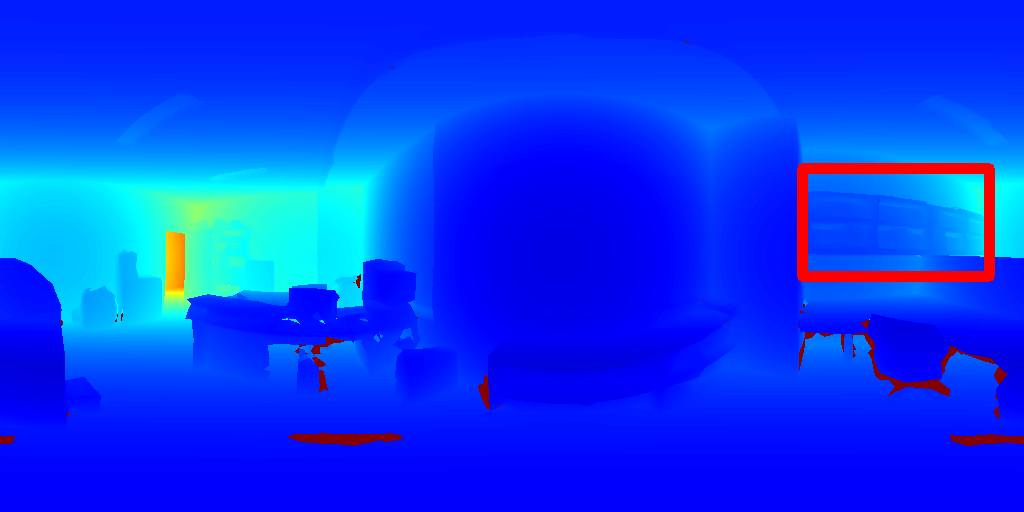}}
	\hspace{0.1em}
	\subfloat[]{\includegraphics[width=.18\linewidth]{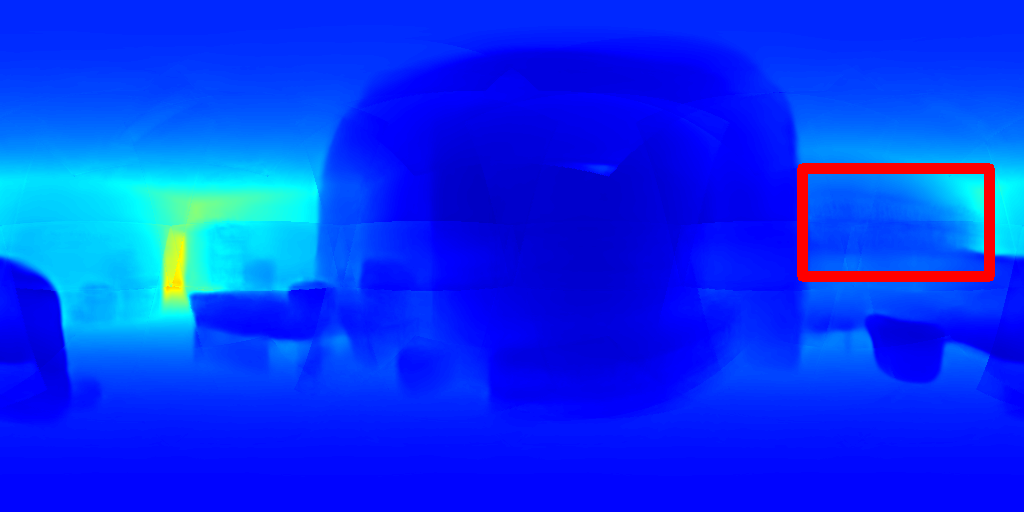}}
	\hspace{0.1em}
	\subfloat[]{\includegraphics[width=.18\linewidth]{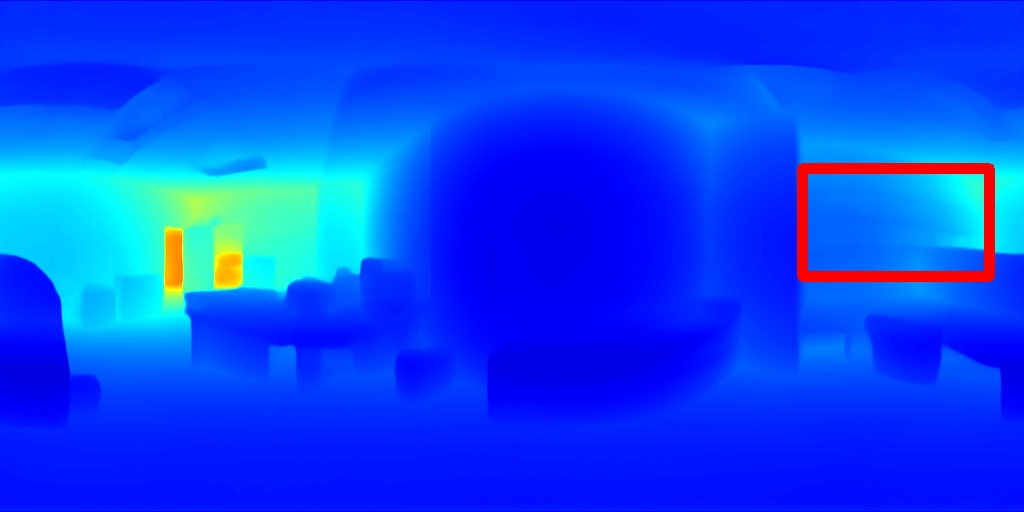}}
	\hspace{0.1em}
	\subfloat[]{\includegraphics[width=.18\linewidth]{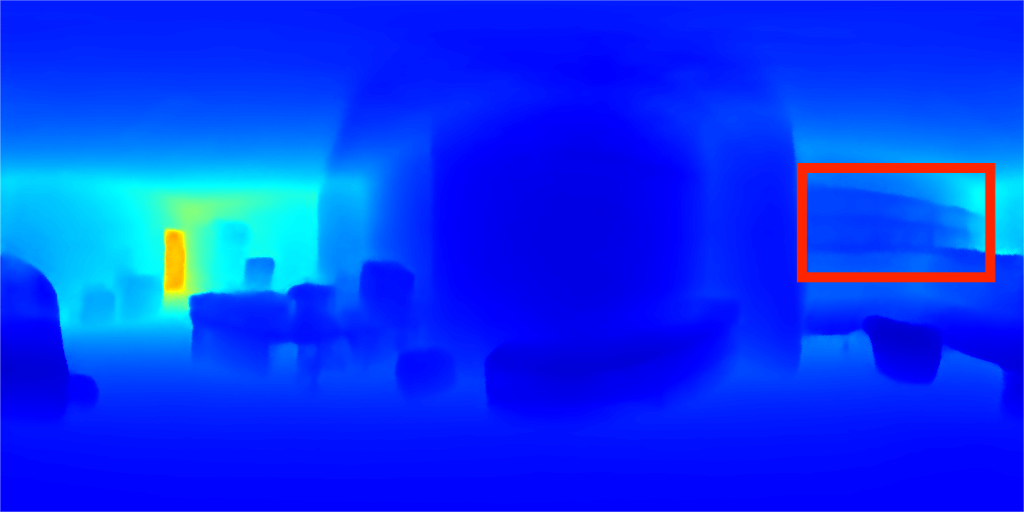}}
 
	\vspace{-1.0em}
	
	\subfloat[RGB]{\includegraphics[width=.18\linewidth]{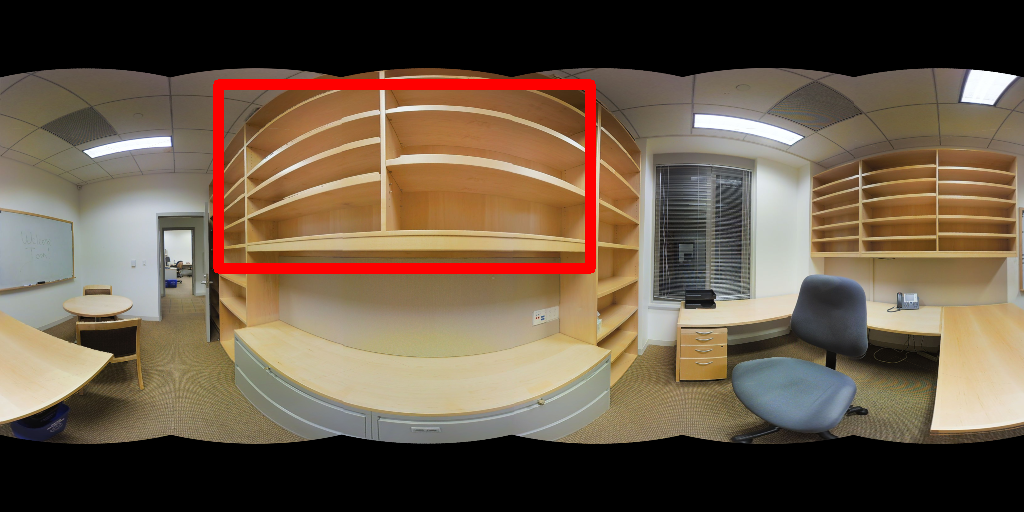}}
	\hspace{0.1em}
	\subfloat[GT]{\includegraphics[width=.18\linewidth]{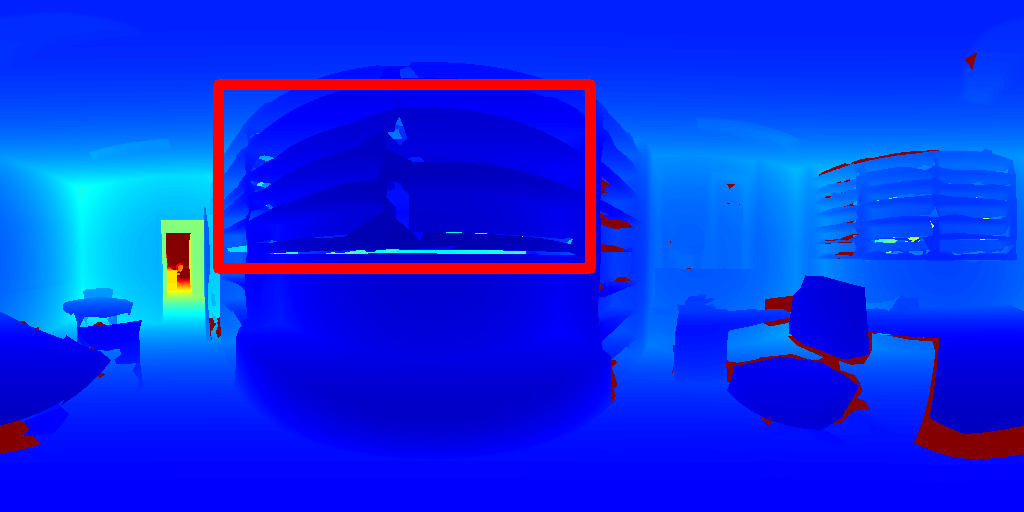}}
	\hspace{0.1em}
	\subfloat[OmniFusion]{\includegraphics[width=.18\linewidth]{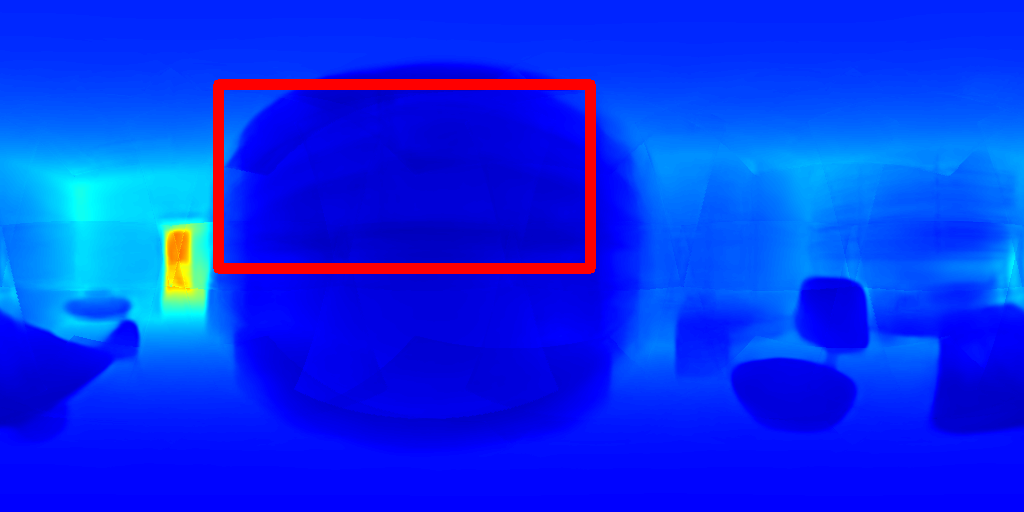}}
	\hspace{0.1em}
	\subfloat[PanoFormer]{\includegraphics[width=.18\linewidth]{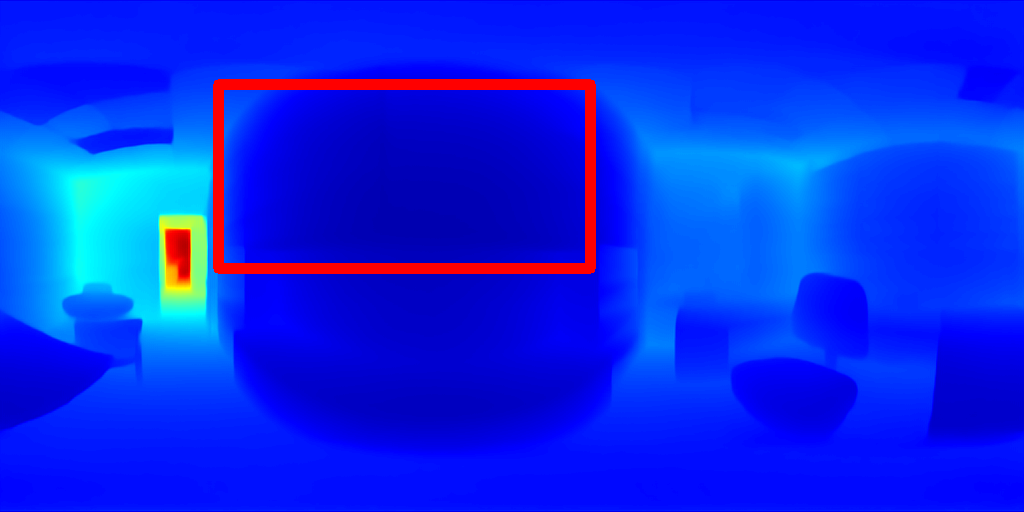}}
	\hspace{0.1em}
	\subfloat[ours]{\includegraphics[width=.18\linewidth]{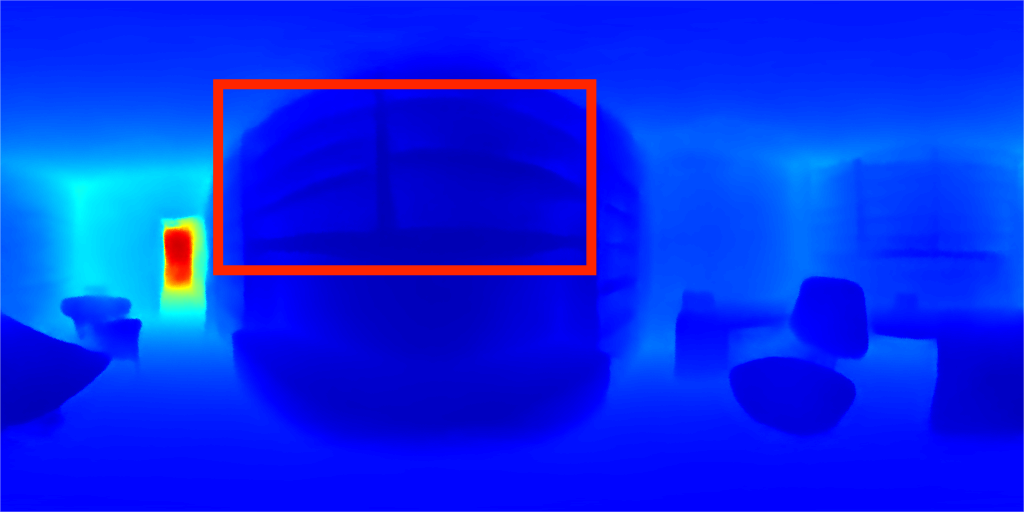}}
	
	\subfloat[]{\includegraphics[width=.23\linewidth]{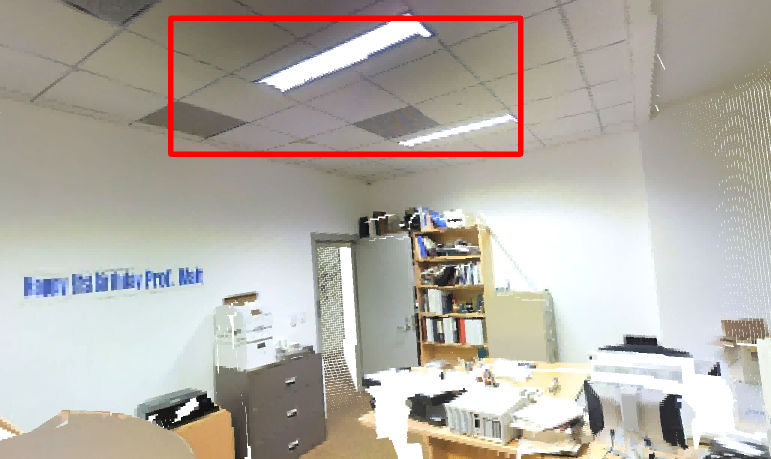}}
	\hspace{0.1em}
	\subfloat[]{\includegraphics[width=.23\linewidth]{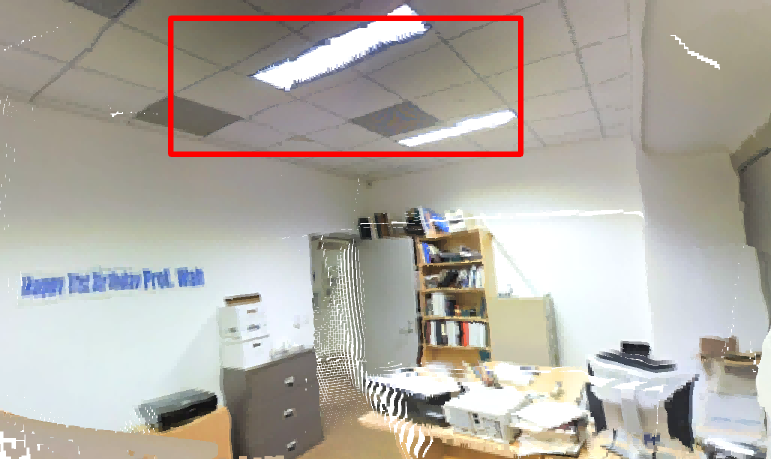}}
	\hspace{0.1em}
	\subfloat[]{\includegraphics[width=.23\linewidth]{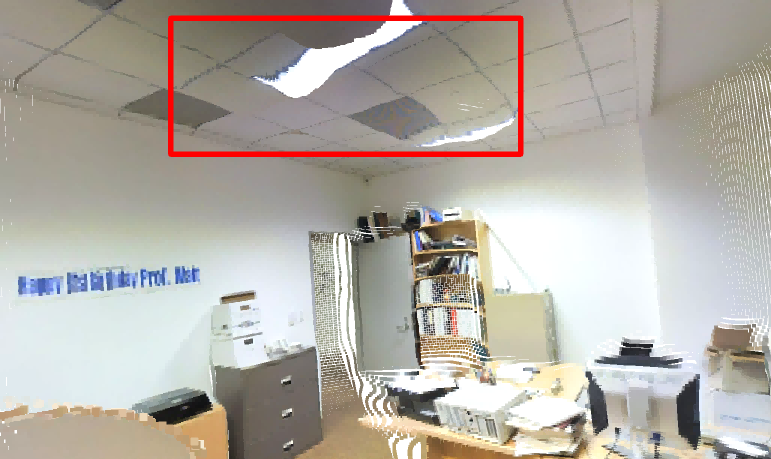}}
	\hspace{0.1em}
	\subfloat[]{\includegraphics[width=.23\linewidth]{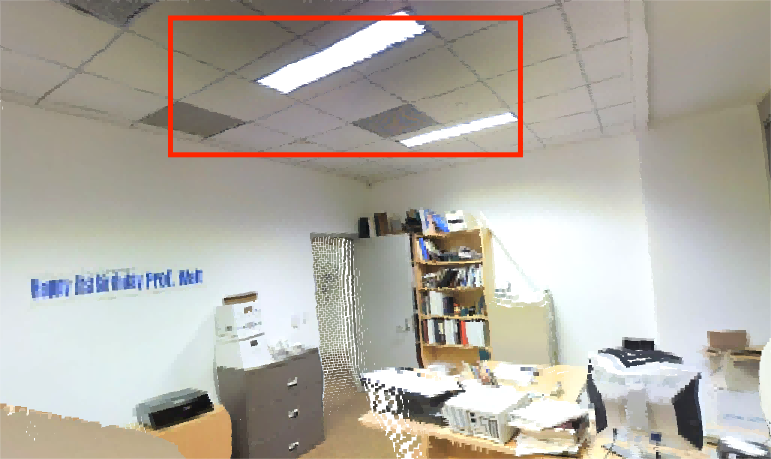}}
	
    \vspace{-1.0em}
	
	\subfloat[GT]{\includegraphics[width=.23\linewidth]{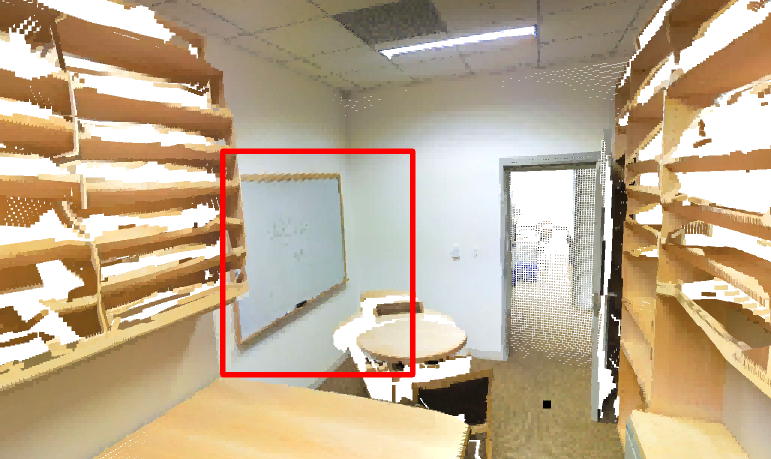}}
	\hspace{0.1em}	
	\subfloat[OmniFusion]{\includegraphics[width=.23\linewidth]{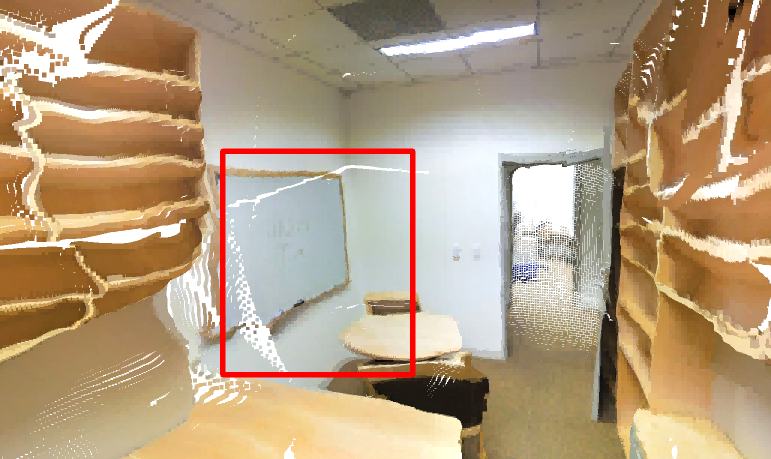}}
	\hspace{0.1em}
	\subfloat[PanoFormer]{\includegraphics[width=.23\linewidth]{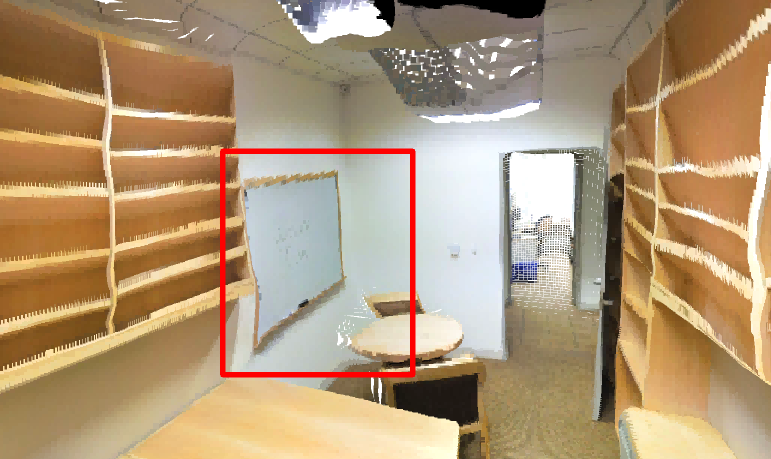}}
	\hspace{0.1em}
	\subfloat[ours]{\includegraphics[width=.23\linewidth]{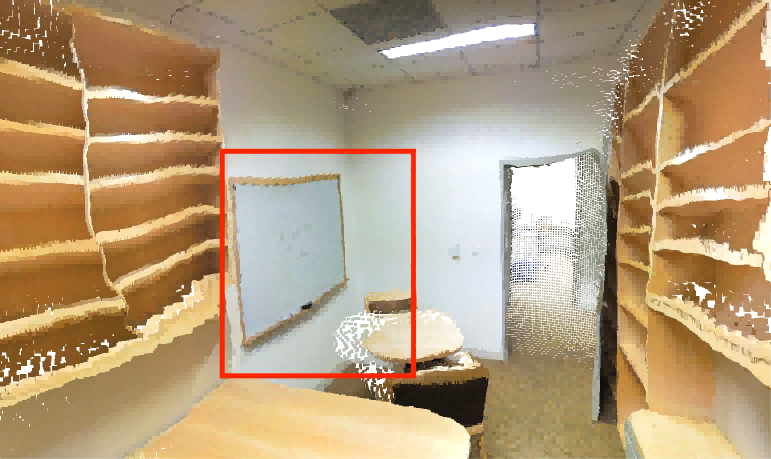}}
	\vspace{-1.0em}
	\caption{
        We select two scenes from Stanford2D3D \cite{armeni2017joint} and show depth maps and corresponding point clouds.
	}
	\label{fig:2d3d_depth_cloud}
    \vspace{-1.5em}
\end{figure}

In addition to the quality of panorama depth maps, we also compare the inference efficiency of different methods. 
We average the inference time by predicting the depth map of 100 panorama images with a resolution of $512 \times 1024$ on a single RTX 3090 to obtain a reliable inference time. 
SphereFusion is the most efficient method, requiring only 0.0174 seconds per image during inference.
Compared with SphereDepth \cite{yan2022spheredepth}, our method achieves higher efficiency by using a lighter mesh network and the cache strategy to store the FAF during inference. 
Although OmniFusion \cite{li2022omnifusion} and PanoFormer \cite{shen2022panoformer} achieve higher depth map quality, they require longer inference time. 

In summary, SphereFusion benefits from choosing the suitable projection and can obtain comparable reconstruction results with more efficient inference efficiency using only a lightweight network model.

\begin{table*} [ht]
	\begin{center}
		\caption{ Ablation studies on network encoder and different fusion strategies. }
		\label{tab:pipeline}
        \vspace{-0.5em}
        
		\addtolength{\tabcolsep}{-0.2pt}
		\resizebox{\linewidth}{!}{\begin{tabular}{ccc|cccc|ccc}
			\hline\noalign{\smallskip}
			2D Encoder & Mesh Encoder & Fusion Module & MRE$\downarrow$ & MAE$\downarrow$ & RMSE$\downarrow$ & RMSE(log)$\downarrow$ & $\delta_1\uparrow$ & $\delta_2\uparrow$ & $\delta_3\uparrow$ \\
			\noalign{\smallskip}
			\hline
			\noalign{\smallskip}
			
			$\checkmark$ & $\times$ & $\times$ & 0.0461 & 0.0969 & 0.2081 & 0.0315 & 0.9833 & 0.9964 & 0.9986 \\
			$\times$ & $\checkmark$ & $\times$  & 0.0572 & 0.1180 & 0.2372 & 0.0374 & 0.9755 & 0.9956 & 0.9985 \\
			$\checkmark$ & $\checkmark$ & BiFuse \cite{wang2020bifuse} & \textbf{0.0415} & \textbf{0.0888} & 0.1824 & 0.0288 & 0.9861 & 0.9969 & 0.9988 \\
			$\checkmark$ & $\checkmark$ & UniFuse \cite{jiang2021unifuse} & 0.0427 & 0.0915 & 0.1837 & 0.0290 & 0.9868 & 0.9969 & 0.9988 \\
			$\checkmark$ & $\checkmark$ & GateFuse (ours)   & 0.0417 & 0.0894 & \textbf{0.1813} & \textbf{0.0286} & \textbf{0.9869} & \textbf{0.9970} & \textbf{0.9989} \\
			
			\hline
		\end{tabular}}
	\end{center}
    \vspace{-2.5em}
\end{table*}

\subsection{Qualitative Evaluation}

For the qualitative evaluation, we visualize depth maps of different methods in Fig. \ref{fig:3d60_depth_cloud}, Fig. \ref{fig:mat3d_depth_cloud}, and Fig. \ref{fig:2d3d_depth_cloud}.
Furthermore, we convert them to point clouds to compare different methods in the 3D space and visualize point clouds by Meshlab \cite{meshlab} with the same rendering settings.

On 360D \cite{zioulis2018omnidepth}, we compare our method with SphereDepth \cite{yan2022spheredepth} and UniFuse \cite{jiang2021unifuse}, and visualize depth maps and corresponding point clouds in Fig. \ref{fig:3d60_depth_cloud}.
UniFuse has better reconstruction results in the middle areas but struggles around the polar regions, such as the lights on the ceiling. SphereDepth reconstructs the ceiling region but suffers from losing details, such as edges of the door and the wall. SphereFusion combines the strengths of two projections and can reconstruct details and polar regions at the same time.

On Matterport3D \cite{chang2017matterport3d}, we compare our method with SliceNet \cite{pintore2021slicenet} and PanoFormer \cite{shen2022panoformer}, and Fig. \ref{fig:mat3d_depth_cloud} shows  results.
SliceNet suffers from poles, as it only uses the equirectangular projection and extracts features by the 2D image encoder.
PanoFormer and SphereFusion achieve better results using the tangent and spherical projections.

On Stanford2D3D \cite{armeni2017joint}, we compare our method with OmniFusion \cite{yan2022spheredepth} and PanoFormer \cite{jiang2021unifuse}.
Fig. \ref{fig:2d3d_depth_cloud}
shows depth maps and corresponding point clouds.  
OmniFusion and PanoFormer use the tangent projection and attempt to reduce discontinuities using more complex feature fusion mechanisms. 
However, OmniFusion fails to merge different tangent patches and has noticeable gaps in the point cloud, while PanoFormer is smoother and loses some details. 
Although OmniFusion and PanoFormer achieve better reconstruction results than SphereFusion, our method only uses a simple encoder based on ResNet, demonstrating the importance of choosing the proper projection.

Overall, SphereFusion utilizes the spherical projection to avoid distortion and discontinuities and the equirectangular projection to extract visual features, achieving comparable results with state-of-the-art methods with a lighter network and higher inference efficiency.

\subsection{Ablation Studies}

We conduct several ablation studies to study the influence of different components of the SphereFusion. We first compare the network encoder to show the importance of using two encoders to extract features from the panorama image. We then study how to fuse features from the spherical and equirectangular projection. Throughout all ablation experiments, we use the 360D dataset.

\subsubsection{Network Encoder}
To evaluate the contribution of each encoder, we build two networks to estimate panorama depth, where each network only uses one type of encoder.
Table \ref{tab:pipeline} shows the performance of different network structures. 
The network that only uses the mesh encoder obtains the worst results, which cannot reconstruct details of the panorama image only through the mesh operation, as SphereDepth \cite{yan2022spheredepth} does.
The network that only uses the 2D image encoder ranks second in Table \ref{tab:pipeline}, which can achieve higher performance but cannot deal with distortion and discontinuity.
SphereFusion outperforms others and achieves the best results, proving that combining the 2D image encoder and the mesh encoder can obtain higher-quality depth maps.

\subsubsection{Fusion Strategy}
\begin{figure} [h]
	\centering
	\captionsetup[subfigure]{labelformat=empty}
	
	\subfloat[]{\includegraphics[width=.18\linewidth]{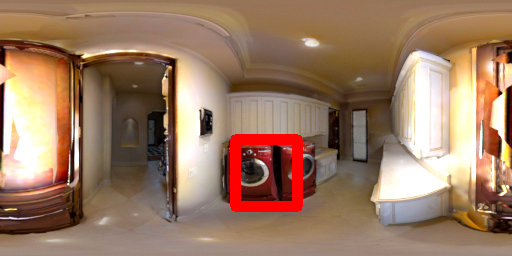}}
	\hspace{0.1em}
	\subfloat[]{\includegraphics[width=.18\linewidth]{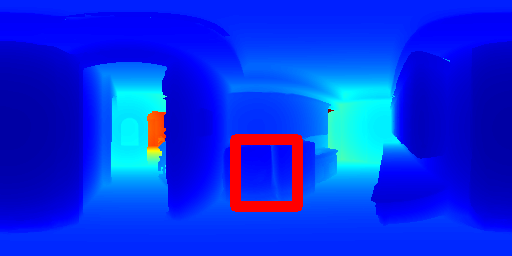}}
	\hspace{0.1em}
	\subfloat[]{\includegraphics[width=.18\linewidth]{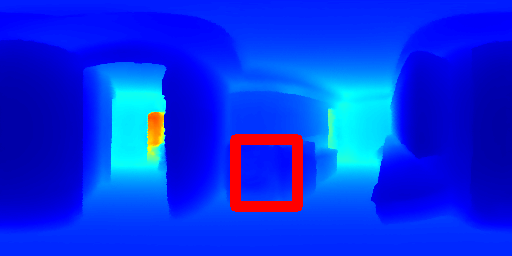}}
	\hspace{0.1em}
	\subfloat[]{\includegraphics[width=.18\linewidth]{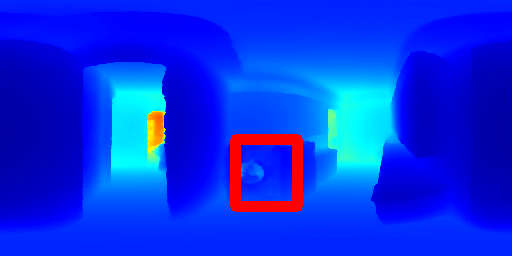}}
	\hspace{0.1em}
	\subfloat[]{\includegraphics[width=.18\linewidth]{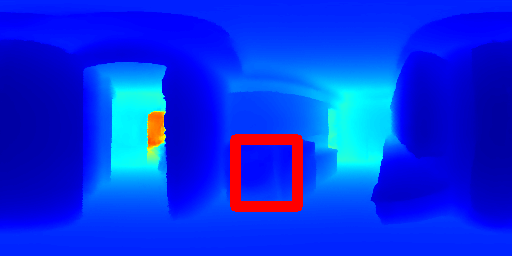}}
	\hspace{0.1em}
 
	\vspace{-1.0em}

	\subfloat[RGB]{\includegraphics[width=.18\linewidth]{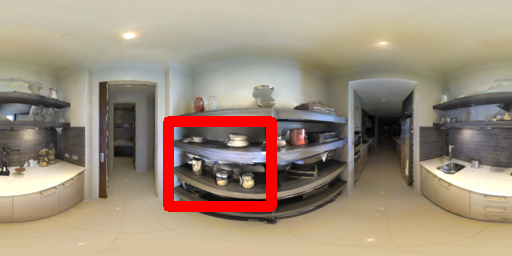}}
	\hspace{0.1em}
	\subfloat[GT]{\includegraphics[width=.18\linewidth]{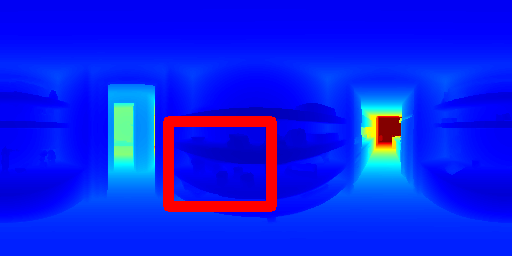}}
	\hspace{0.1em}
	\subfloat[BiFuse]{\includegraphics[width=.18\linewidth]{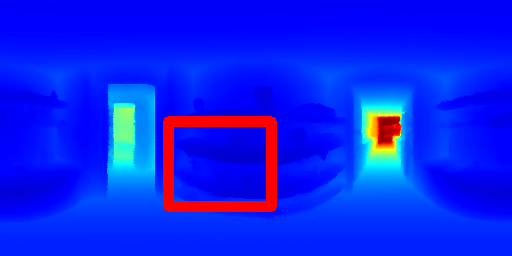}}
	\hspace{0.1em}
	\subfloat[UniFuse]{\includegraphics[width=.18\linewidth]{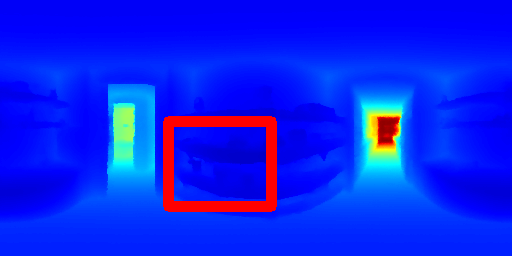}}
	\hspace{0.1em}
	\subfloat[ours]{\includegraphics[width=.18\linewidth]{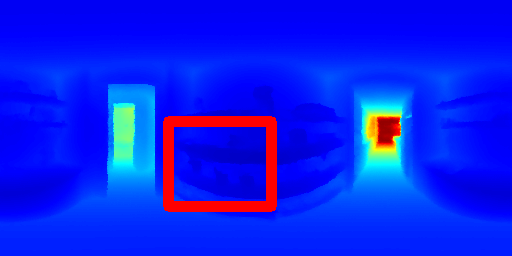}}
	\hspace{0.1em}

    \vspace{-1.0em}

	\caption{
		We select two scenes from 360D \cite{zioulis2018omnidepth} and compare depth maps from different fusion strategies.
		Our GateFuse can reconstruct more details, and we mark out these regions with red boxes.
	}
	\label{fig:viz_fusion}

    \vspace{-1.5em}
 
\end{figure}

The fusion strategy fuses features from different panorama projections.
To compare different fusion strategies, we implement BiFuse \cite{wang2020bifuse}, UniFuse \cite{jiang2021unifuse}, and our GateFuse to fuse features in the spherical projection.
Table \ref{tab:pipeline} shows the results of different fusion strategies.
UniFuse obtains the worst results, and BiFuse ranks second.
Our GateFuse achieves the best results on RMSE, RMSE(log), $\delta$, and ranks second on MAE and MRE.
We visualize depth maps from different fusion strategies for better comparison in Fig. \ref{fig:viz_fusion}, where BiFuse and UniFuse fail to reconstruct details.

\subsection{Limitations}

We propose SphereFuion for panorama depth estimation by using the 2D image convolution and the mesh convolution, which achieves competitive results with a lighter network and the highest inference efficiency.
However, the cache strategy requires additional GPU memory to store FAF information during training and testing.
Furthermore, SphereFusion requires huge GPU memory during training and does not support ultra-high resolution panorama images, such as $1024 \times 2048$ \cite{rey2022360monodepth}. 
We still need more in-depth research to improve the mesh operation and the quality of panorama depth estimation.

\section{Conclusion}
\label{conclusion}

This paper introduces a novel panorama depth estimation method, SphereFusion, which combines the strengths of equirectangular and spherical projection. 
SphereFusion uses 2D image convolution to complement mesh convolution by using a gate fusion module to select reliable features from two encoders and estimates the panorama depth map in the spherical domain to avoid distortion and discontinuity. 
Meanwhile, SphereFusion utilizes a lighter network and a cache strategy to improve the inference efficiency. 
Experiments conducted on three popular datasets indicate that SphereFusion achieves competitive results and maintains an impressive inference efficiency of up to 60 FPS for $512 \times 1024$ panorama images on one NVIDIA RTX3090.

\section{Acknowledgments}
This work was supported by the Shenzhen Science and Technology Program (Grant No. KJZD20240903 104103005, JCYJ20220818102414030, JSGGKQTD202 21101115655027).

{
    \small
    \bibliographystyle{ieeenat_fullname}
    \bibliography{main}
}
\clearpage
\setcounter{page}{1}
\maketitlesupplementary

\section{Panorama Projections}

To capture the texture features and avoid distortion and discontinuity, SphereFusion relies on two panorama projections: the equirectangular projection and the spherical mesh. In this section, we describe the details of their conversion, and how to convert the panorama image in spherical mesh to equirectangular for depth map evaluation orre-project it to 3D space for visualization.

\paragraph{The E2S}

Based on the definition of the equirectangular projection and the spherical mesh, we define the E2S module, which converts a panorama image from the equirectangular projection to the spherical projection. Given a triangle center $(x, y, z)$ from the spherical mesh, we first calculate its position on the image plane $(u, v)$ by Eq. \ref{eq:e2s}, and then use a bi-linear sample to capture value on the image plane, finally assign corresponding value to the triangle.

\begin{align}
	\label{eq:e2s}
	\begin{cases}
        u = (1+atan(y,x)/\pi) \times W/2  \\
        v = (0.5+atan(z,\sqrt{x^2+y^2})/\pi ) \times H
	\end{cases}
\end{align}

\paragraph{The S2E}

Meanwhile, we can also define the S2E module, which converts a panorama image from the spherical projection to the equirectangular projection. Given a pixel $(u, v)$ on the image plane, we first calculate its position on the sphere surface, then calculate its 3D position to find out the closest triangle on the spherical mesh, and finally assign the value of the triangle to the pixel.

\paragraph{Projection To 3D Point Cloud}

For better comparison, we visualize point clouds using different methods. Although we name the result obtained from the panorama depth estimation as the depth map, it represents the distance map, which means the Euclidean distance between a 3D point and the camera center. Therefore, we can reproject a panorama depth map to 3D space by Eq. \ref{eq:to_point} after converting it into the sphere coordinates, where $d$ is the distance.

\begin{align}
	\label{eq:to_point}
	\begin{cases}
		X = cos(latitude)cos(longitude) \times d\\
		Y = cos(latitude)sin(longitude) \times d\\ 
		Z = sin(latitude) \times d\\
	\end{cases}
\end{align}

\section{Evaluation}

To compare with other methods, we convert our depth map in the spherical domain to equirectangular projection. Following BiFuse \cite{wang2020bifuse}, we use five evaluation metrics, including MAE, MRE, RMSE, RMSE(log), and $\delta ^n$. Eq. \ref{eq:metrics} shows how to calculate them, where $gt$ is the ground truth, $pr$ is the predicted depth, $V$ is valid pixels, and $N$ is the number of valid pixels. For MAE, MRE, RMSE, and RMSE(log), the smaller is better. For the $\delta$, the bigger is better. During the evaluation, we set the depth range to $0.1 \sim 10$ meters.

\begin{equation} 
\label{eq:metrics}
\begin{split}
MAE &= \sum_{i\in V}{|gt_i-pr_i|}   \\
MRE &= \sum_{i\in V}{\frac{|gt_i-pr_i|}{gt_i}} \\
RMSE &= \sqrt{\frac{\sum_{i\in V}(gt_i-pr_i)^2}{N}} \\
RMSE_{log} & = \sqrt{\frac{\sum_{i\in V}(log_{10}(gt_i)-log_{10}(pr_i))^2}{N}} \\
\delta^n & = \frac{\sum_{i\in V}max(\frac{gt_i}{pr_i},\frac{pr_i}{gt_i})<1.25^n}{N}
\end{split}
\end{equation}

\section{Visulization}

\begin{figure*}[t]
	\centering
	\captionsetup[subfigure]{labelformat=empty}
	
	\begin{subfigure}{0.18\linewidth}
		\includegraphics[width=.98\linewidth]{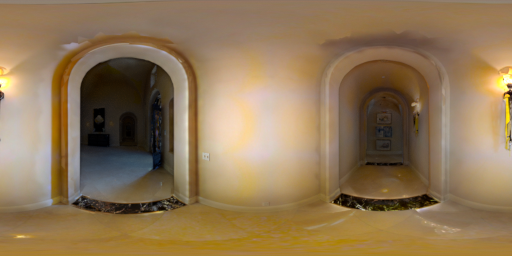}
	\end{subfigure}
	\begin{subfigure}{0.18\linewidth}
		\includegraphics[width=.98\linewidth]{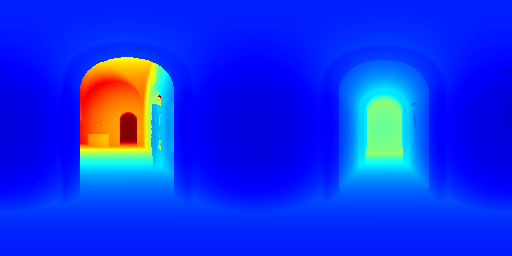}
	\end{subfigure}
	\begin{subfigure}{0.18\linewidth}
		\includegraphics[width=.98\linewidth]{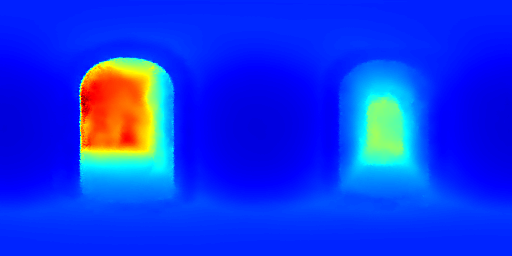}
	\end{subfigure}
	\begin{subfigure}{0.18\linewidth}
		\includegraphics[width=.98\linewidth]{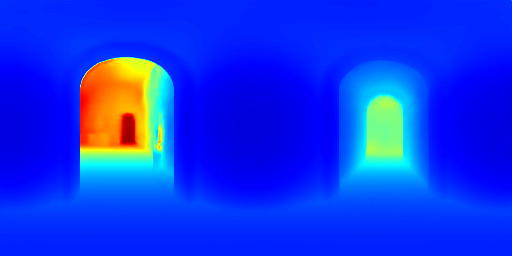}
	\end{subfigure}
	\begin{subfigure}{0.18\linewidth}
		\includegraphics[width=.98\linewidth]{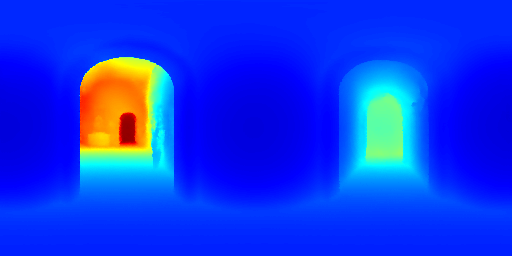}
	\end{subfigure}

	\vspace{1pt}

	\begin{subfigure}{0.18\linewidth}
		\includegraphics[width=.98\linewidth]{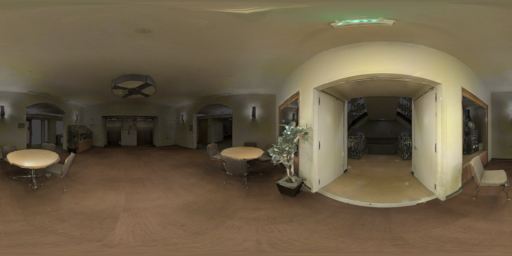}
	\end{subfigure}
	\begin{subfigure}{0.18\linewidth}
		\includegraphics[width=.98\linewidth]{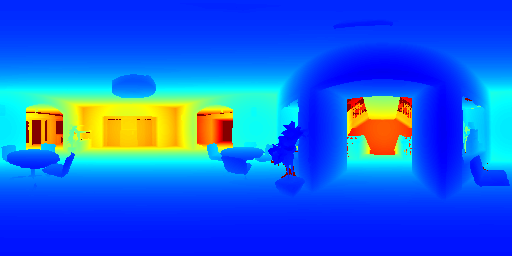}
	\end{subfigure}
	\begin{subfigure}{0.18\linewidth}
		\includegraphics[width=.98\linewidth]{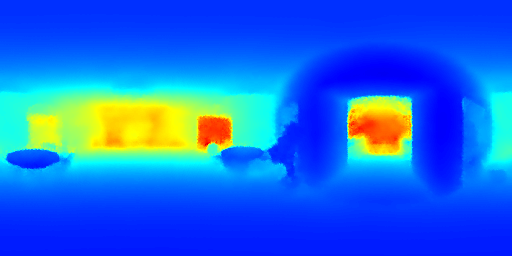}
	\end{subfigure}
	\begin{subfigure}{0.18\linewidth}
		\includegraphics[width=.98\linewidth]{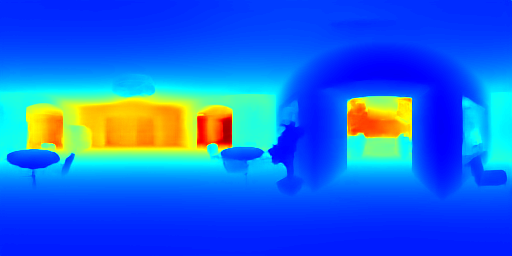}
	\end{subfigure}
	\begin{subfigure}{0.18\linewidth}
		\includegraphics[width=.98\linewidth]{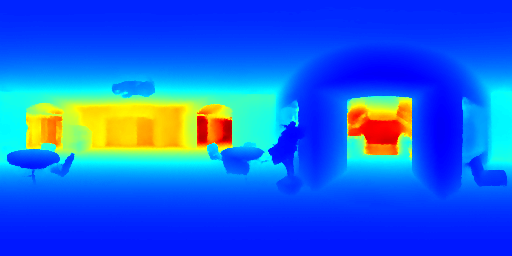}
	\end{subfigure}
	
	\vspace{1pt}

	\begin{subfigure}{0.18\linewidth}
		\includegraphics[width=.98\linewidth]{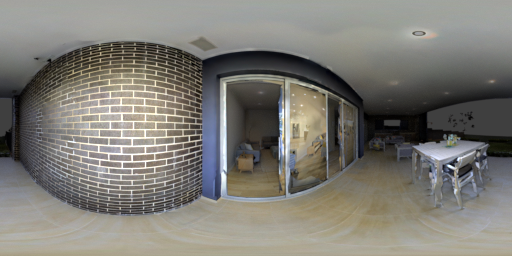}
	\end{subfigure}
	\begin{subfigure}{0.18\linewidth}
		\includegraphics[width=.98\linewidth]{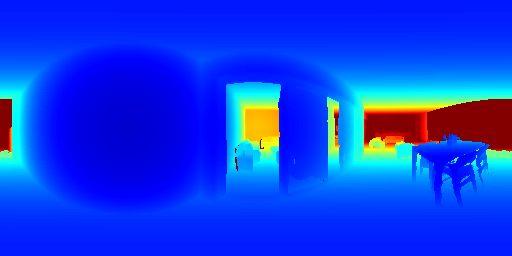}
	\end{subfigure}
	\begin{subfigure}{0.18\linewidth}
		\includegraphics[width=.98\linewidth]{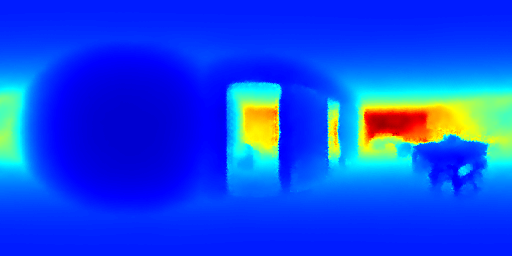}
	\end{subfigure}
	\begin{subfigure}{0.18\linewidth}
		\includegraphics[width=.98\linewidth]{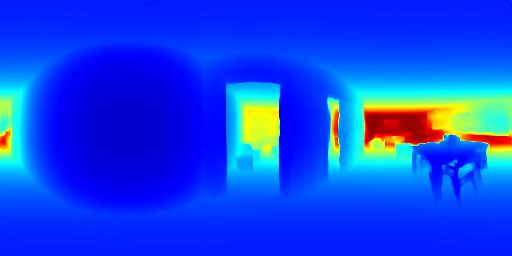}
	\end{subfigure}
	\begin{subfigure}{0.18\linewidth}
		\includegraphics[width=.98\linewidth]{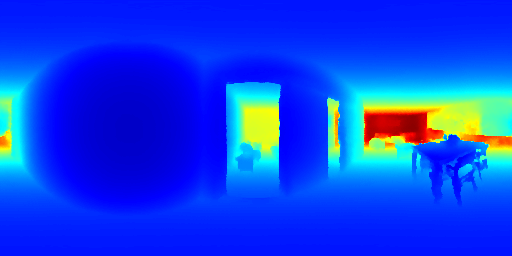}
	\end{subfigure}
	
	\vspace{1pt}

	\begin{subfigure}{0.18\linewidth}
		\includegraphics[width=.98\linewidth]{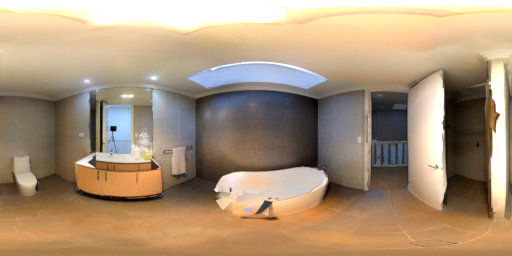}
		\caption{RGB}
	\end{subfigure}
	\begin{subfigure}{0.18\linewidth}
		\includegraphics[width=.98\linewidth]{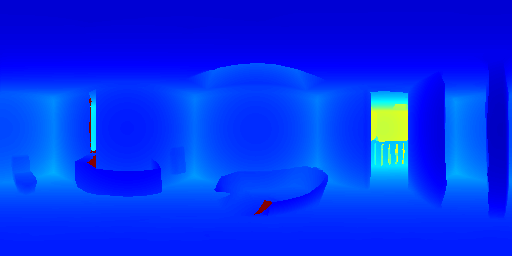}
		\caption{GT}
	\end{subfigure}
	\begin{subfigure}{0.18\linewidth}
		\includegraphics[width=.98\linewidth]{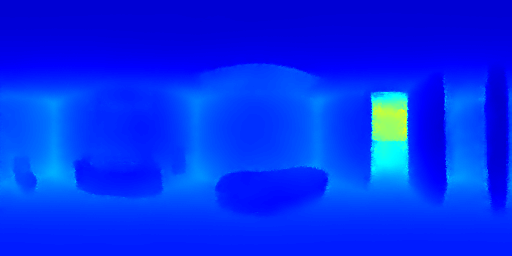}
		\caption{SphereDepth \cite{yan2022spheredepth}}
	\end{subfigure}
	\begin{subfigure}{0.18\linewidth}
		\includegraphics[width=.98\linewidth]{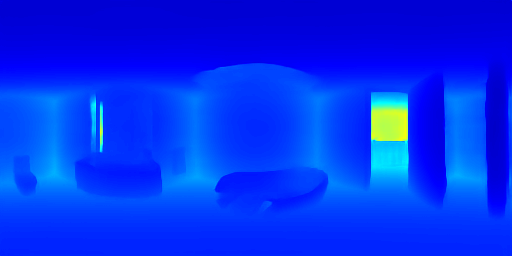}
		\caption{UniFuse \cite{jiang2021unifuse}}
	\end{subfigure}
	\begin{subfigure}{0.18\linewidth}
		\includegraphics[width=.98\linewidth]{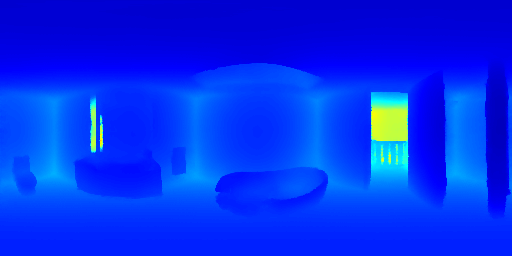}
		\caption{SphereFusion (ours)}
	\end{subfigure}

	\caption{
		\textbf{Depth Maps of 360D.} Invalid parts of the depth map are set to red. 
	}
	\label{fig:3d60_depth}
    \vspace{-1.0em}
\end{figure*}

\begin{figure*}[t]
	\centering
	\captionsetup[subfigure]{labelformat=empty}

	\begin{subfigure}{0.3\linewidth}
		\includegraphics[width=.98\linewidth]{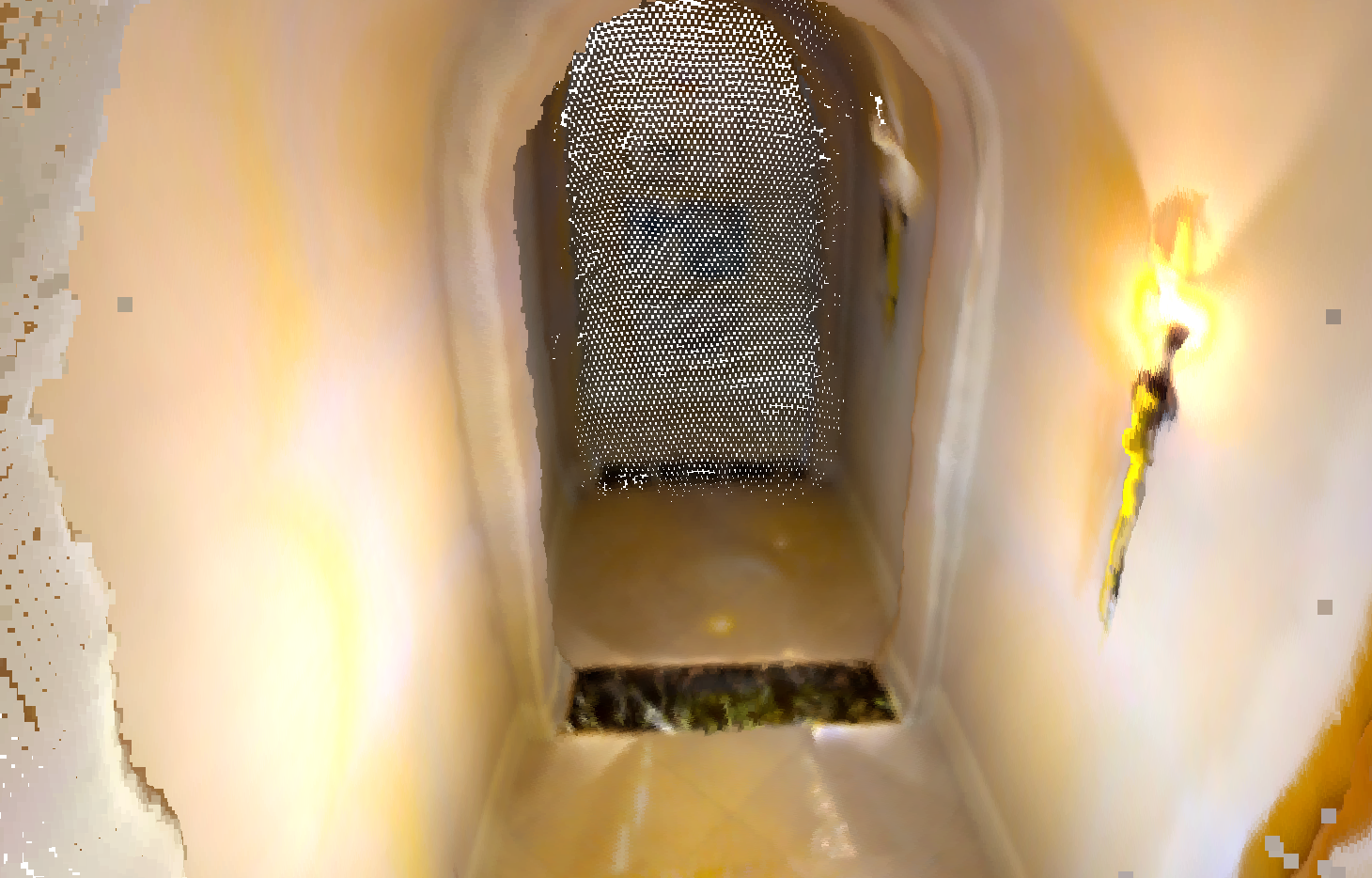}
	\end{subfigure}
	\begin{subfigure}{0.3\linewidth}
		\includegraphics[width=.98\linewidth]{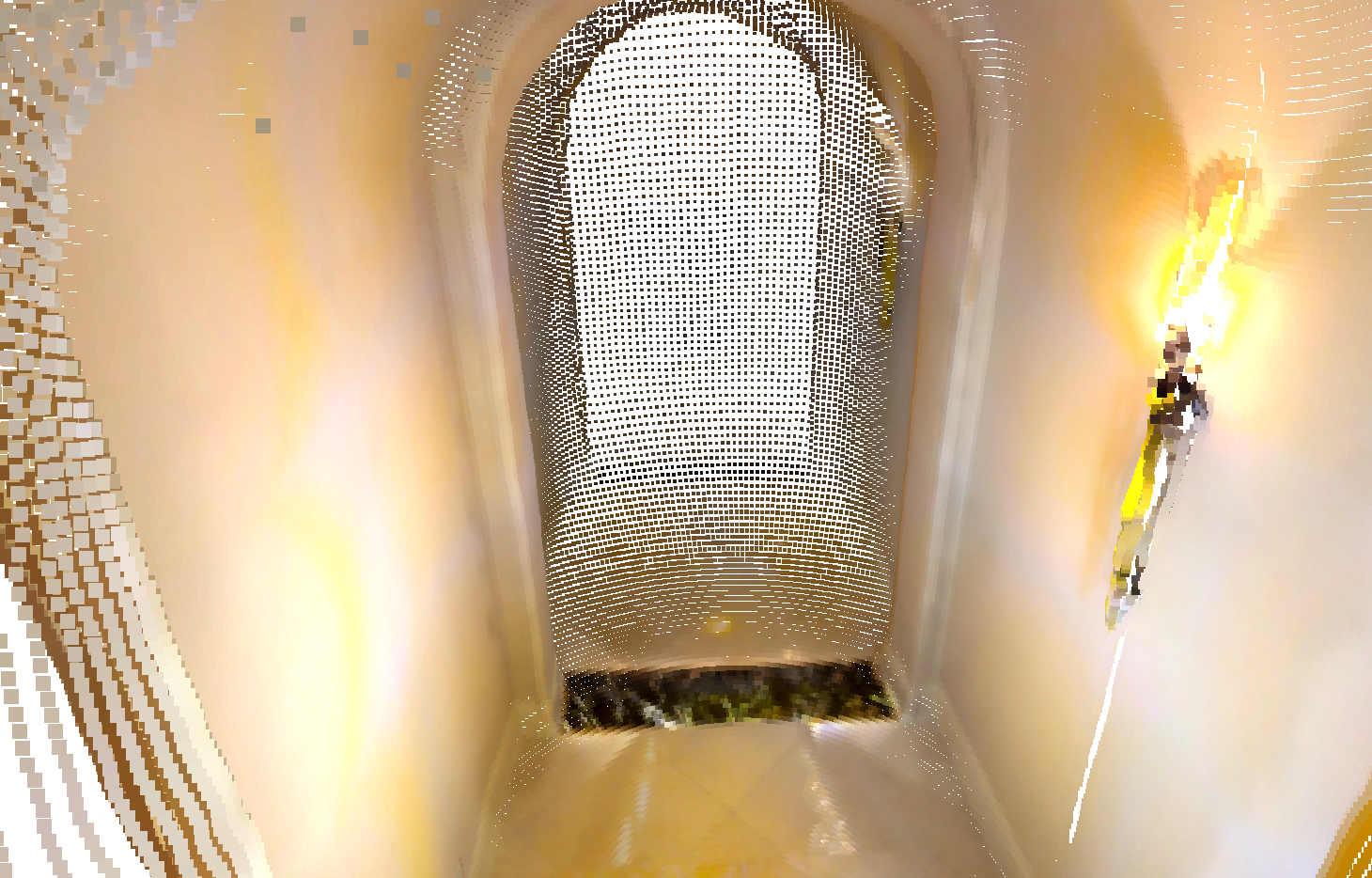}
	\end{subfigure}
	\begin{subfigure}{0.3\linewidth}
		\includegraphics[width=.98\linewidth]{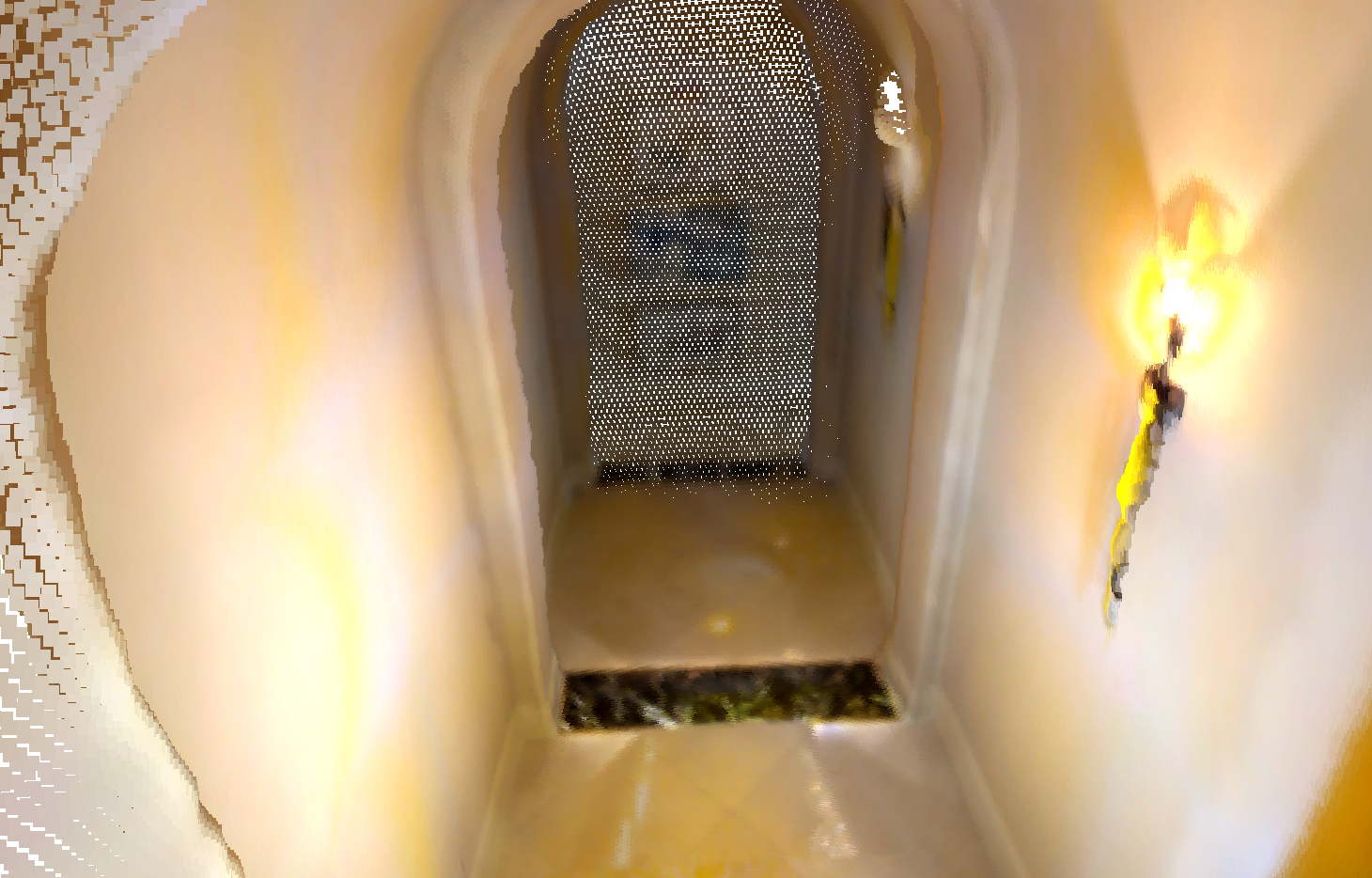}
	\end{subfigure}
	
	\vspace{1pt}
	
	\begin{subfigure}{0.3\linewidth}
		\includegraphics[width=.98\linewidth]{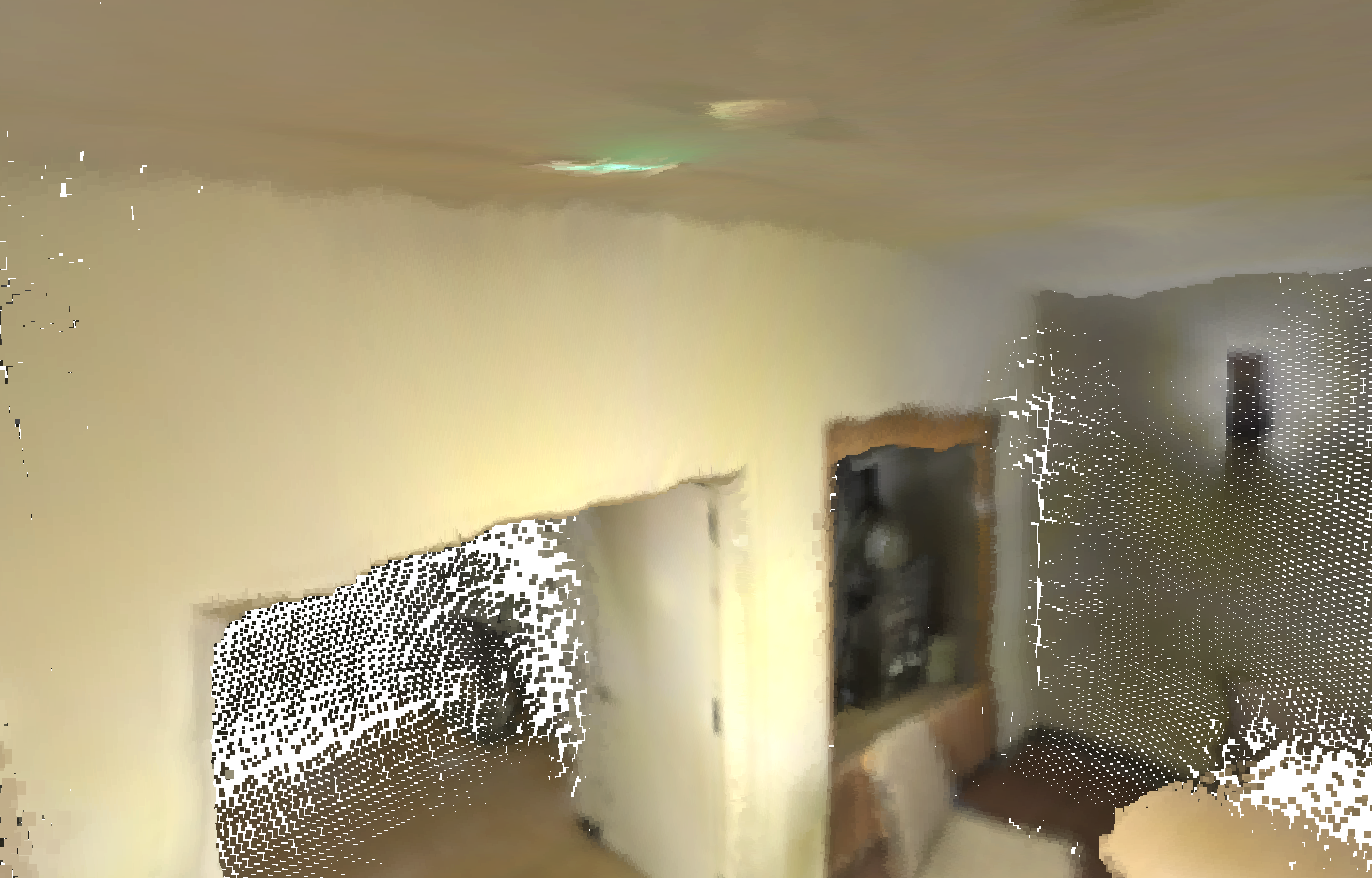}
	\end{subfigure}
	\begin{subfigure}{0.3\linewidth}
		\includegraphics[width=.98\linewidth]{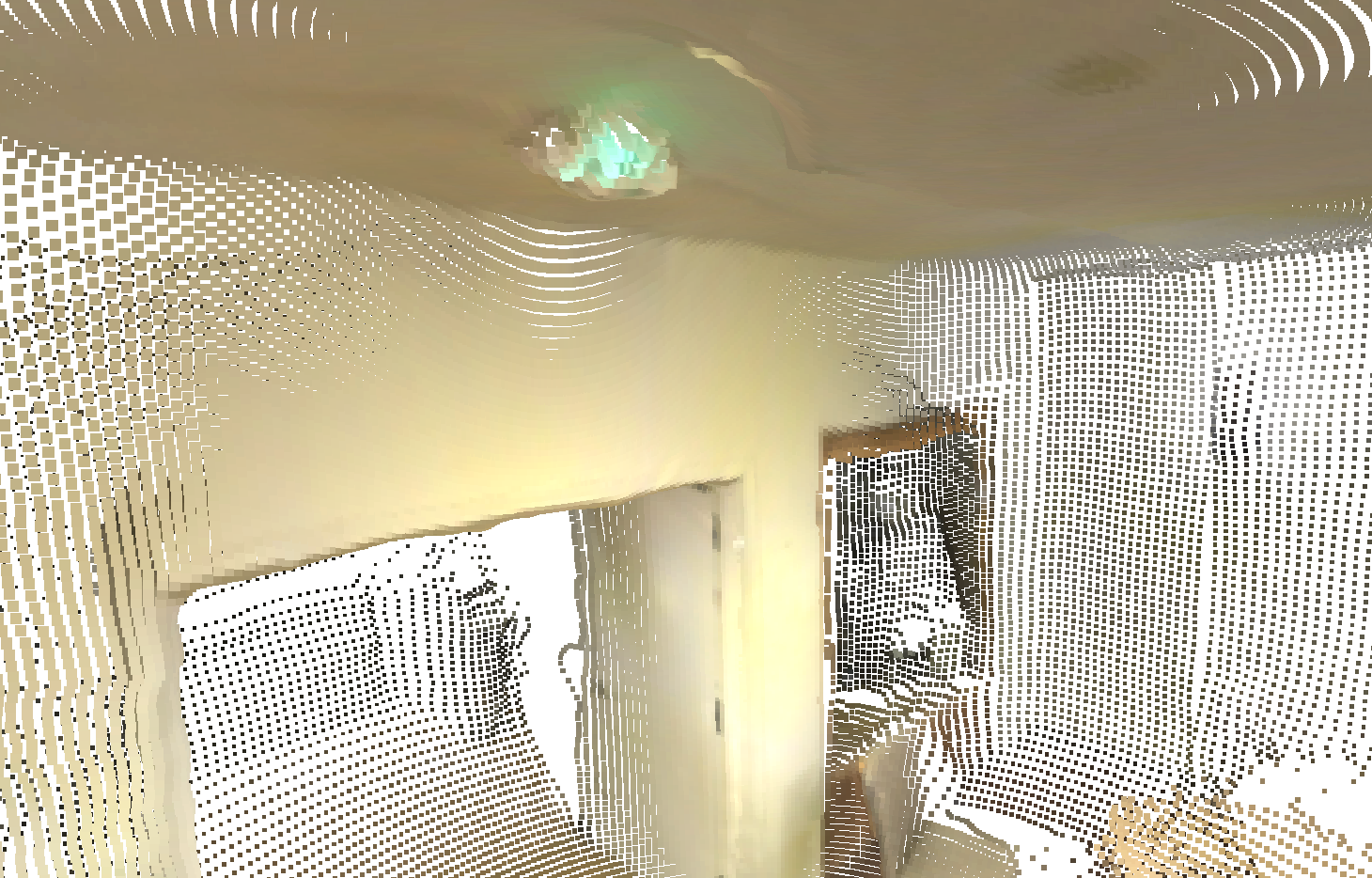}
	\end{subfigure}
	\begin{subfigure}{0.3\linewidth}
		\includegraphics[width=.98\linewidth]{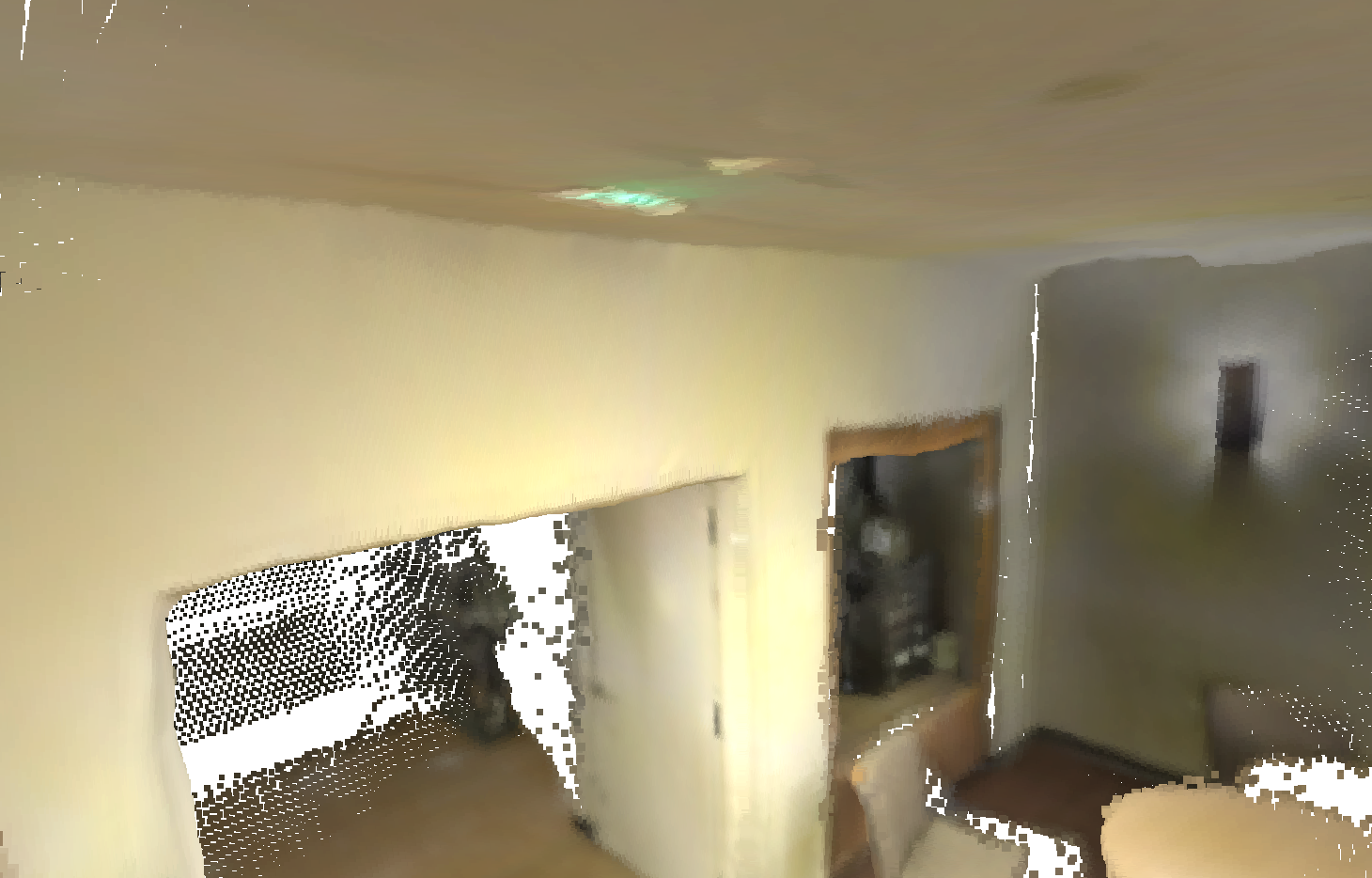}
	\end{subfigure}
	
	\vspace{1pt}

	\begin{subfigure}{0.3\linewidth}
		\includegraphics[width=.98\linewidth]{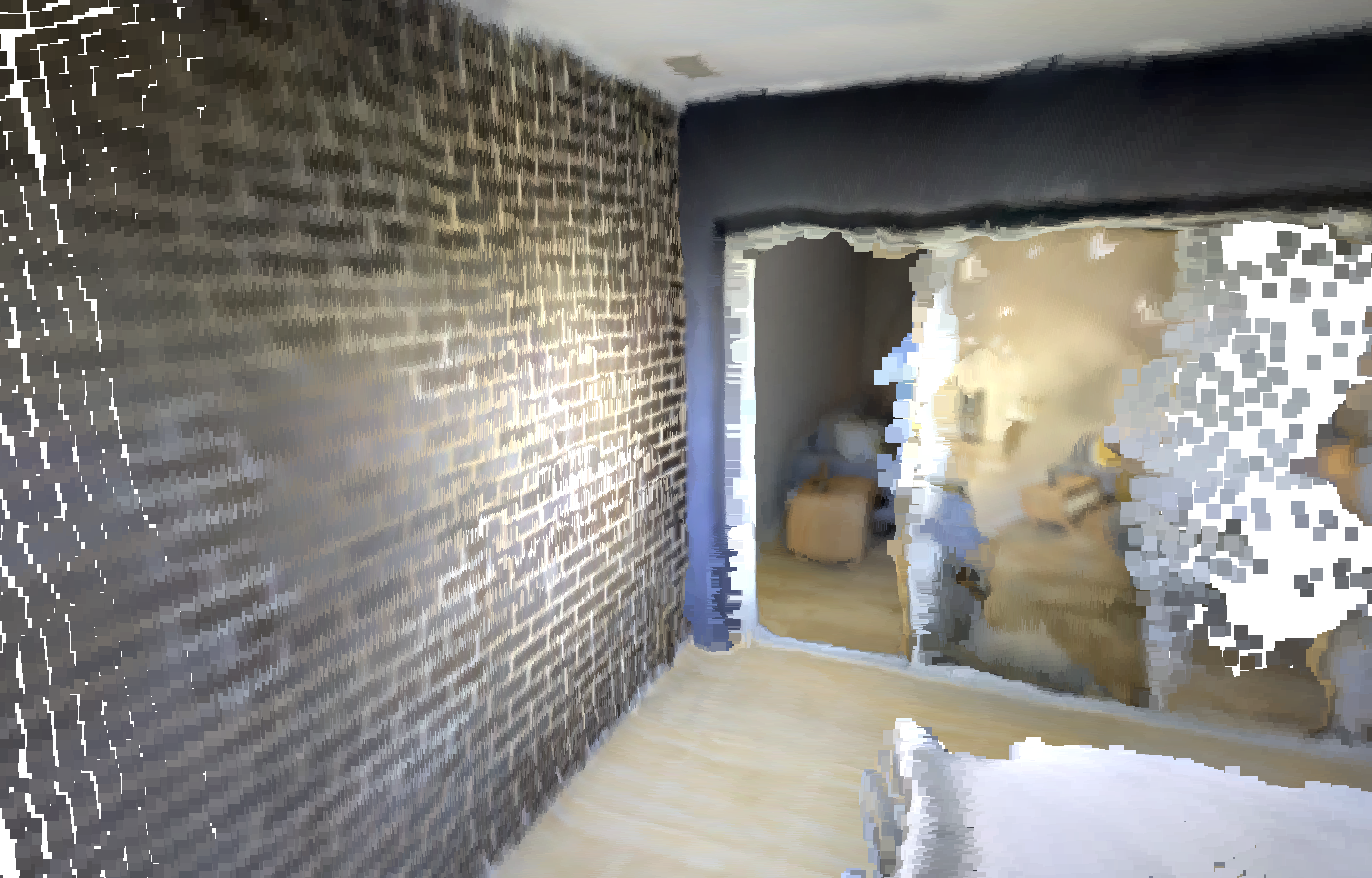}
	\end{subfigure}
	\begin{subfigure}{0.3\linewidth}
		\includegraphics[width=.98\linewidth]{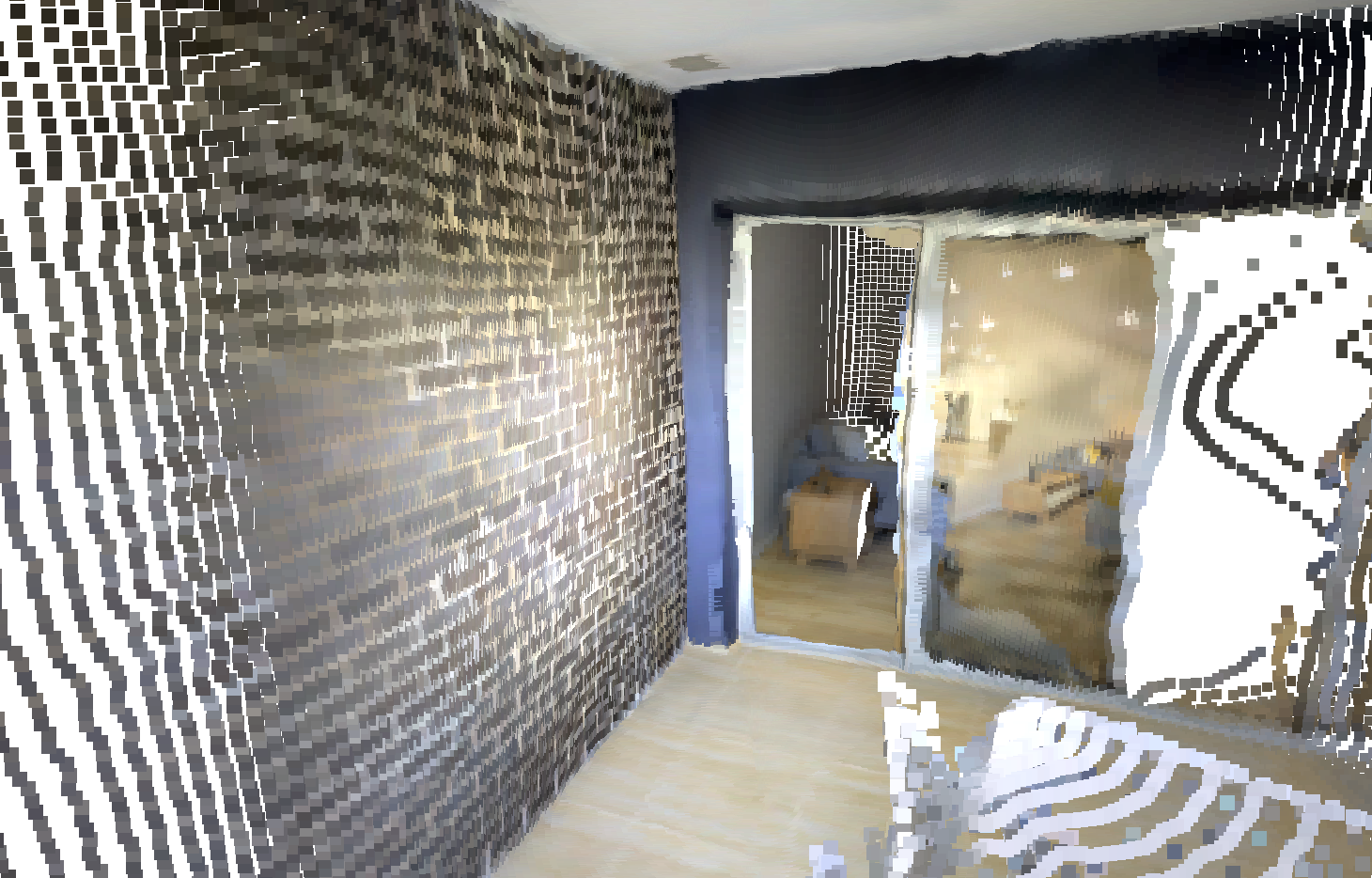}
	\end{subfigure}
	\begin{subfigure}{0.3\linewidth}
		\includegraphics[width=.98\linewidth]{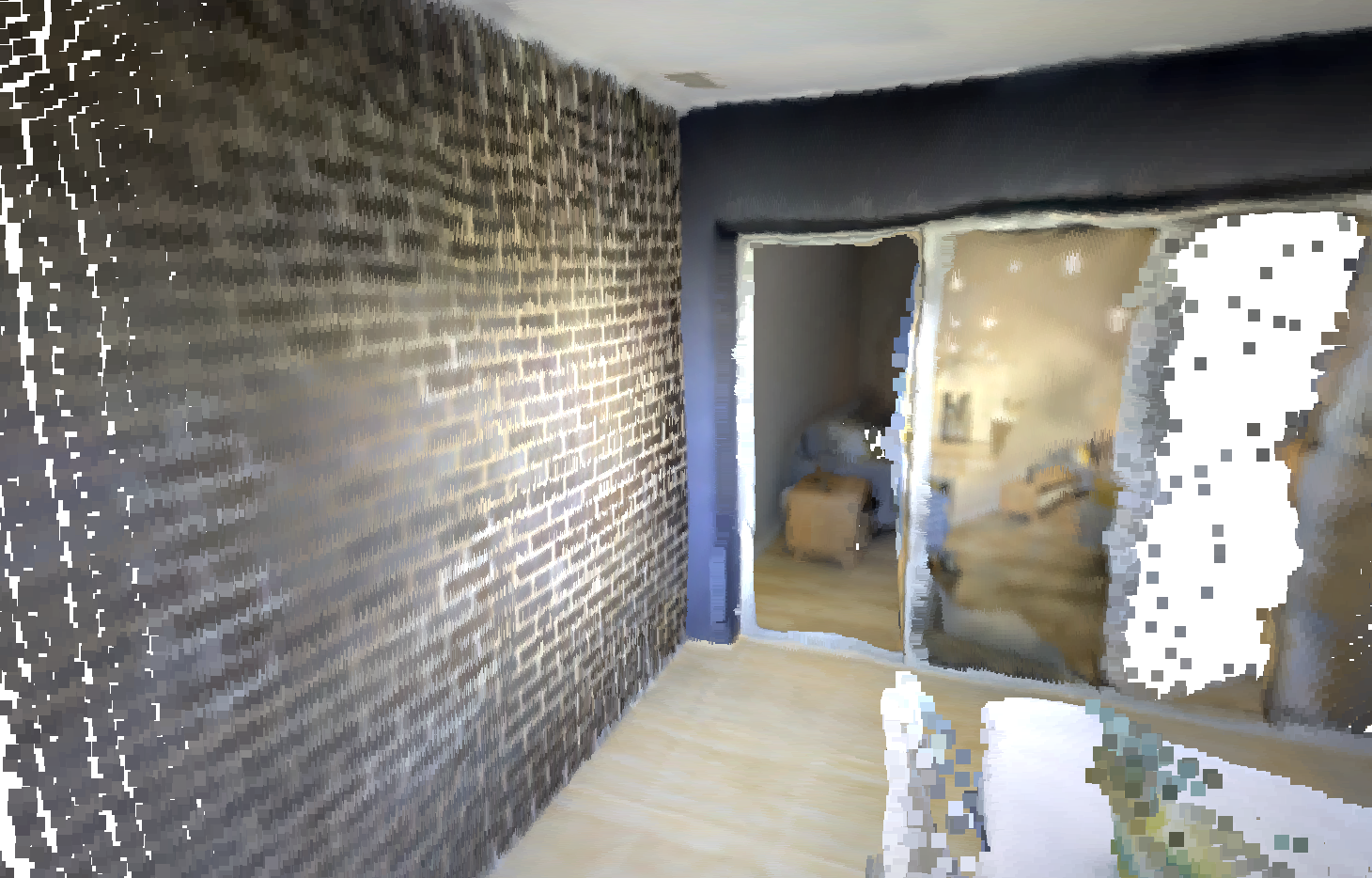}
	\end{subfigure}
	
	\vspace{1pt}

	\begin{subfigure}{0.3\linewidth}
		\includegraphics[width=.98\linewidth]{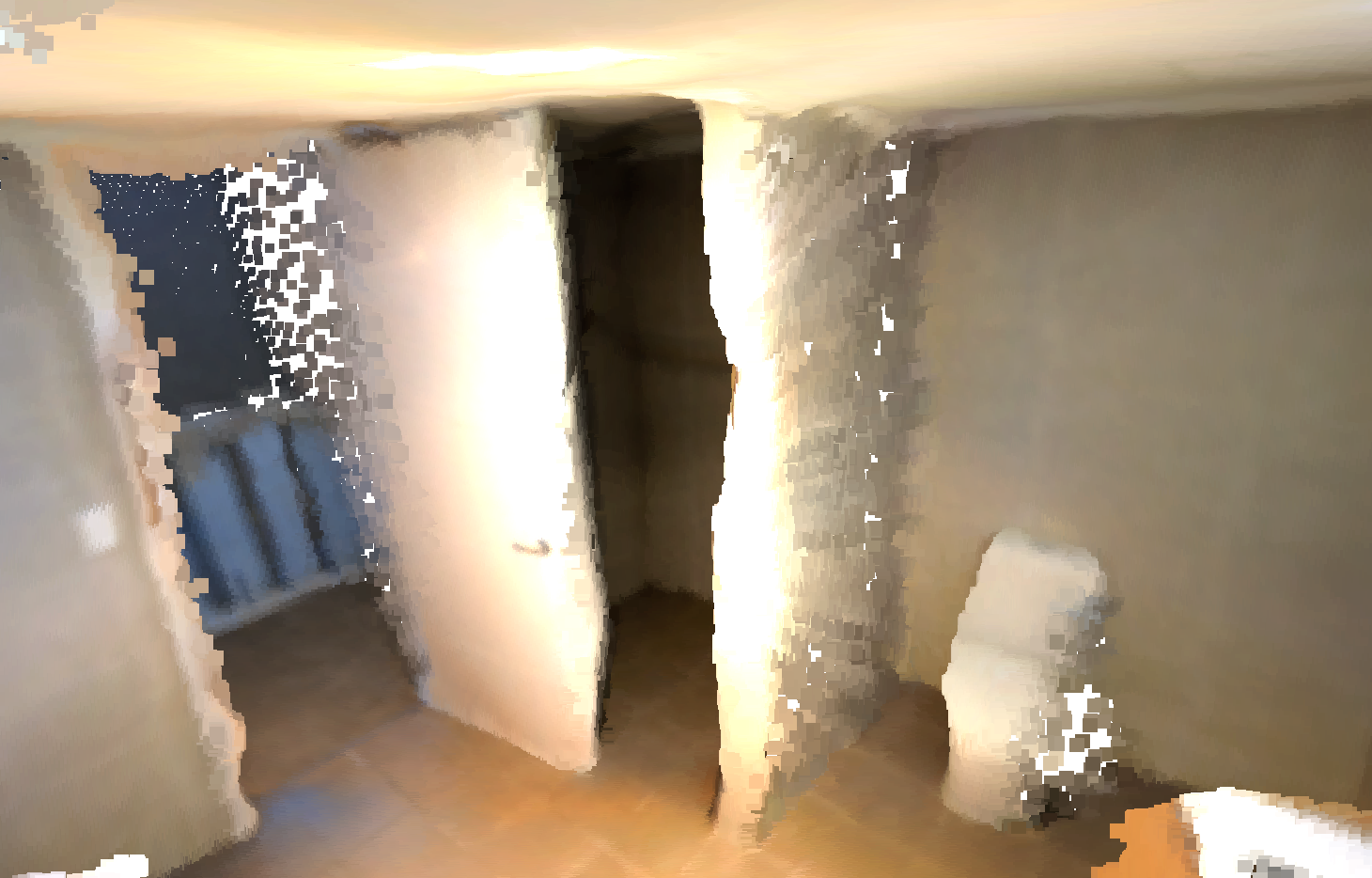}
		\caption{SphereDepth \cite{yan2022spheredepth}}
	\end{subfigure}
	\begin{subfigure}{0.3\linewidth}
		\includegraphics[width=.98\linewidth]{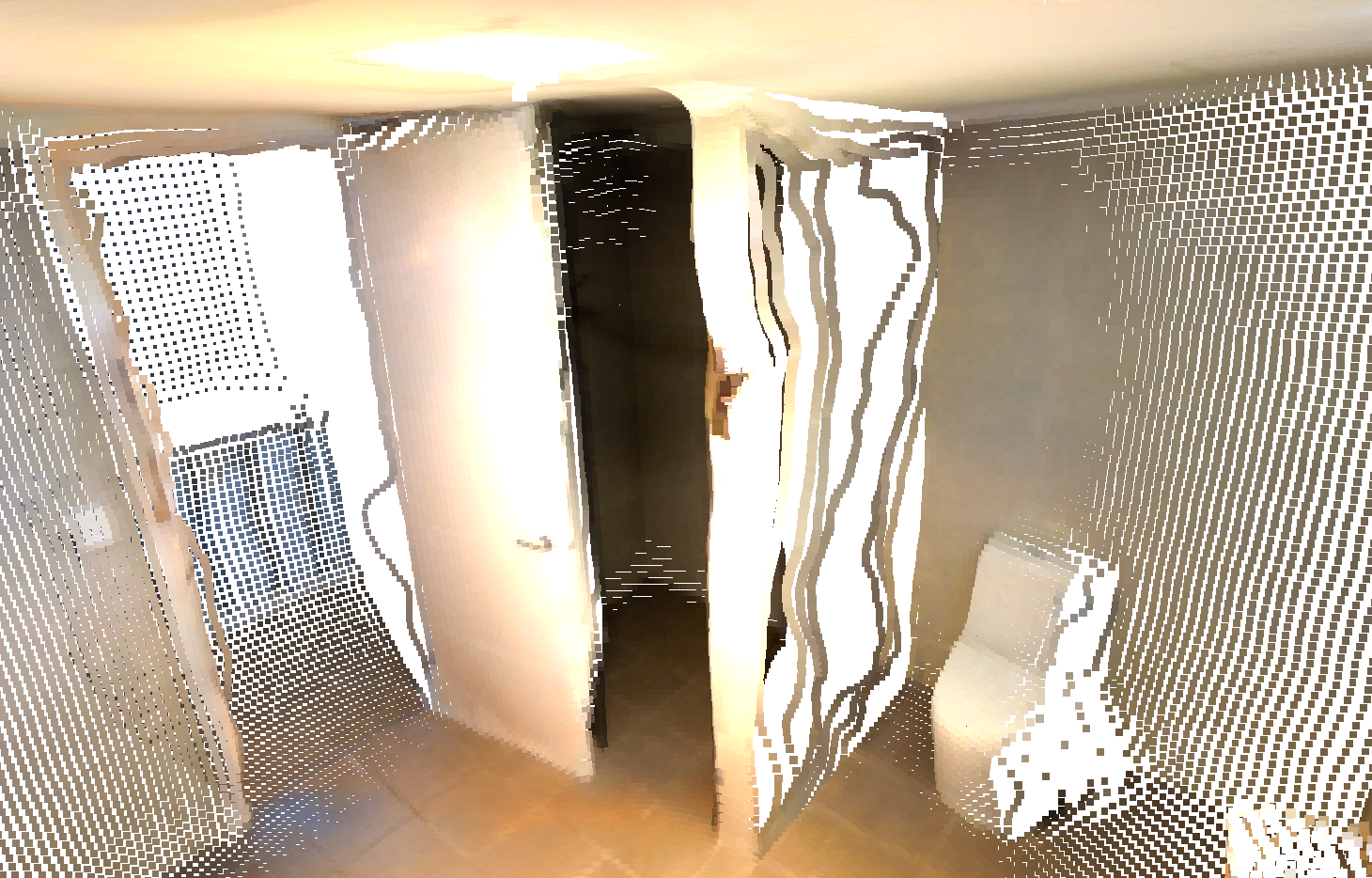}
		\caption{UniFuse \cite{jiang2021unifuse}}
	\end{subfigure}
	\begin{subfigure}{0.3\linewidth}
		\includegraphics[width=.98\linewidth]{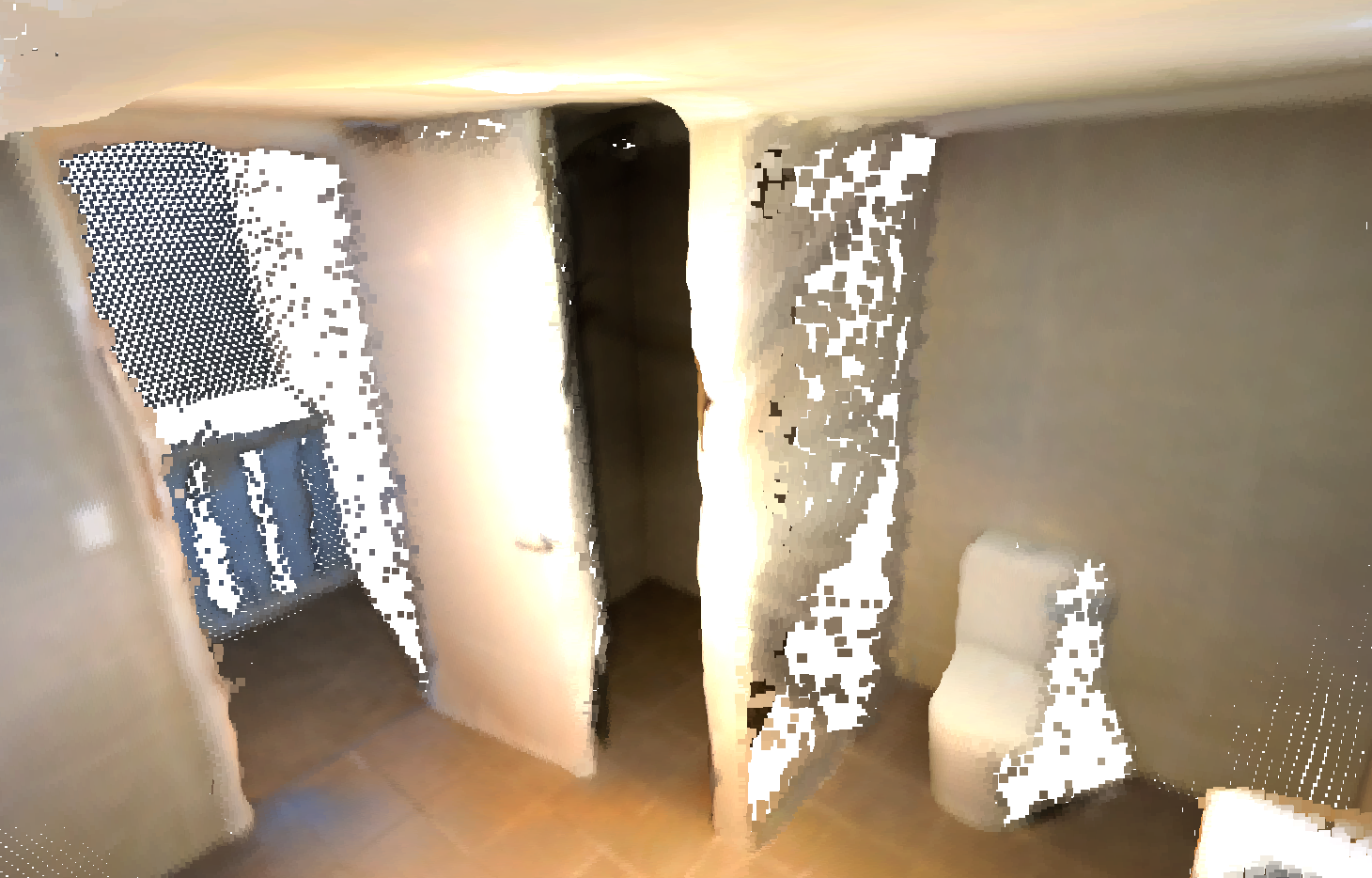}
		\caption{SphereFusion (ours)}
	\end{subfigure}

	\caption{
		\textbf{Point Clouds of 360D.} Our method reconstructs more details of the scene.
	}
	\label{fig:3d60_cloud}
    \vspace{-1.0em}
\end{figure*}

\begin{figure*}[t]
	\centering
	\captionsetup[subfigure]{labelformat=empty}
	
	\begin{subfigure}{0.18\linewidth}
		\includegraphics[width=.98\linewidth]{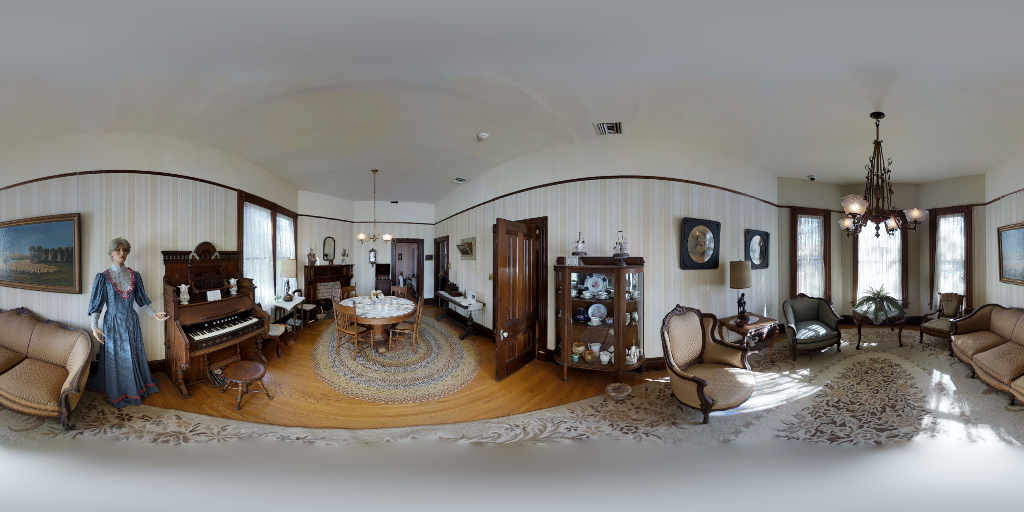}
	\end{subfigure}
	\begin{subfigure}{0.18\linewidth}
		\includegraphics[width=.98\linewidth]{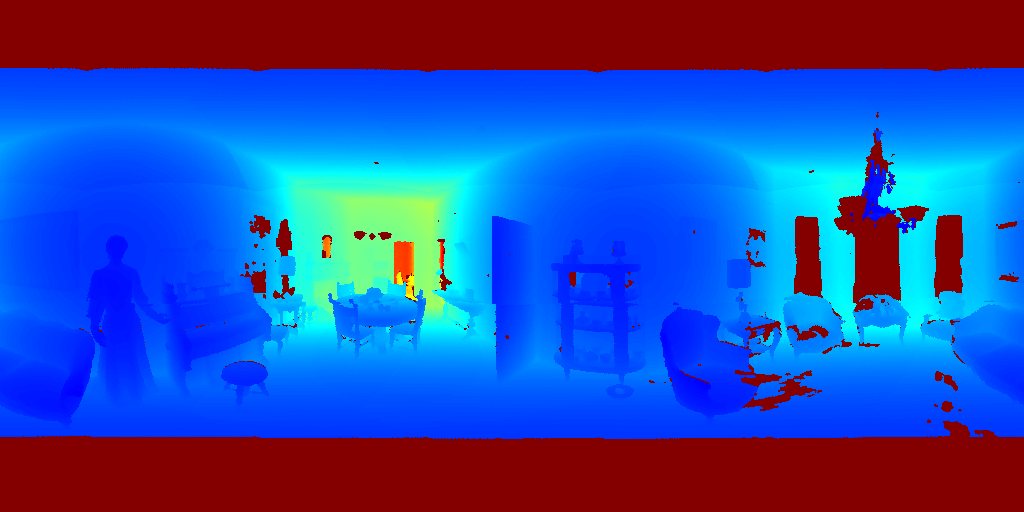}
	\end{subfigure}
	\begin{subfigure}{0.18\linewidth}
		\includegraphics[width=.98\linewidth]{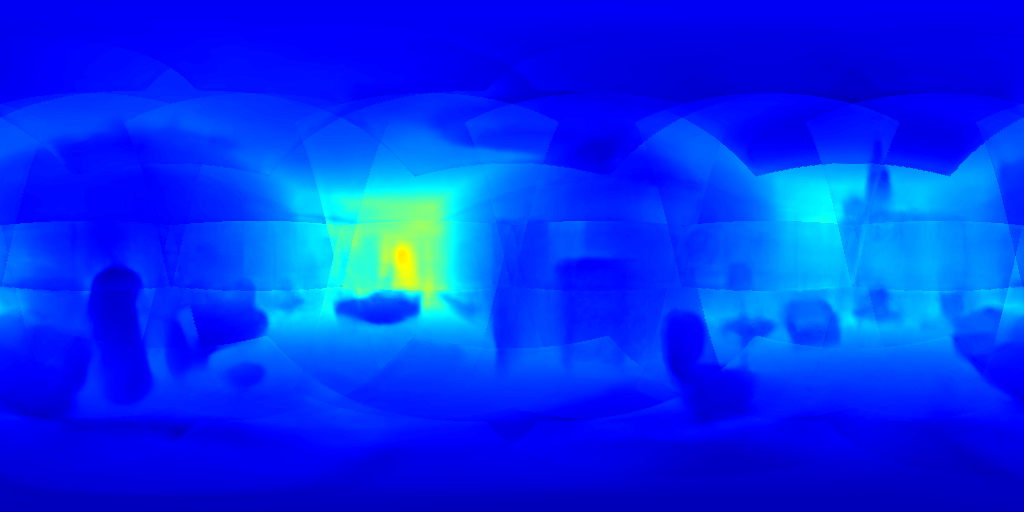}
	\end{subfigure}
	\begin{subfigure}{0.18\linewidth}
		\includegraphics[width=.98\linewidth]{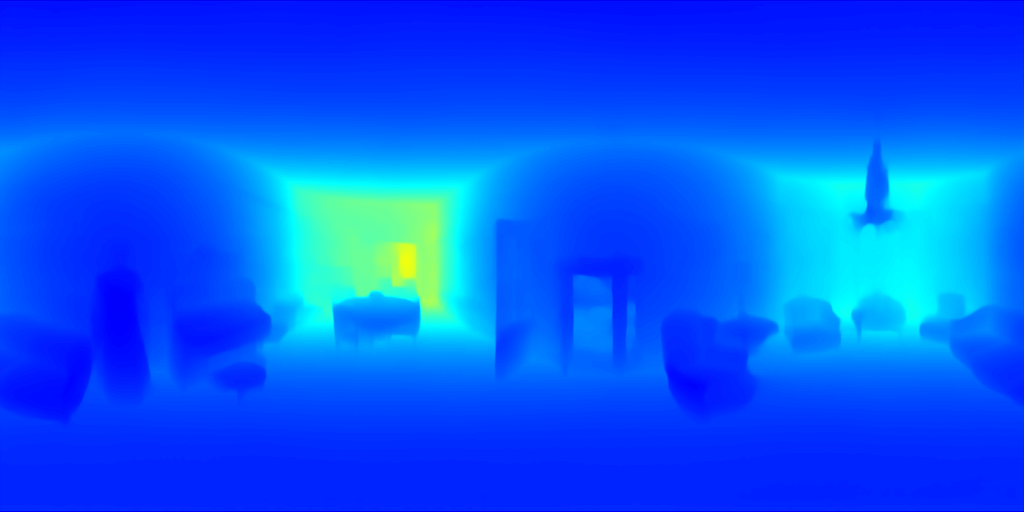}
	\end{subfigure}
	\begin{subfigure}{0.18\linewidth}
		\includegraphics[width=.98\linewidth]{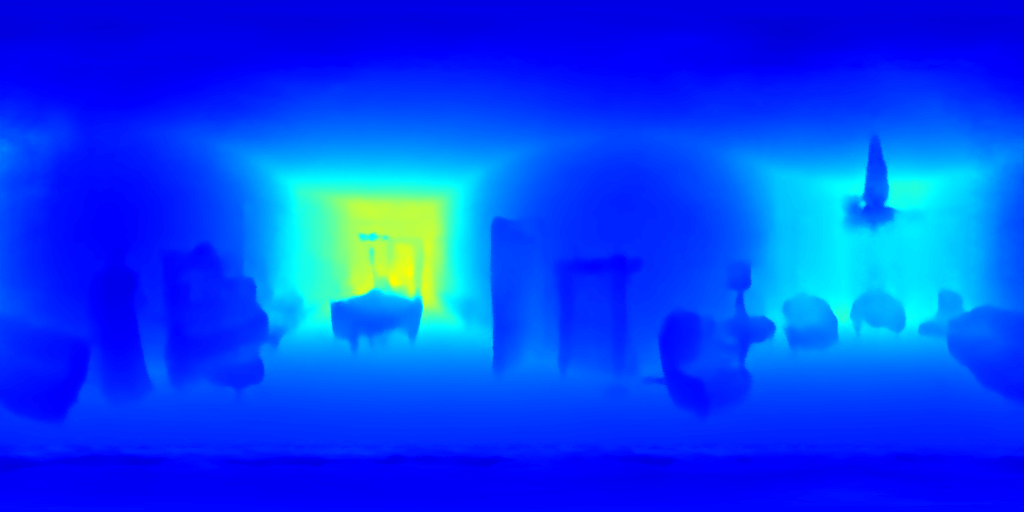}
	\end{subfigure}
	
	\vspace{1pt}

	\begin{subfigure}{0.18\linewidth}
		\includegraphics[width=.98\linewidth]{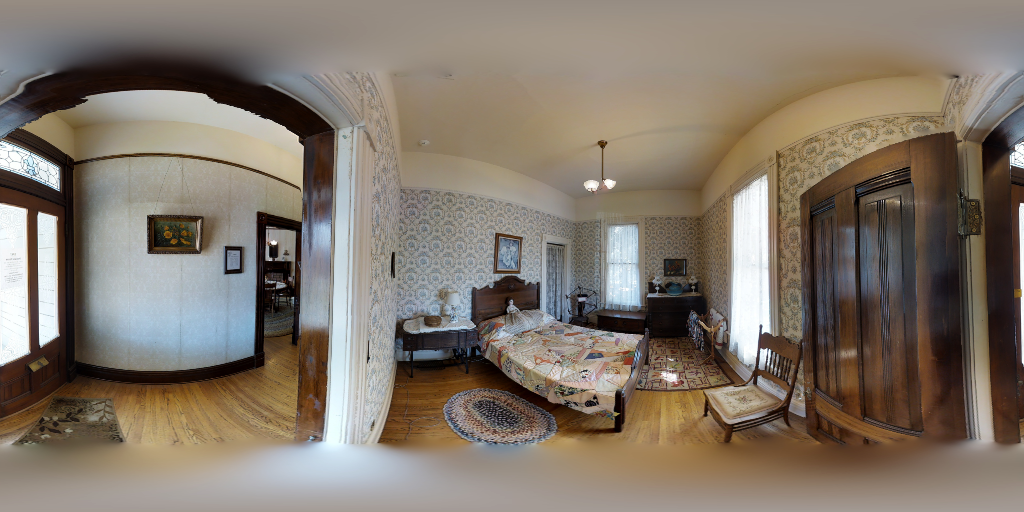}
	\end{subfigure}
	\begin{subfigure}{0.18\linewidth}
		\includegraphics[width=.98\linewidth]{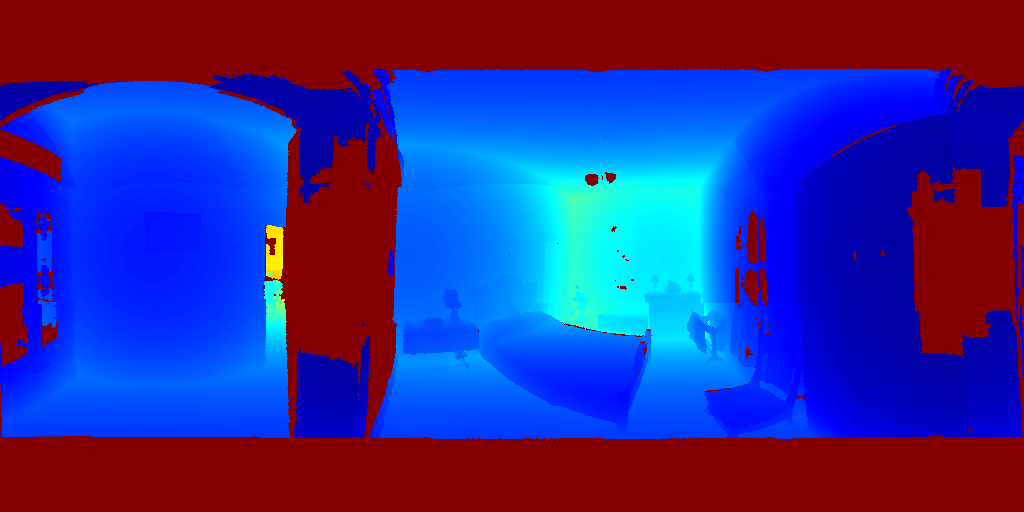}
	\end{subfigure}
	\begin{subfigure}{0.18\linewidth}
		\includegraphics[width=.98\linewidth]{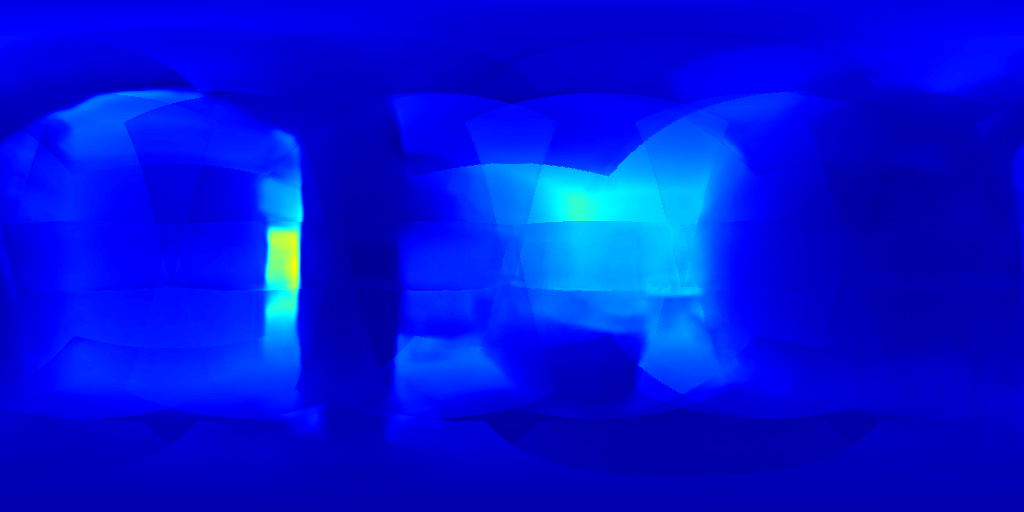}
	\end{subfigure}
	\begin{subfigure}{0.18\linewidth}
		\includegraphics[width=.98\linewidth]{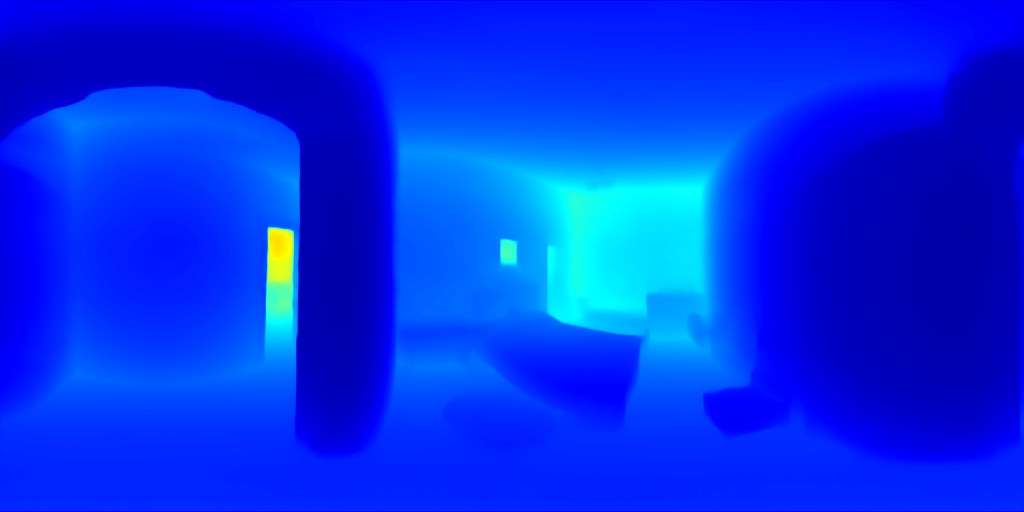}
	\end{subfigure}
	\begin{subfigure}{0.18\linewidth}
		\includegraphics[width=.98\linewidth]{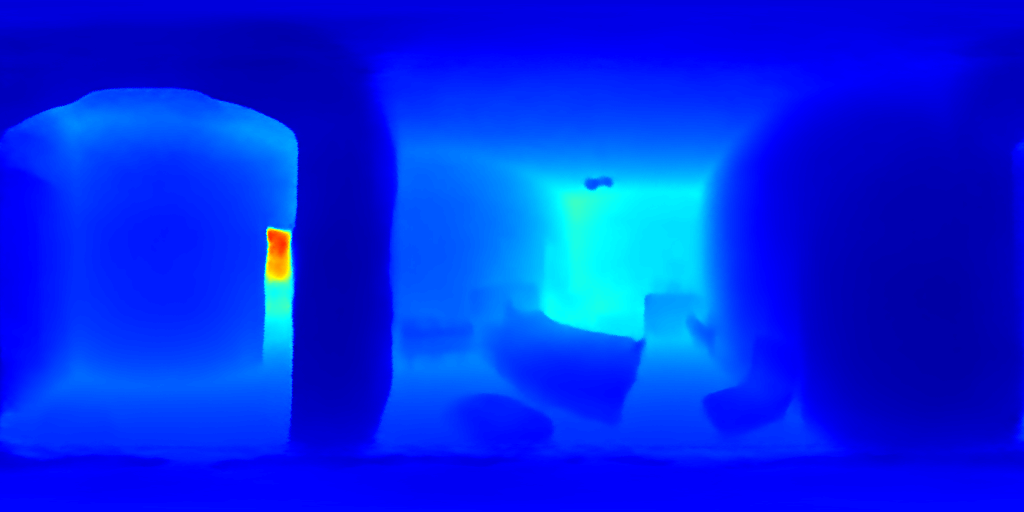}
	\end{subfigure}
	
	\vspace{1pt}
	
	\begin{subfigure}{0.18\linewidth}
		\includegraphics[width=.98\linewidth]{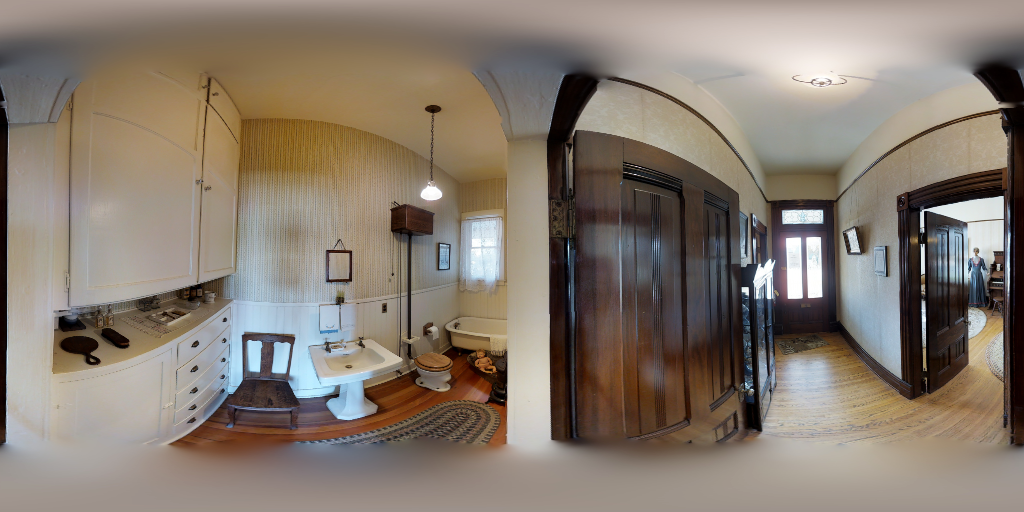}
	\end{subfigure}
	\begin{subfigure}{0.18\linewidth}
		\includegraphics[width=.98\linewidth]{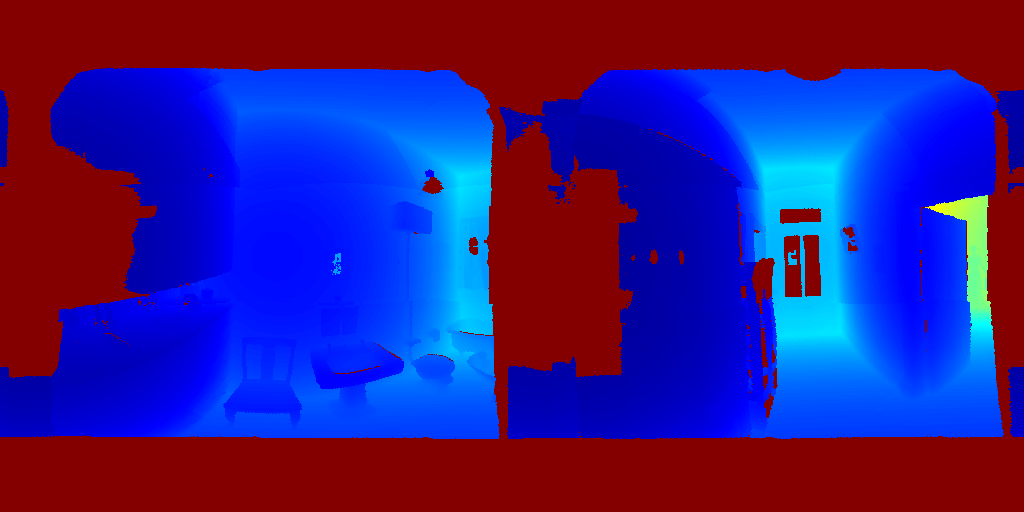}
	\end{subfigure}
	\begin{subfigure}{0.18\linewidth}
		\includegraphics[width=.98\linewidth]{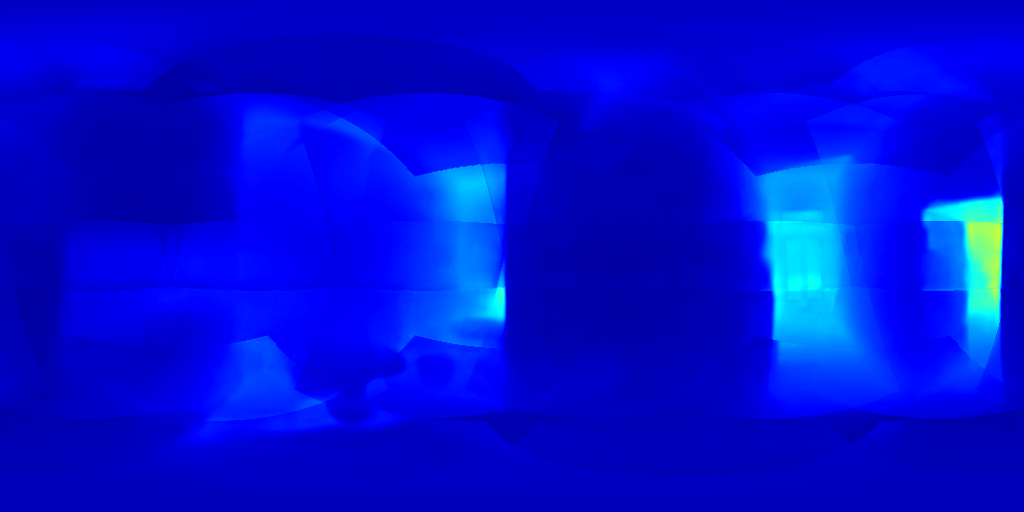}
	\end{subfigure}
	\begin{subfigure}{0.18\linewidth}
		\includegraphics[width=.98\linewidth]{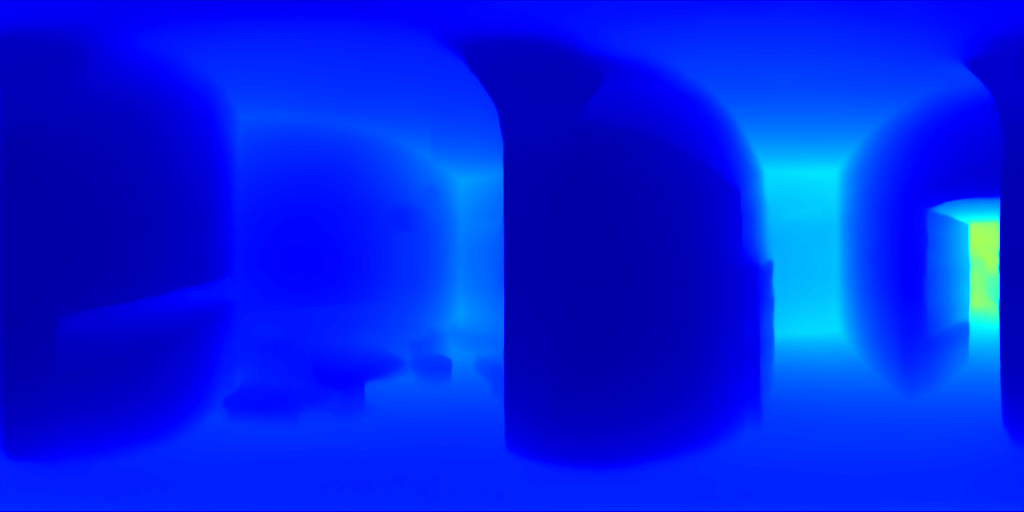}
	\end{subfigure}
	\begin{subfigure}{0.18\linewidth}
		\includegraphics[width=.98\linewidth]{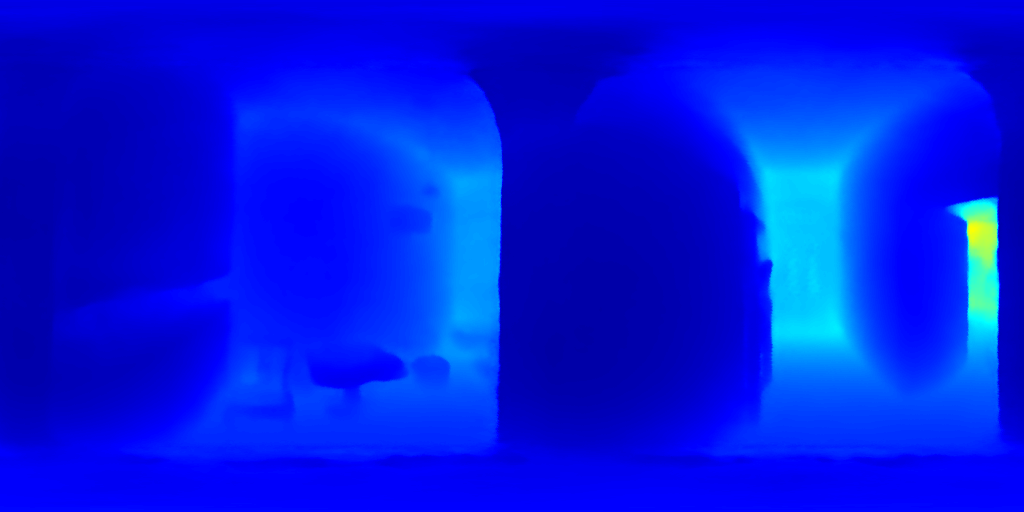}
	\end{subfigure}
	
	\vspace{1pt}
	
	\begin{subfigure}{0.18\linewidth}
		\includegraphics[width=.98\linewidth]{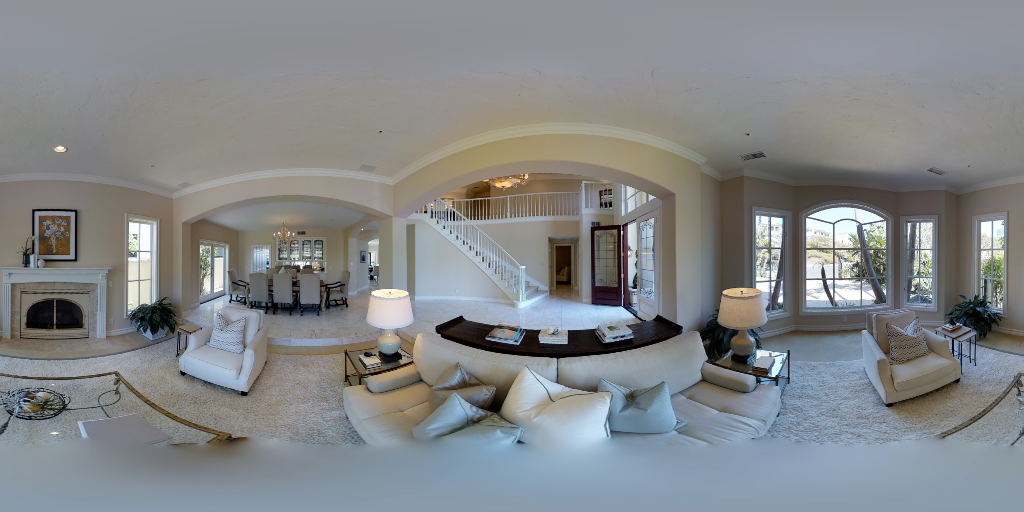}
		\caption{RGB}
	\end{subfigure}
	\begin{subfigure}{0.18\linewidth}
		\includegraphics[width=.98\linewidth]{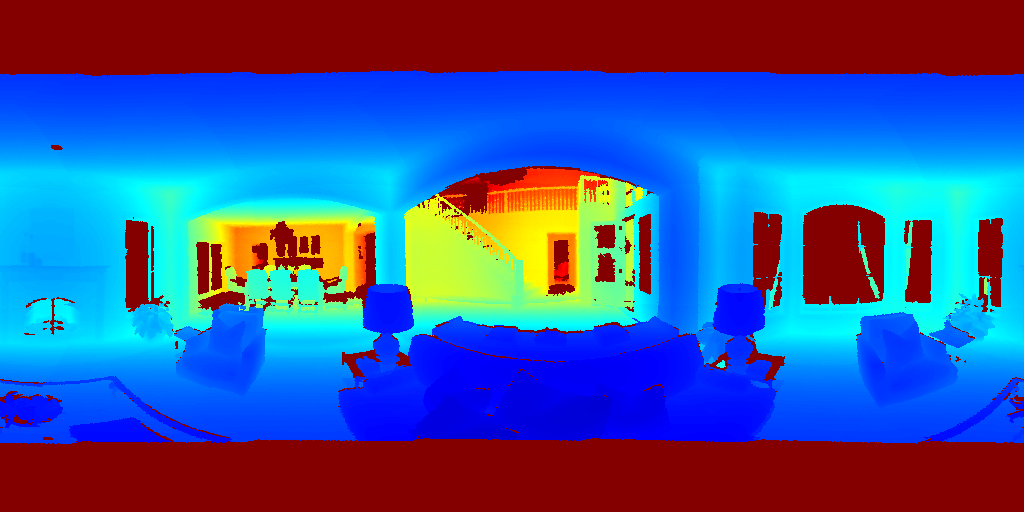}
		\caption{GT}
	\end{subfigure}
	\begin{subfigure}{0.18\linewidth}
		\includegraphics[width=.98\linewidth]{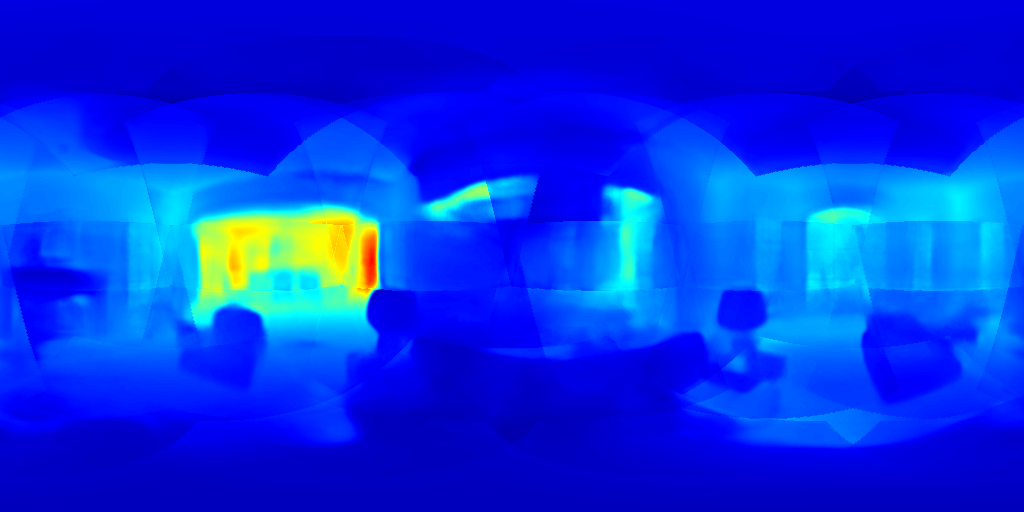}
		\caption{OmniFusion \cite{li2022omnifusion}}
	\end{subfigure}
	\begin{subfigure}{0.18\linewidth}
		\includegraphics[width=.98\linewidth]{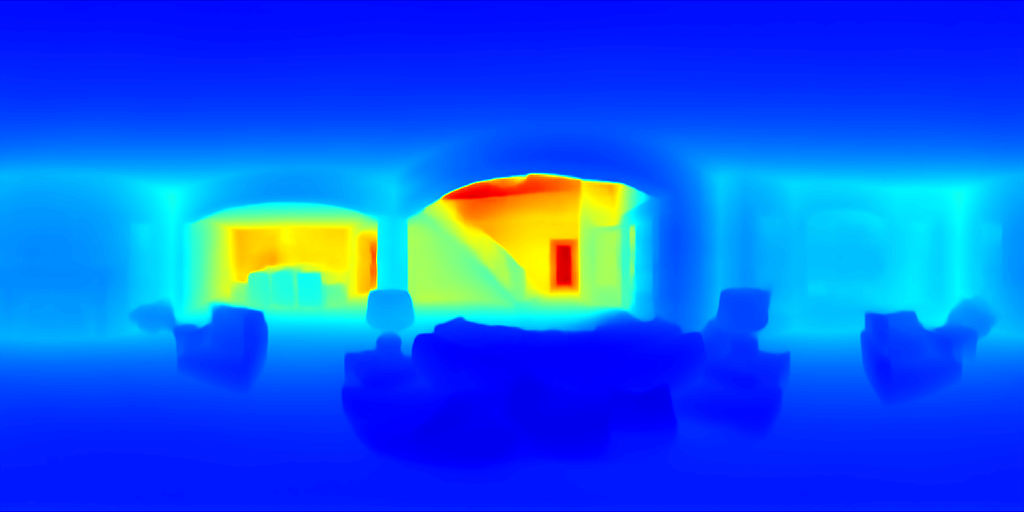}
		\caption{PanoFormer \cite{shen2022panoformer}}
	\end{subfigure}
	\begin{subfigure}{0.18\linewidth}
		\includegraphics[width=.98\linewidth]{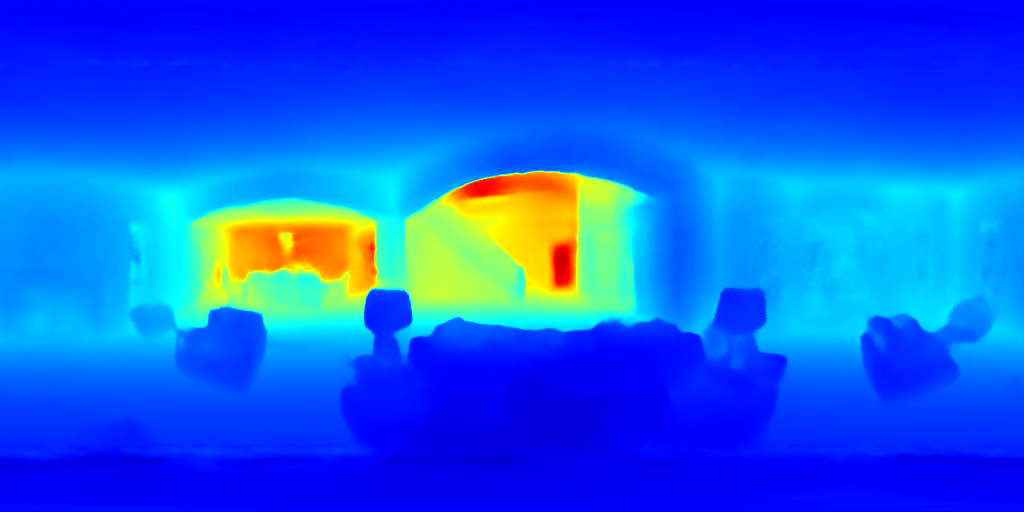}
		\caption{SphereFusion (ours)}
	\end{subfigure}

	\caption{
		\textbf{Depth Maps of Matterport3D.} Invalid parts of the depth map are set to red. 
	}
	\label{fig:mat3d_depth}
    \vspace{-1.0em}
\end{figure*}

\begin{figure*}[t]
	\centering
	\captionsetup[subfigure]{labelformat=empty}

	\begin{subfigure}{0.3\linewidth}
		\includegraphics[width=.98\linewidth]{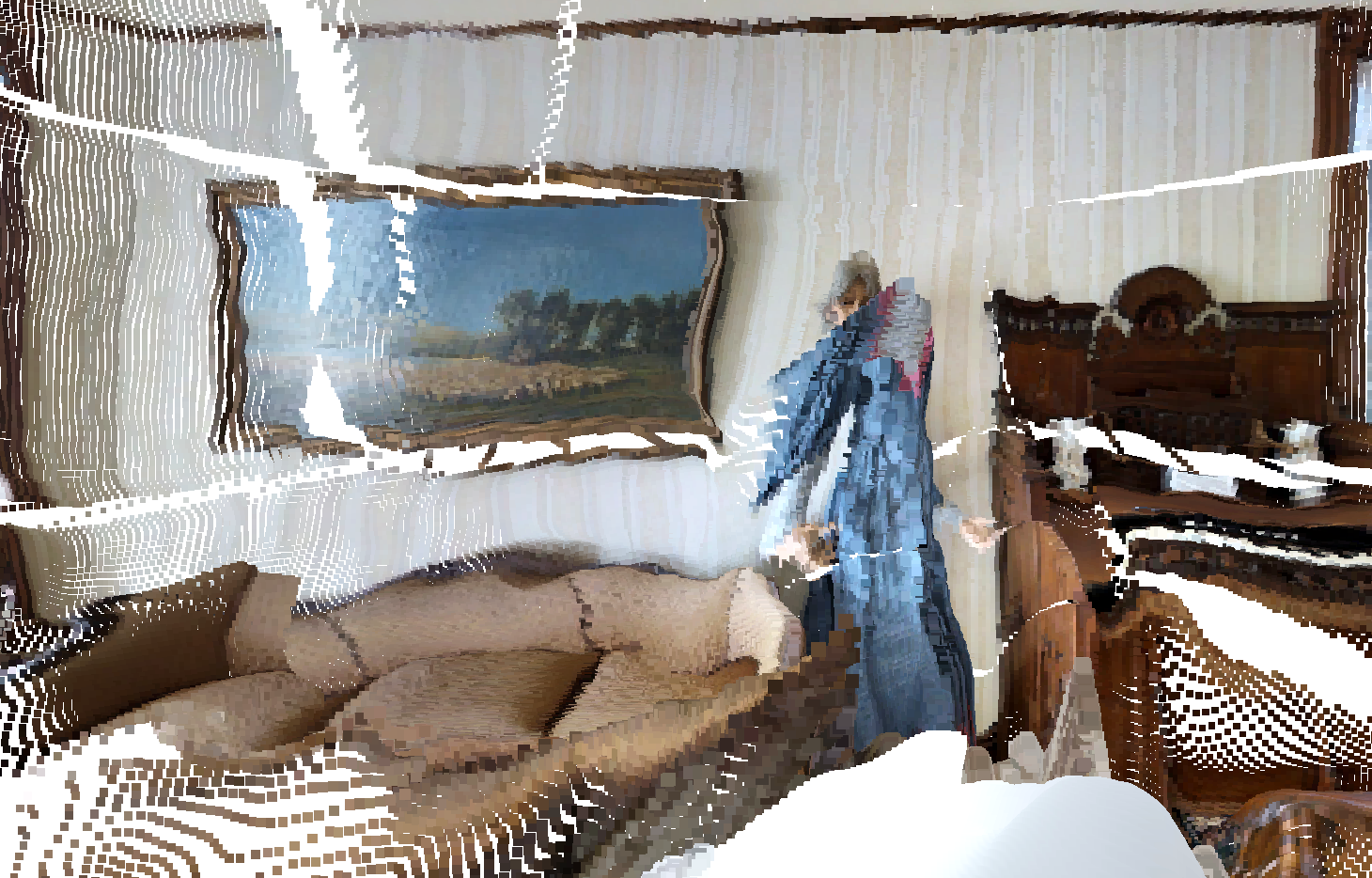}
	\end{subfigure}
	\begin{subfigure}{0.3\linewidth}
		\includegraphics[width=.98\linewidth]{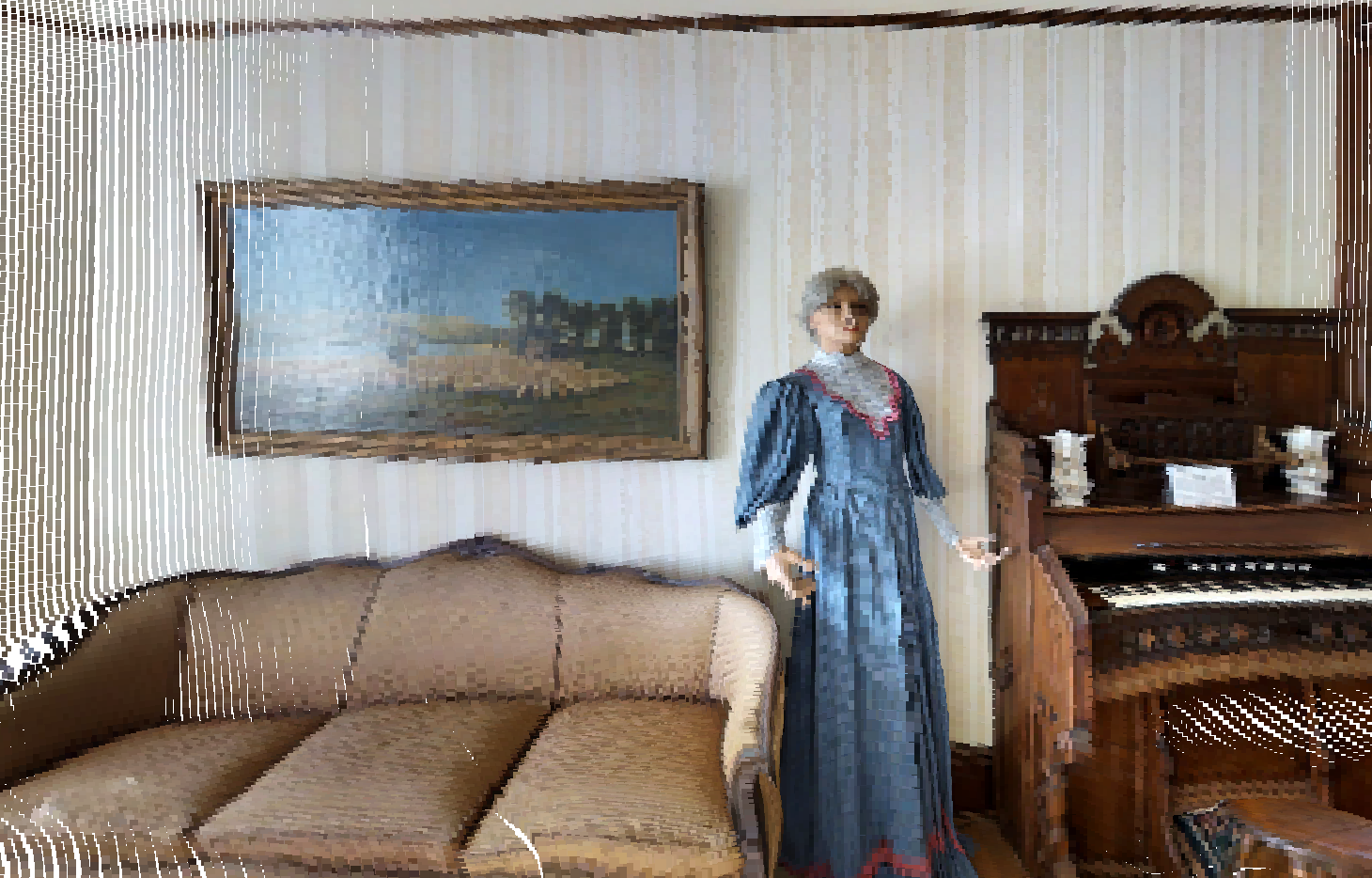}
	\end{subfigure}
	\begin{subfigure}{0.3\linewidth}
		\includegraphics[width=.98\linewidth]{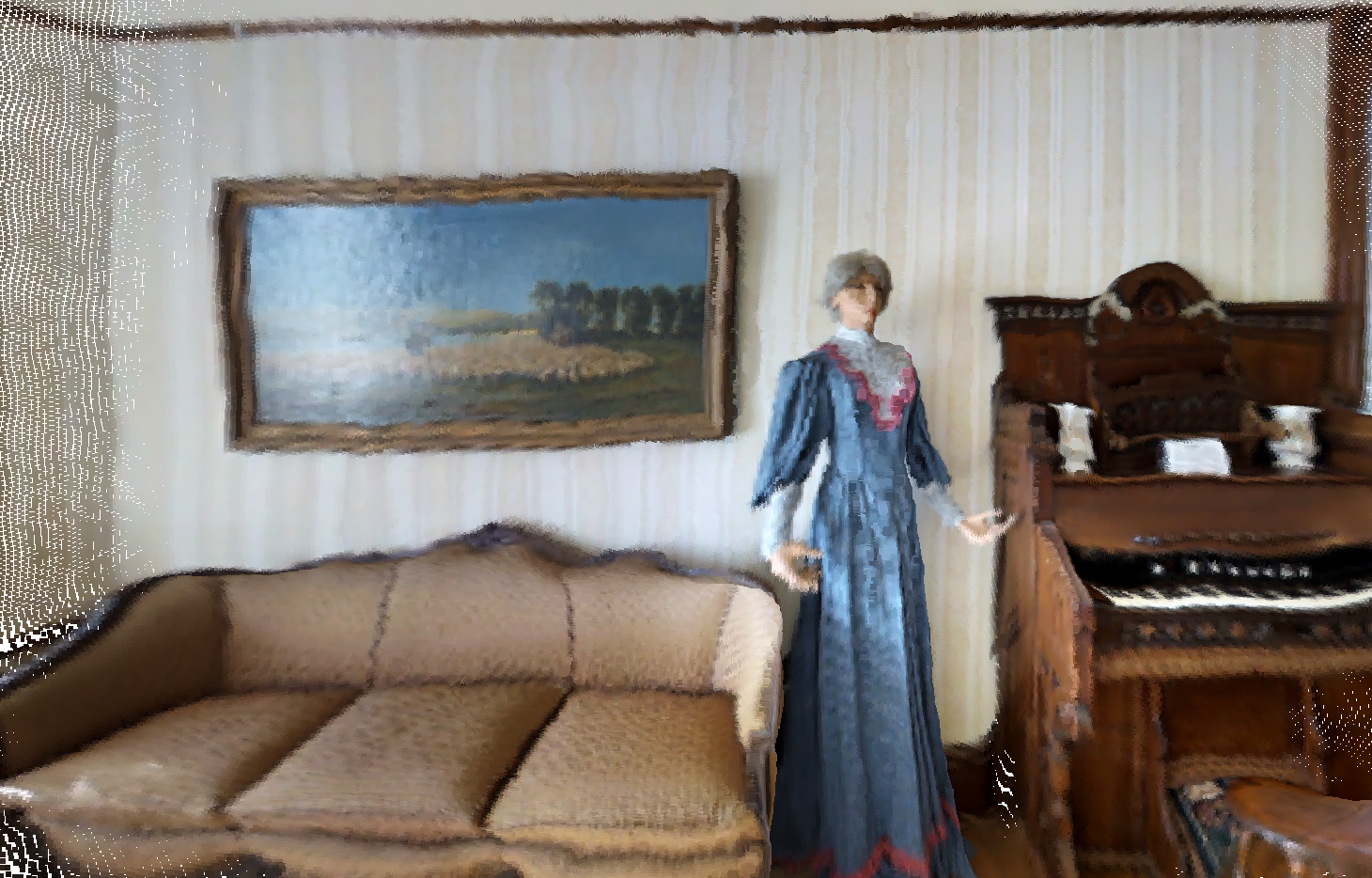}
	\end{subfigure}
	
	\vspace{1pt}

	\begin{subfigure}{0.3\linewidth}
		\includegraphics[width=.98\linewidth]{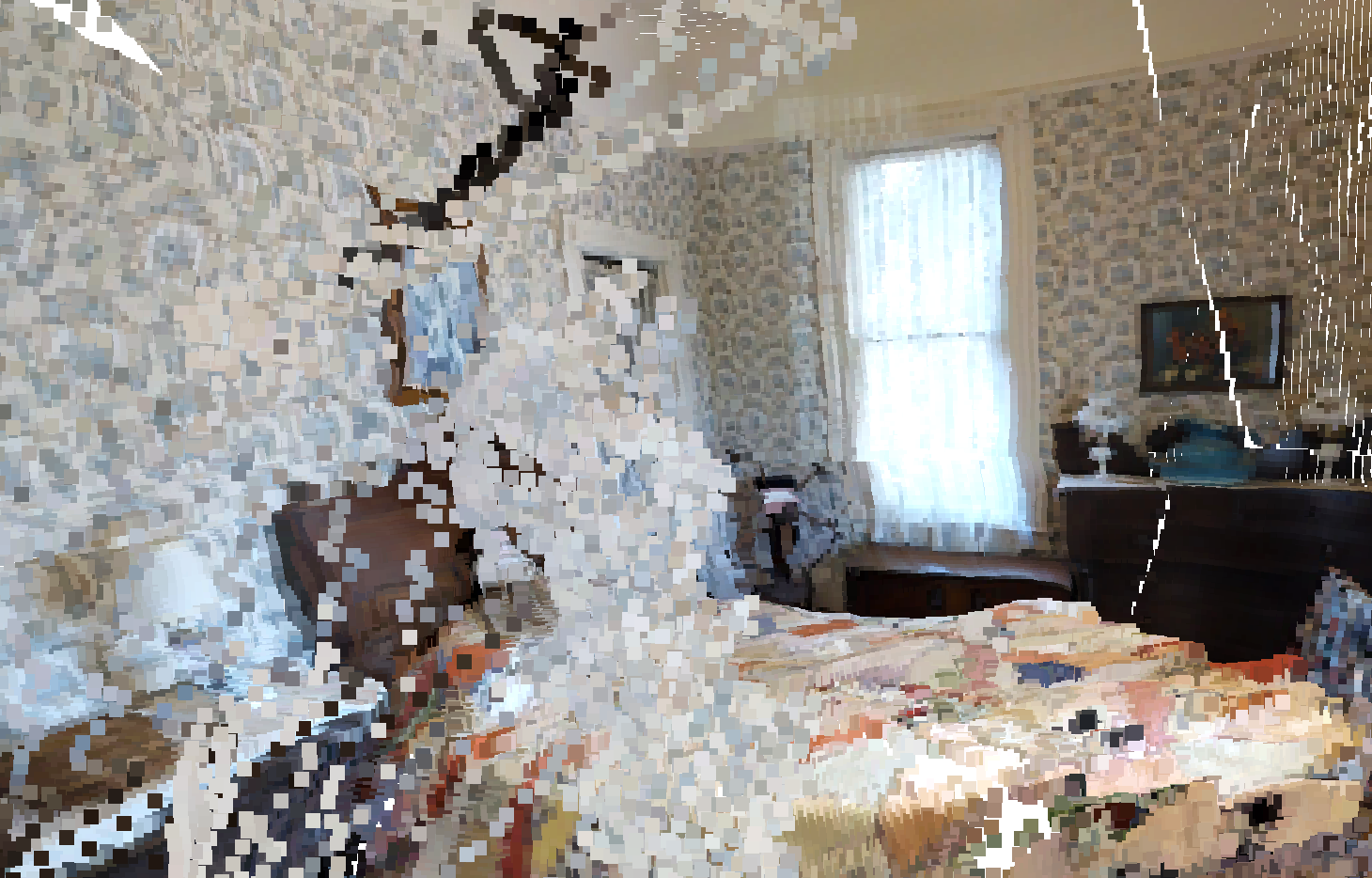}
	\end{subfigure}
	\begin{subfigure}{0.3\linewidth}
		\includegraphics[width=.98\linewidth]{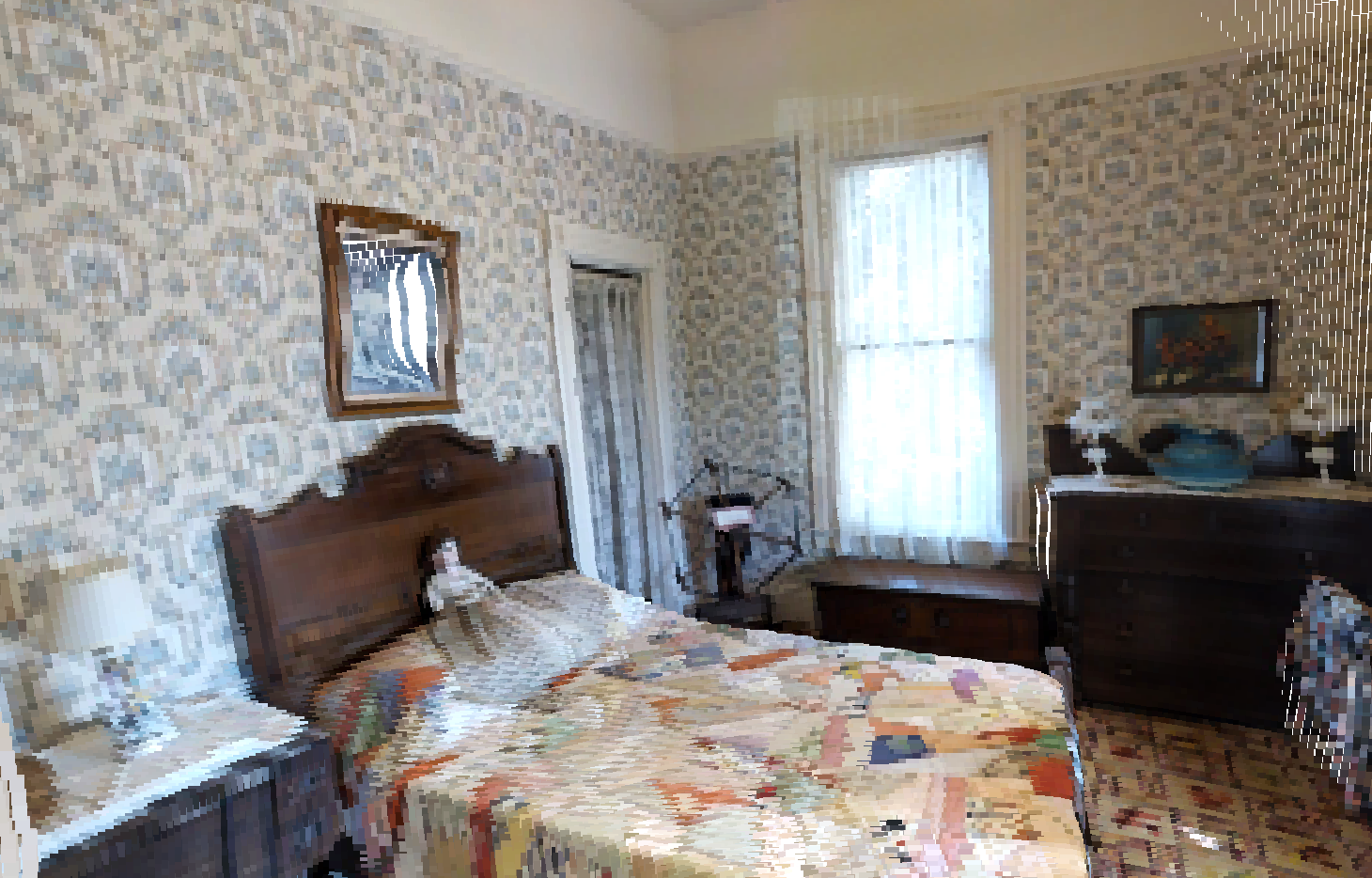}
	\end{subfigure}
	\begin{subfigure}{0.3\linewidth}
		\includegraphics[width=.98\linewidth]{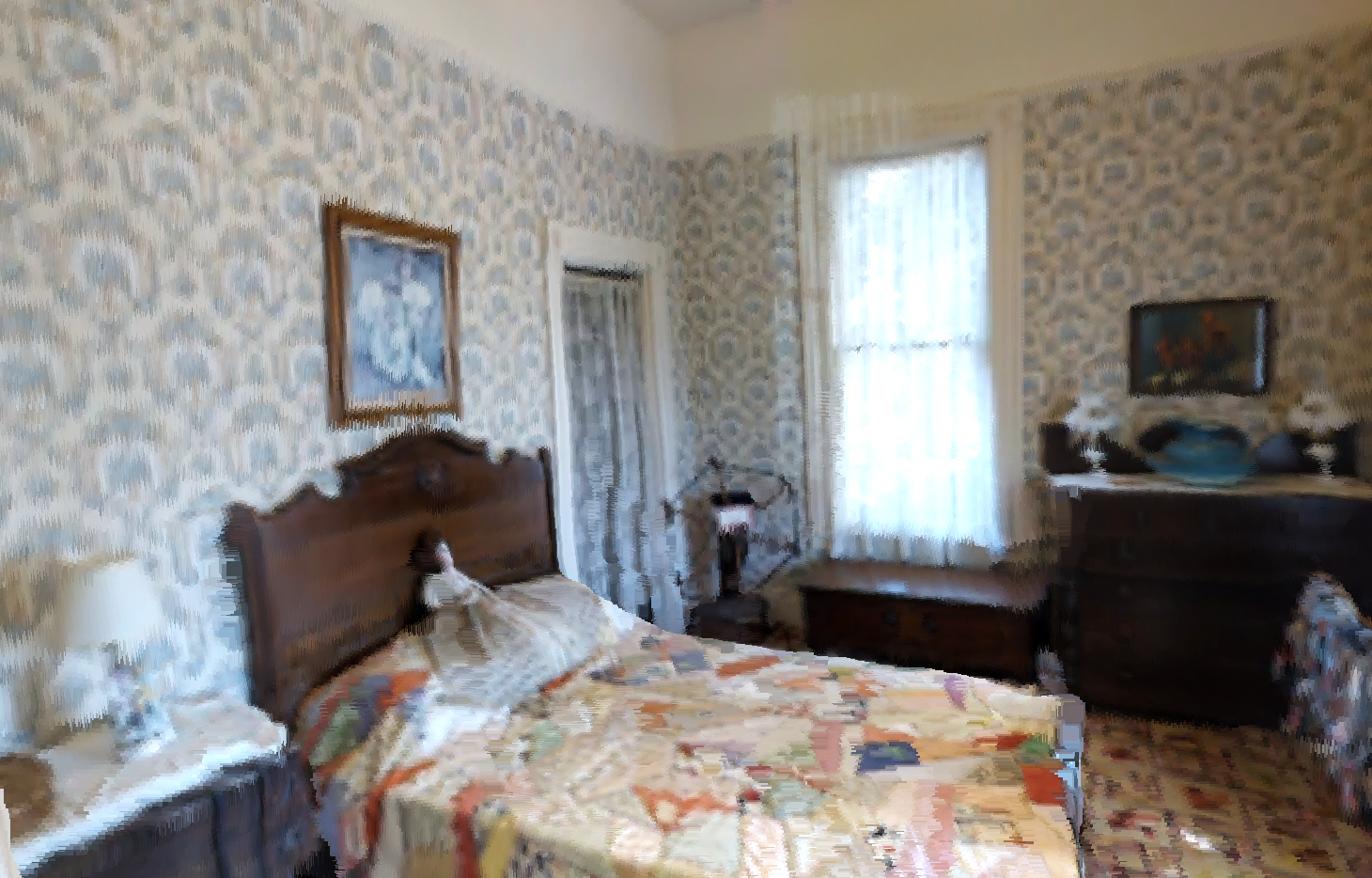}
	\end{subfigure}

	\vspace{1pt}
	
	\begin{subfigure}{0.3\linewidth}
		\includegraphics[width=.98\linewidth]{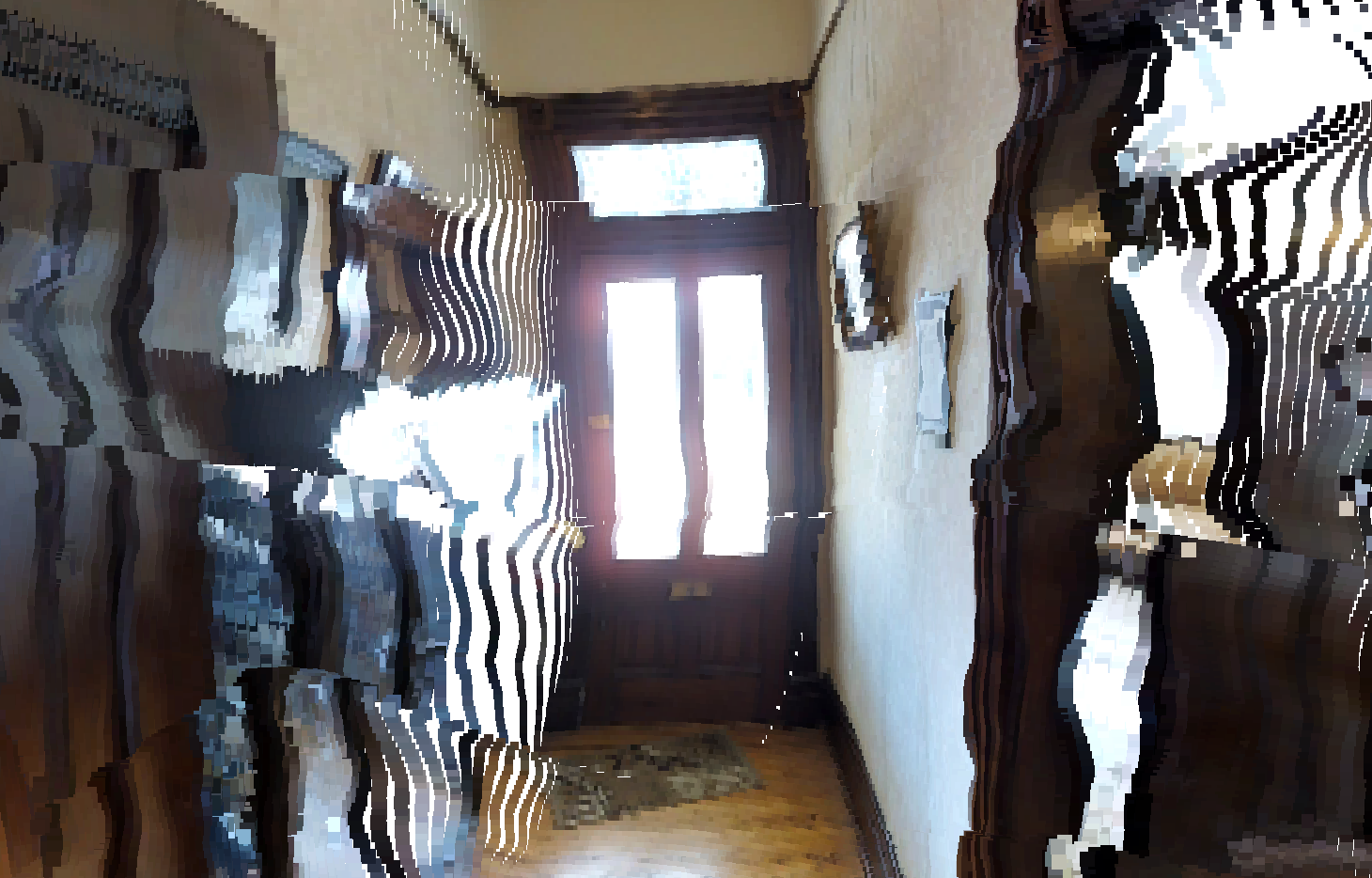}
	\end{subfigure}
	\begin{subfigure}{0.3\linewidth}
		\includegraphics[width=.98\linewidth]{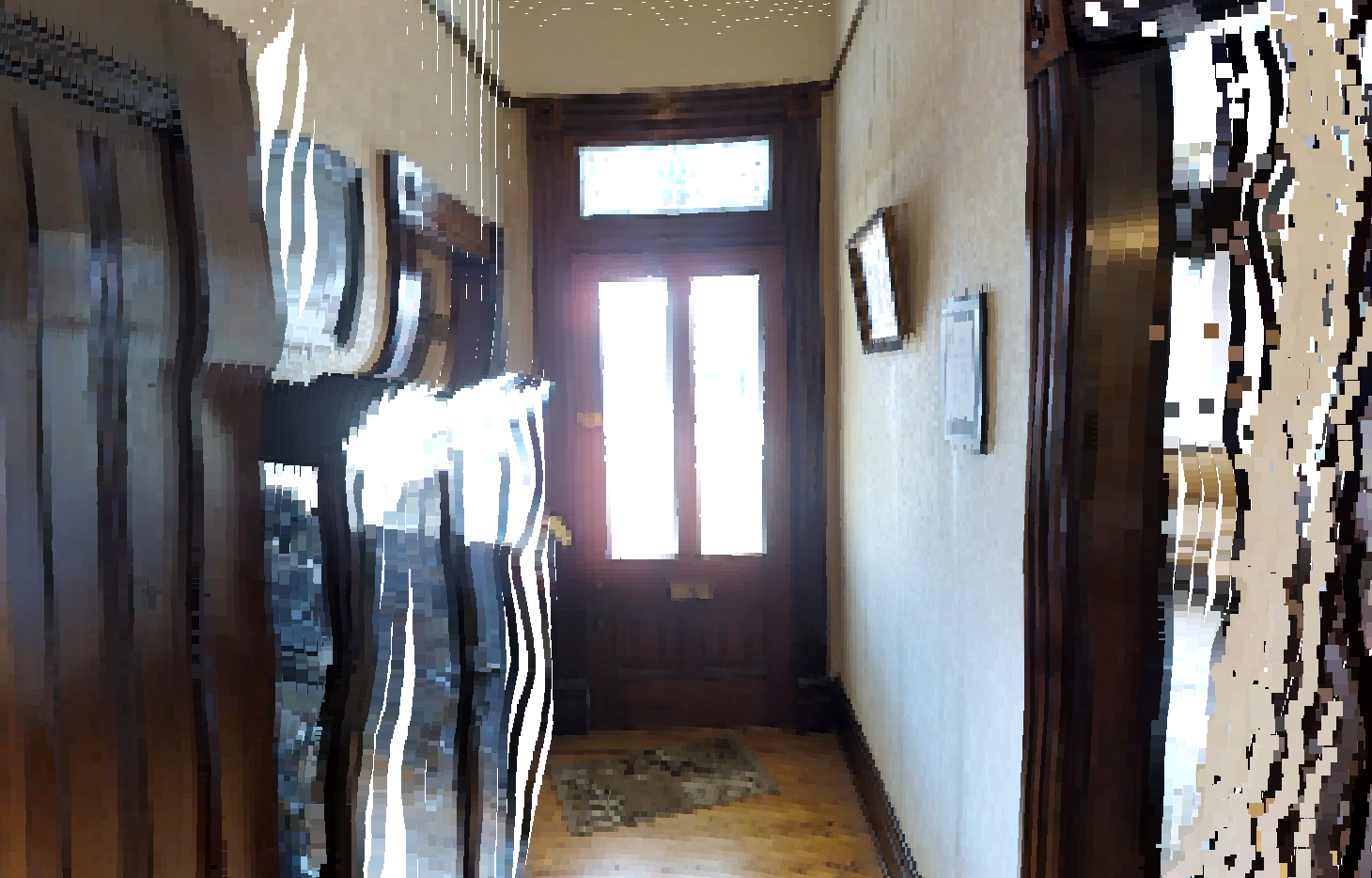}
	\end{subfigure}
	\begin{subfigure}{0.3\linewidth}
		\includegraphics[width=.98\linewidth]{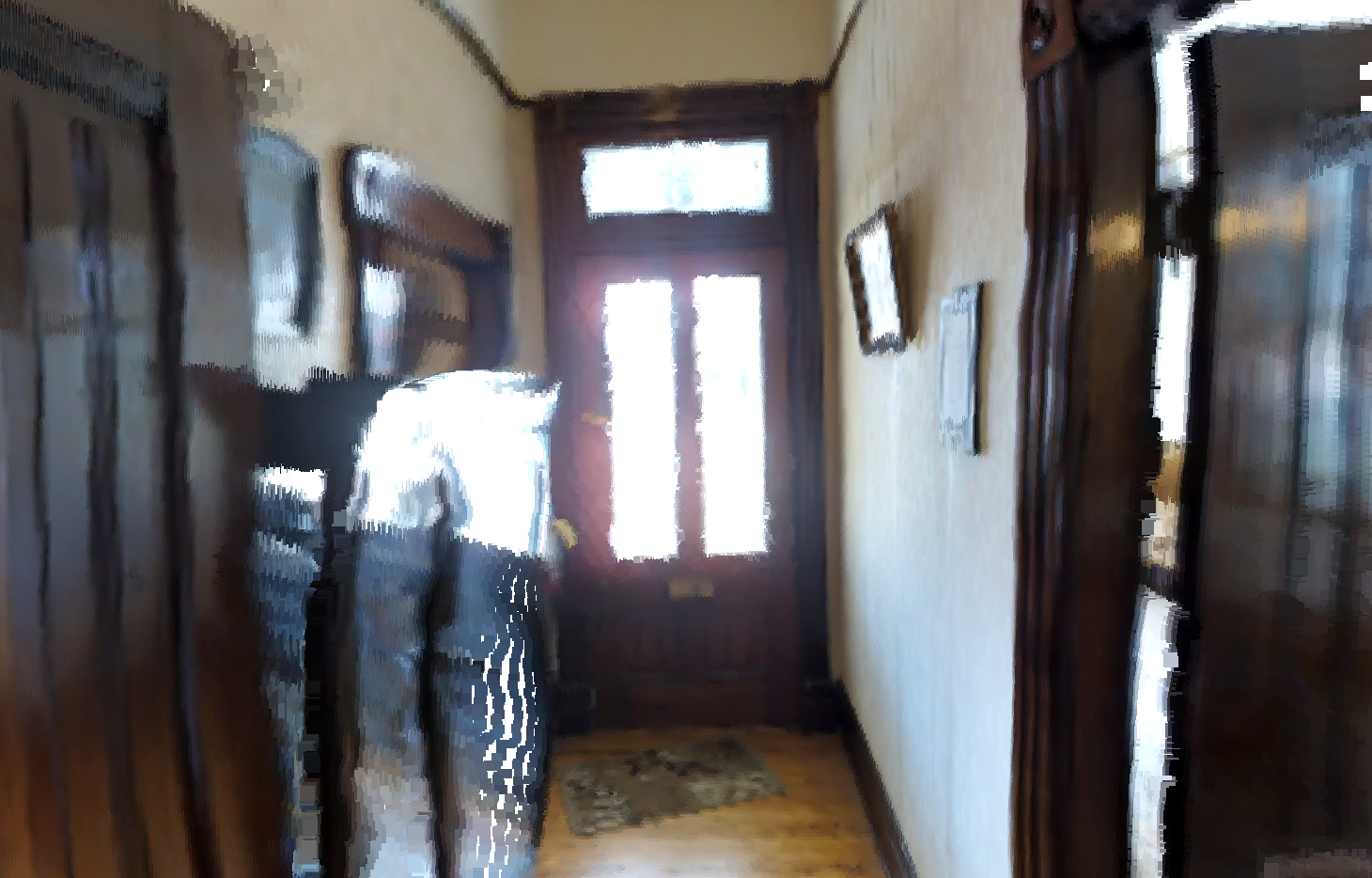}
	\end{subfigure}

	\vspace{1pt}
	
	\begin{subfigure}{0.3\linewidth}
		\includegraphics[width=.98\linewidth]{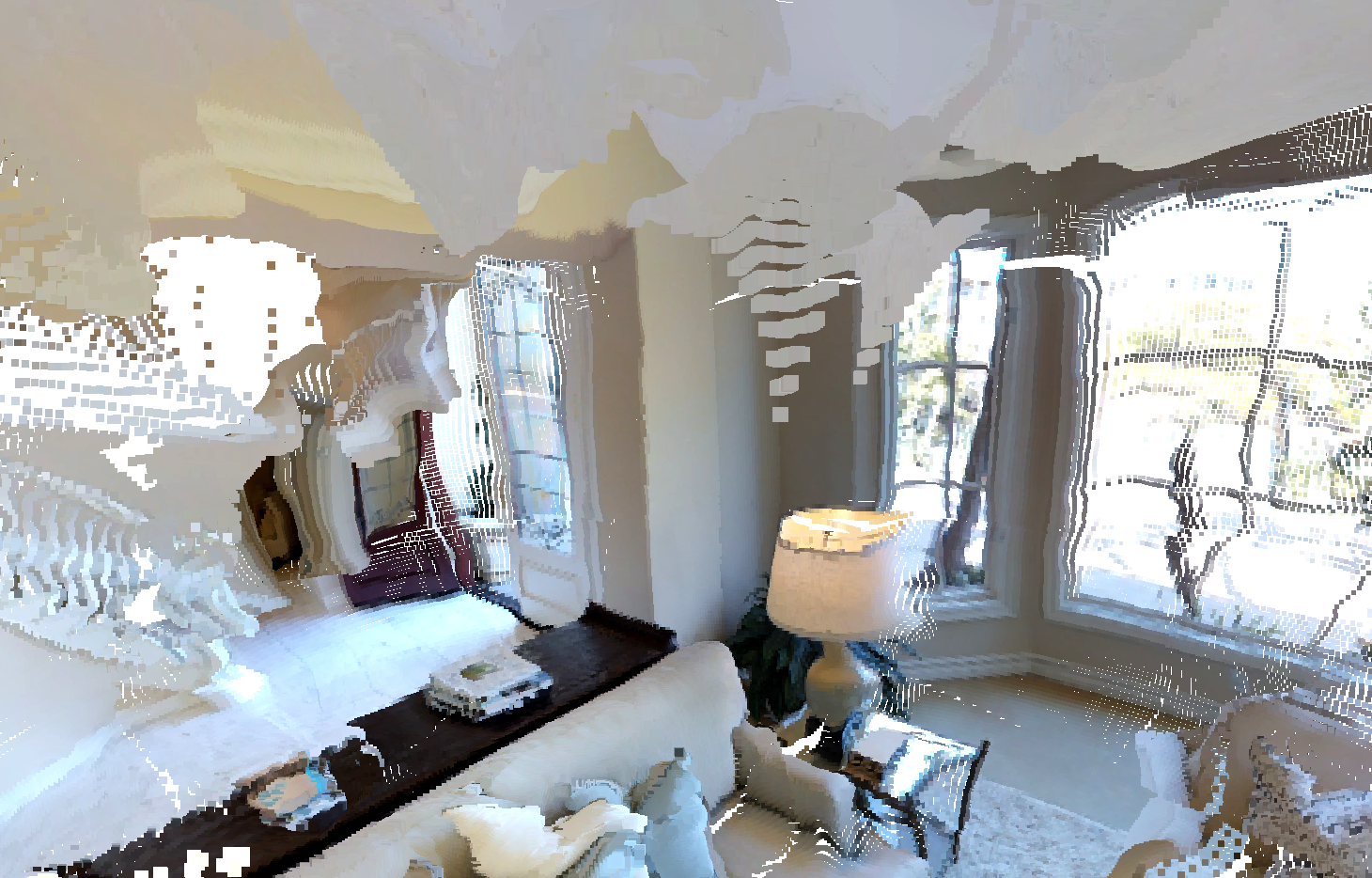}
		\caption{OmniFusion \cite{li2022omnifusion}}
	\end{subfigure}
	\begin{subfigure}{0.3\linewidth}
		\includegraphics[width=.98\linewidth]{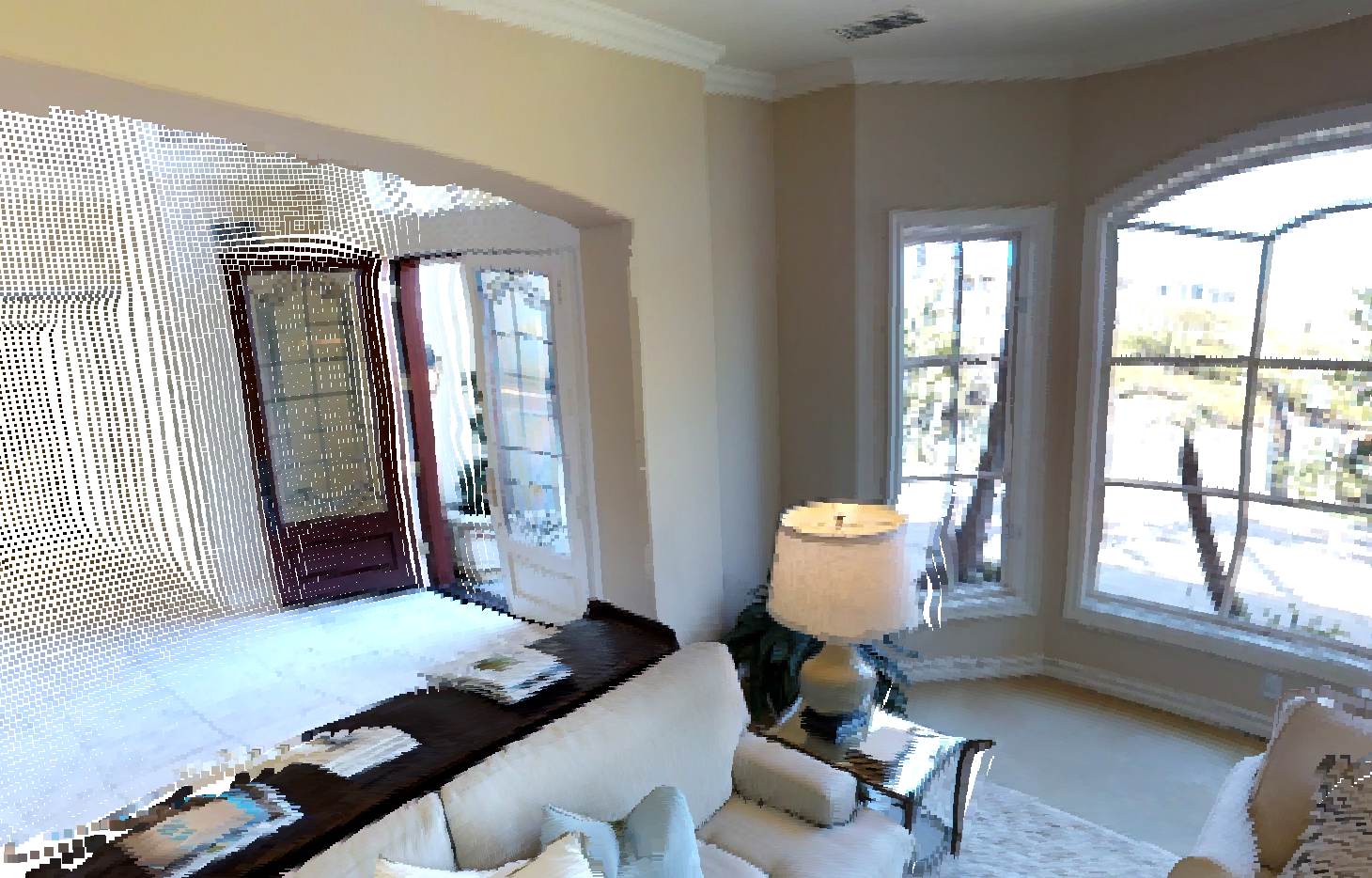}
		\caption{PanoFormer \cite{shen2022panoformer}}
	\end{subfigure}
	\begin{subfigure}{0.3\linewidth}
		\includegraphics[width=.98\linewidth]{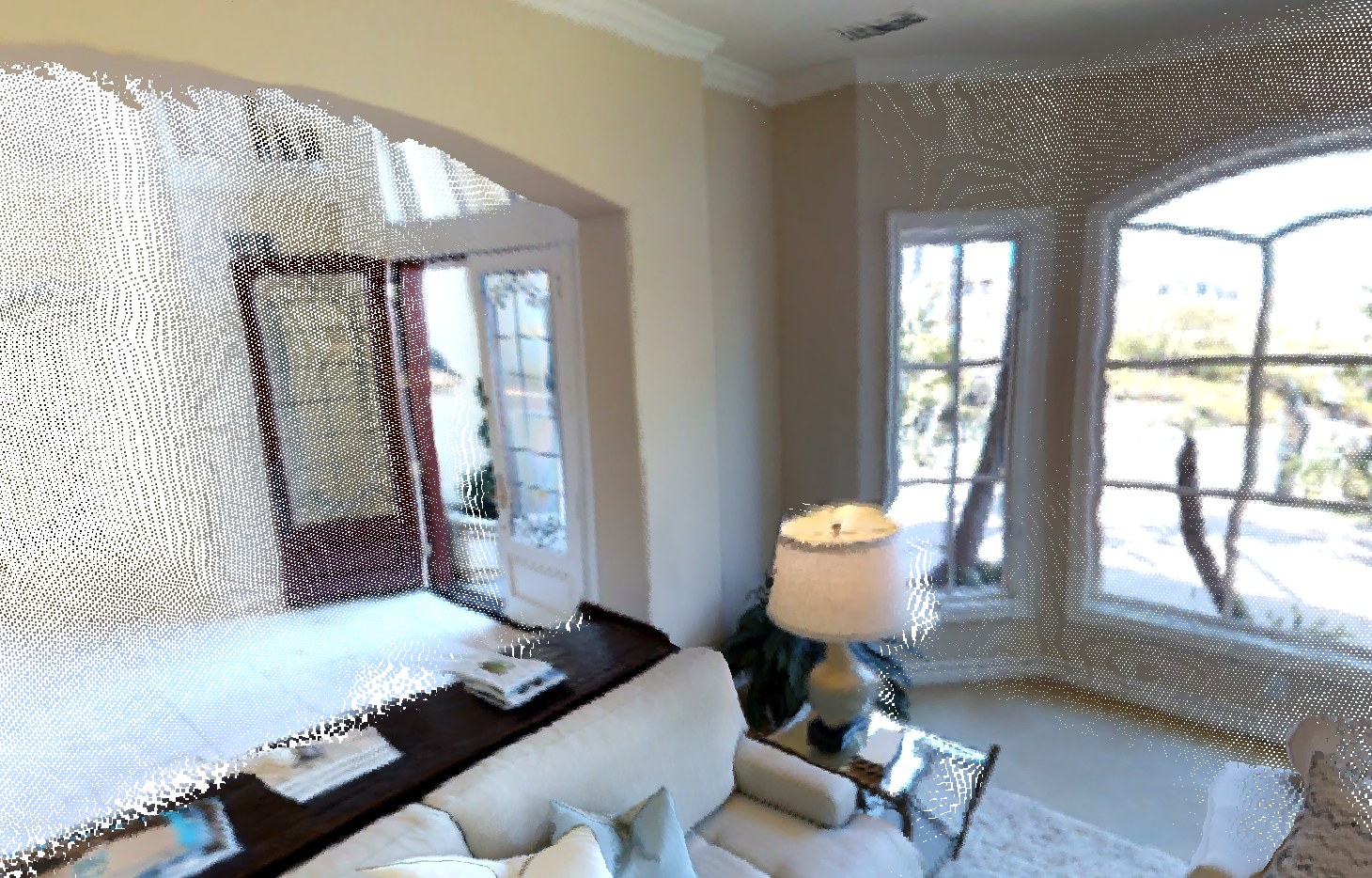}
		\caption{SphereFusion (ours)}
	\end{subfigure}

	\caption{
		\textbf{Depth Maps of Matterport3D.} Our method has less noise and maintains the structure of the scene.
	}
	\label{fig:mat3d_cloud}
    \vspace{-1.0em}
\end{figure*}

\begin{figure*}[t]
	\centering
	\captionsetup[subfigure]{labelformat=empty}
	
	\begin{subfigure}{0.18\linewidth}
		\includegraphics[width=.98\linewidth]{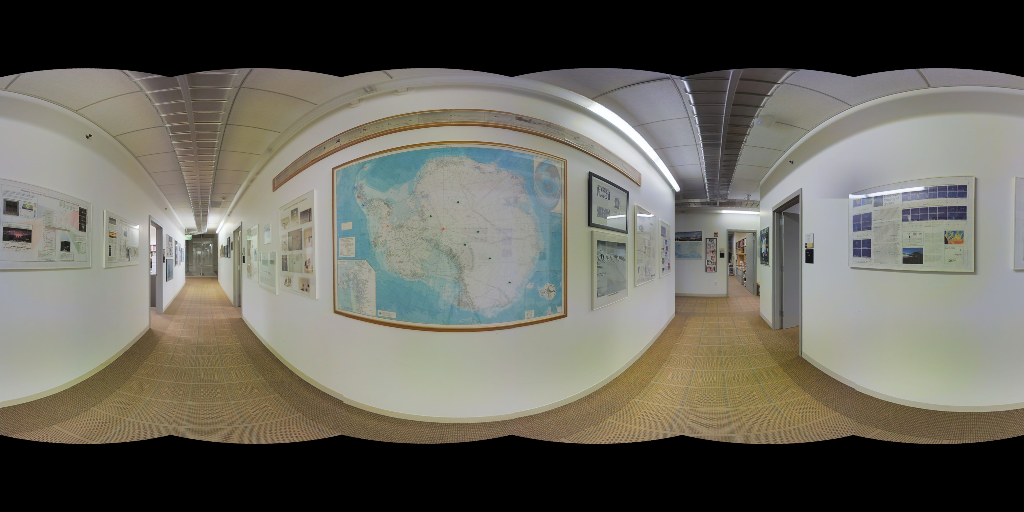}
	\end{subfigure}
	\begin{subfigure}{0.18\linewidth}
		\includegraphics[width=.98\linewidth]{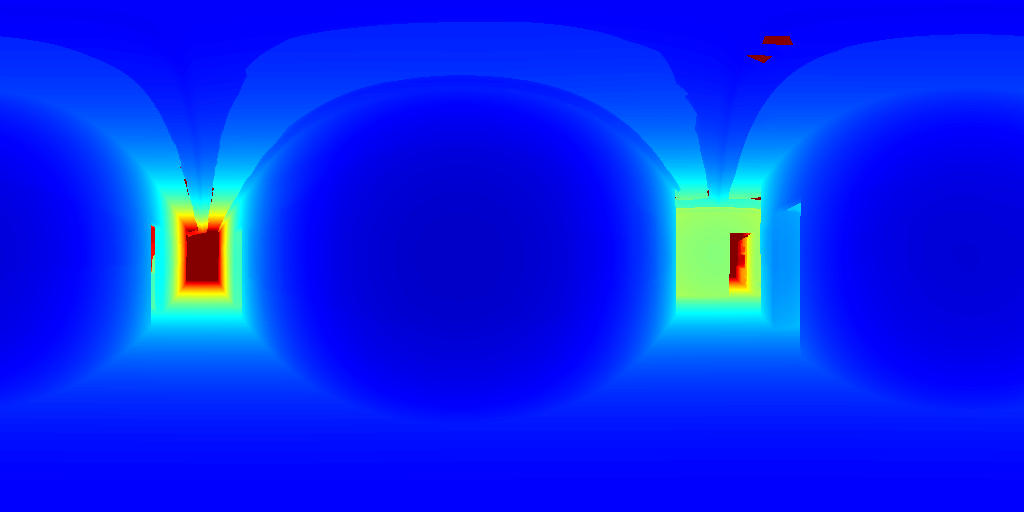}
	\end{subfigure}
	\begin{subfigure}{0.18\linewidth}
		\includegraphics[width=.98\linewidth]{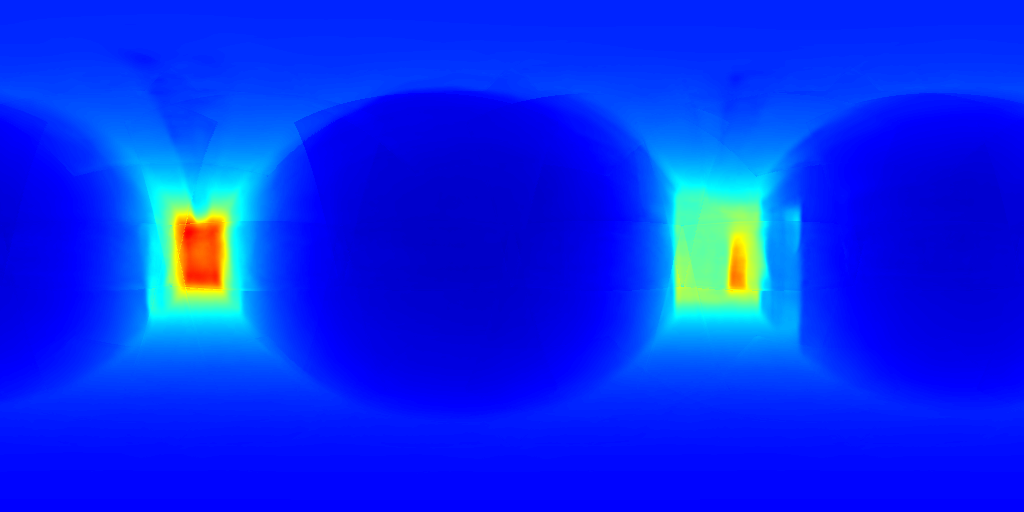}
	\end{subfigure}
	\begin{subfigure}{0.18\linewidth}
		\includegraphics[width=.98\linewidth]{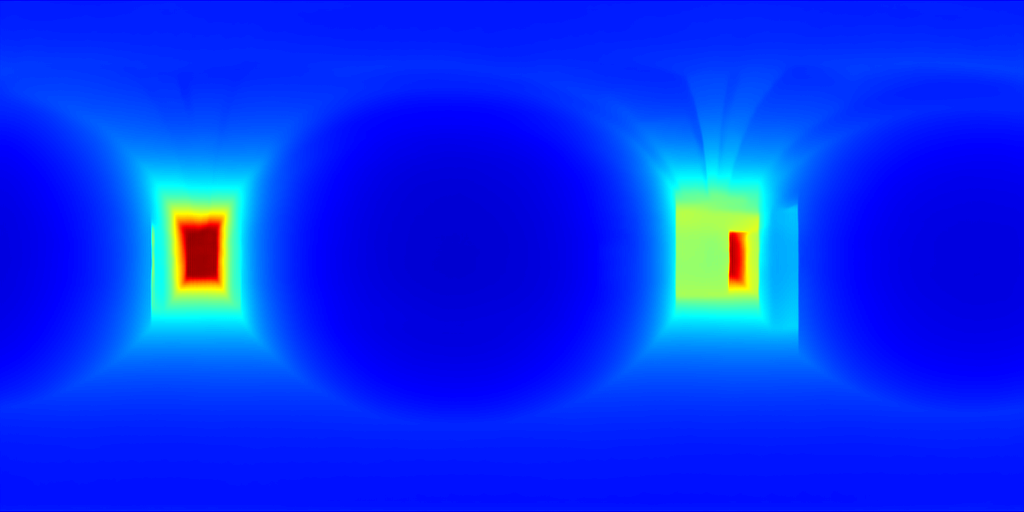}
	\end{subfigure}
	\begin{subfigure}{0.18\linewidth}
		\includegraphics[width=.98\linewidth]{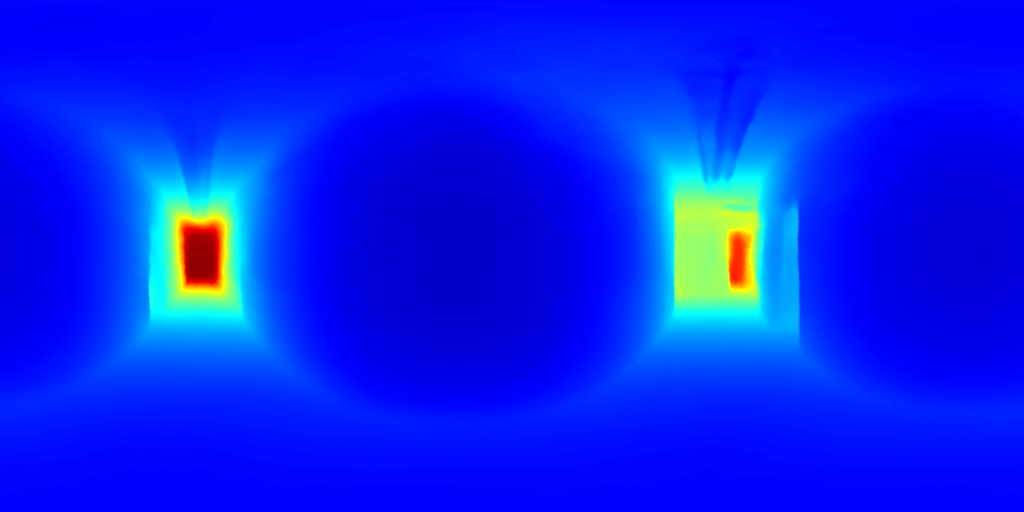}
	\end{subfigure}
	
	\vspace{1pt}

	\begin{subfigure}{0.18\linewidth}
		\includegraphics[width=.98\linewidth]{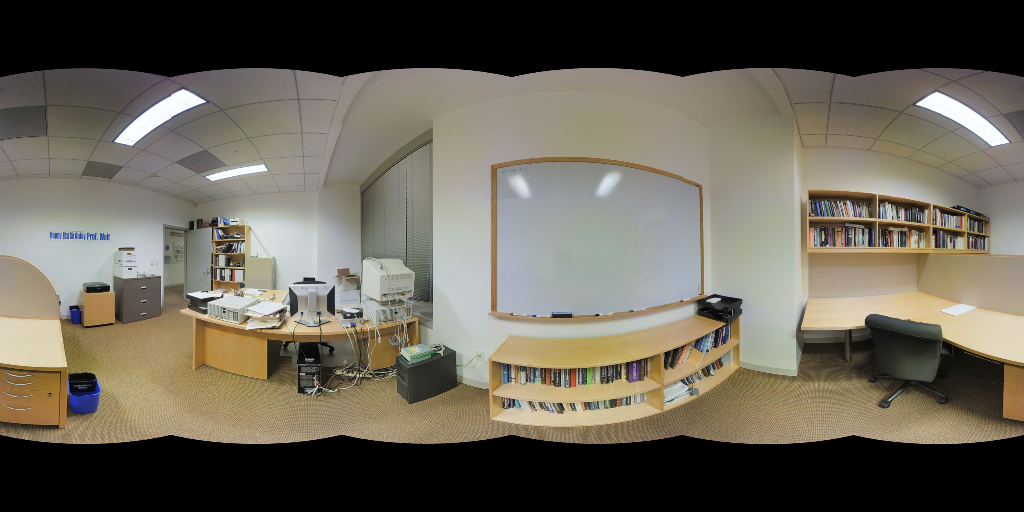}
	\end{subfigure}
	\begin{subfigure}{0.18\linewidth}
		\includegraphics[width=.98\linewidth]{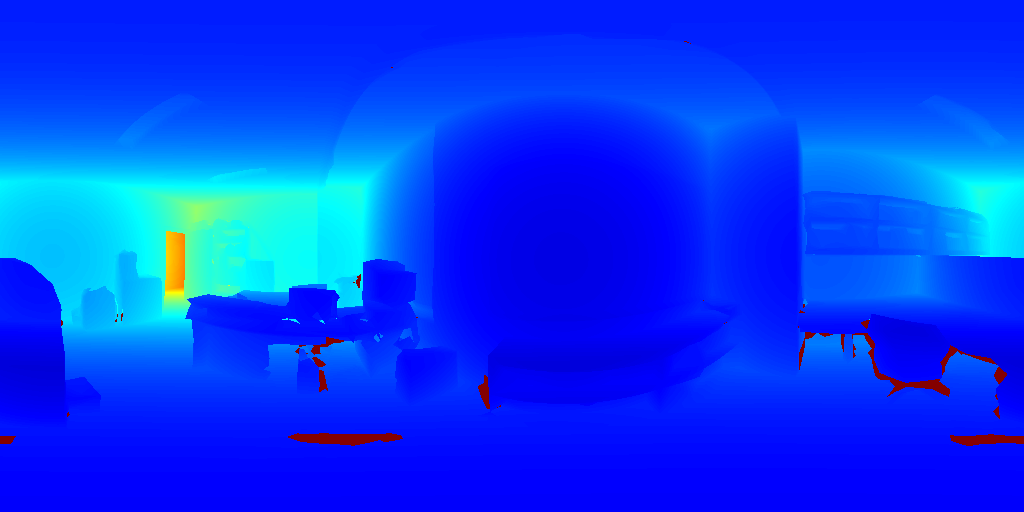}
	\end{subfigure}
	\begin{subfigure}{0.18\linewidth}
		\includegraphics[width=.98\linewidth]{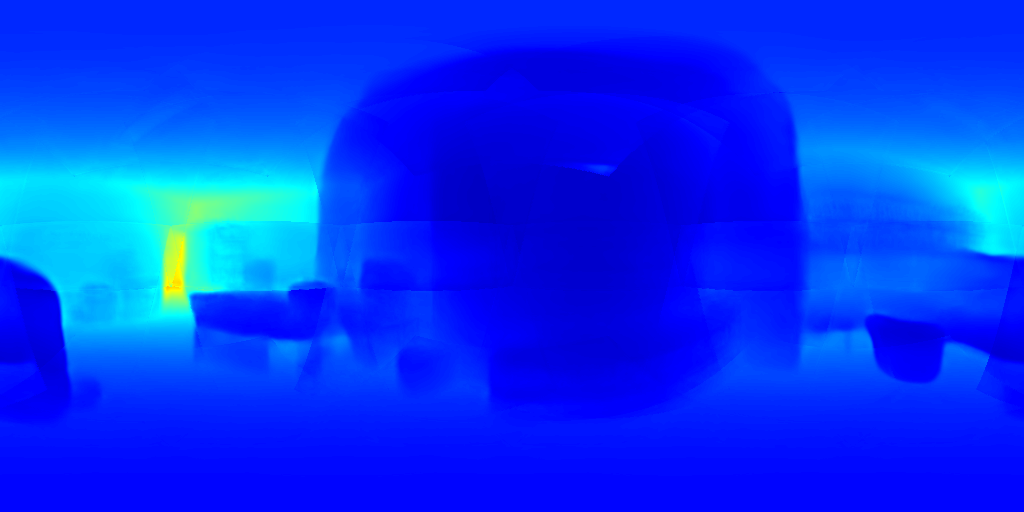}
	\end{subfigure}
	\begin{subfigure}{0.18\linewidth}
		\includegraphics[width=.98\linewidth]{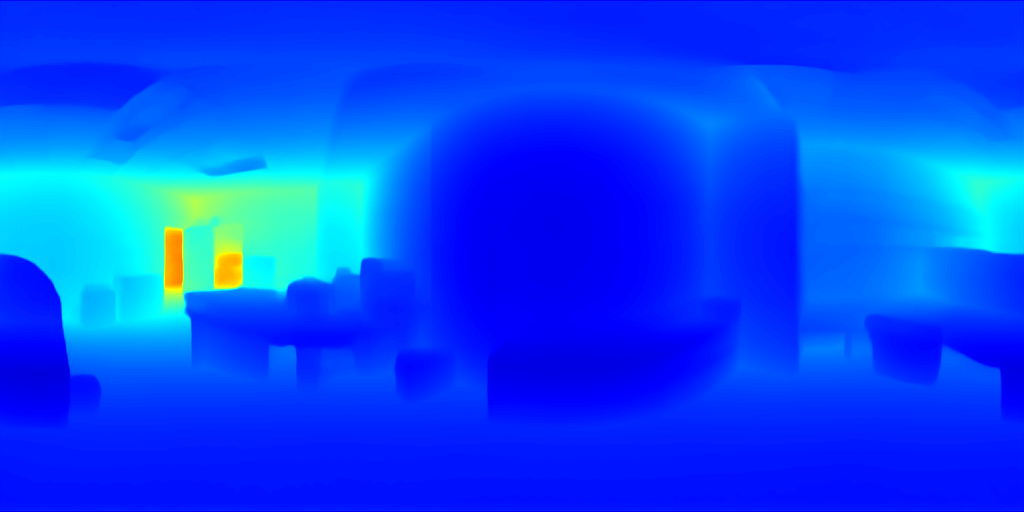}
	\end{subfigure}
	\begin{subfigure}{0.18\linewidth}
		\includegraphics[width=.98\linewidth]{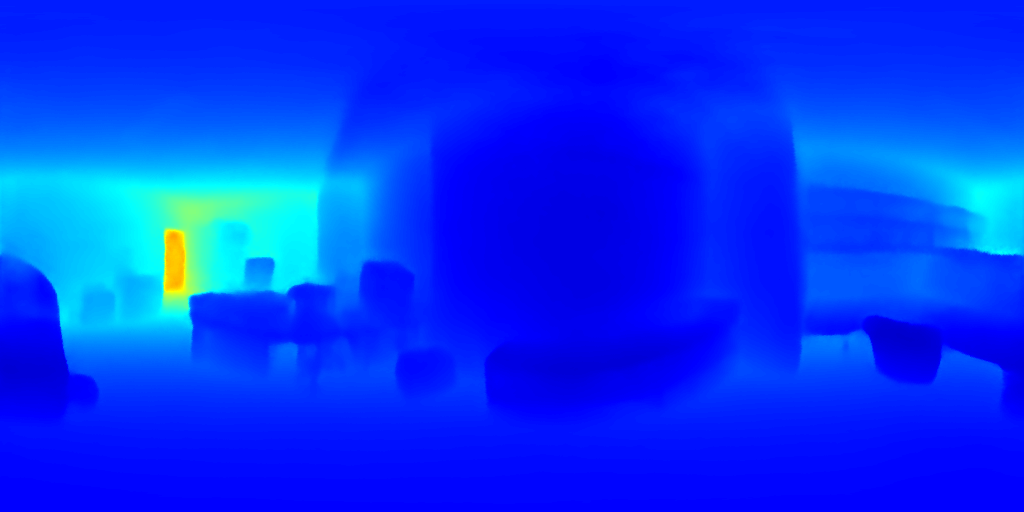}
	\end{subfigure}
	
	\vspace{1pt}
	
	\begin{subfigure}{0.18\linewidth}
		\includegraphics[width=.98\linewidth]{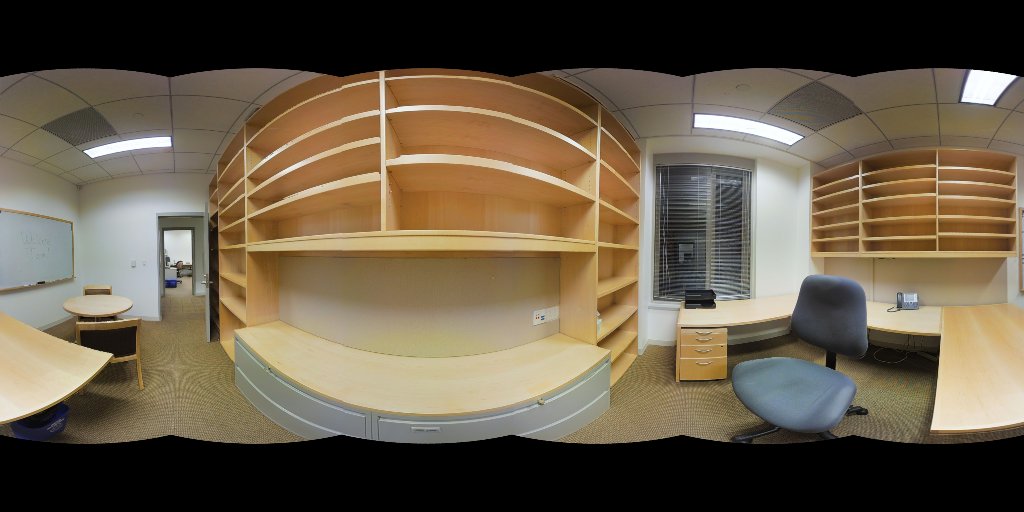}
	\end{subfigure}
	\begin{subfigure}{0.18\linewidth}
		\includegraphics[width=.98\linewidth]{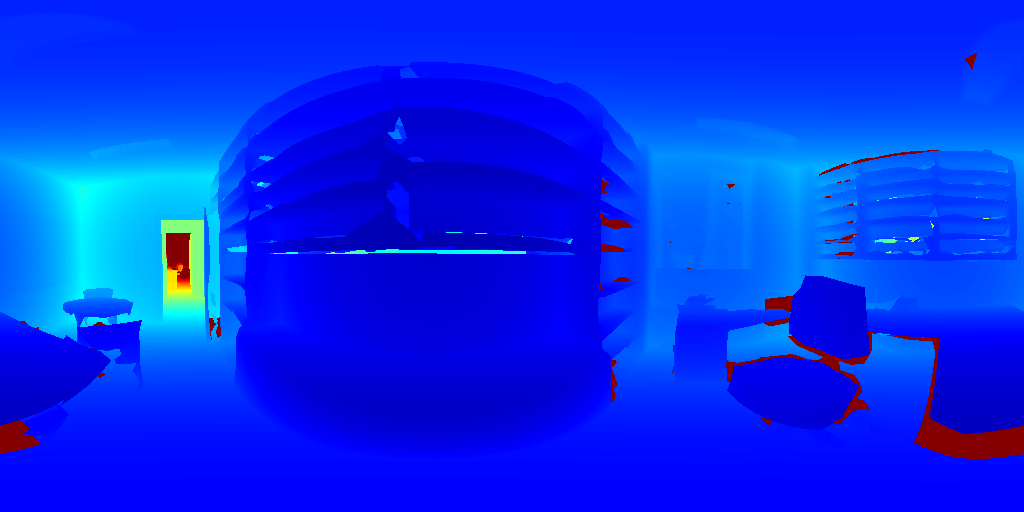}
	\end{subfigure}
	\begin{subfigure}{0.18\linewidth}
		\includegraphics[width=.98\linewidth]{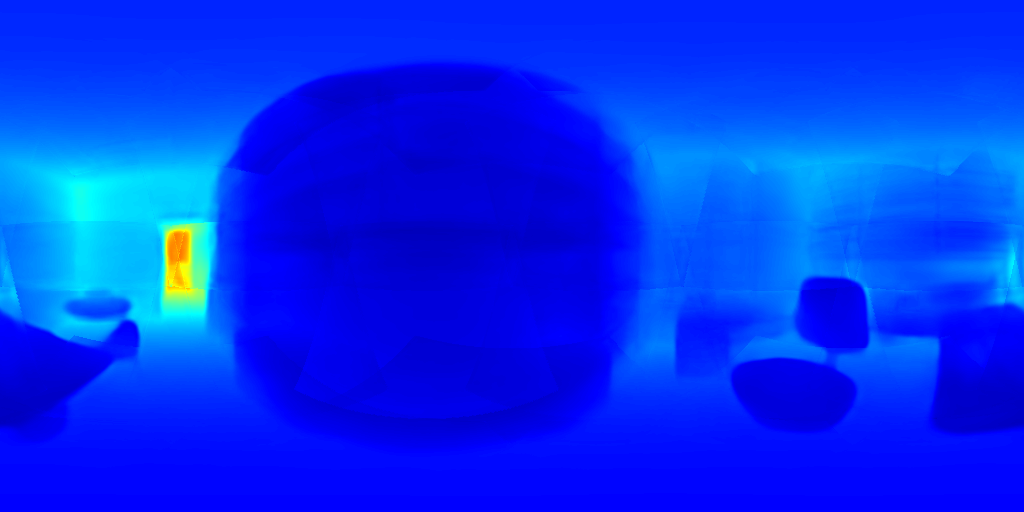}
	\end{subfigure}
	\begin{subfigure}{0.18\linewidth}
		\includegraphics[width=.98\linewidth]{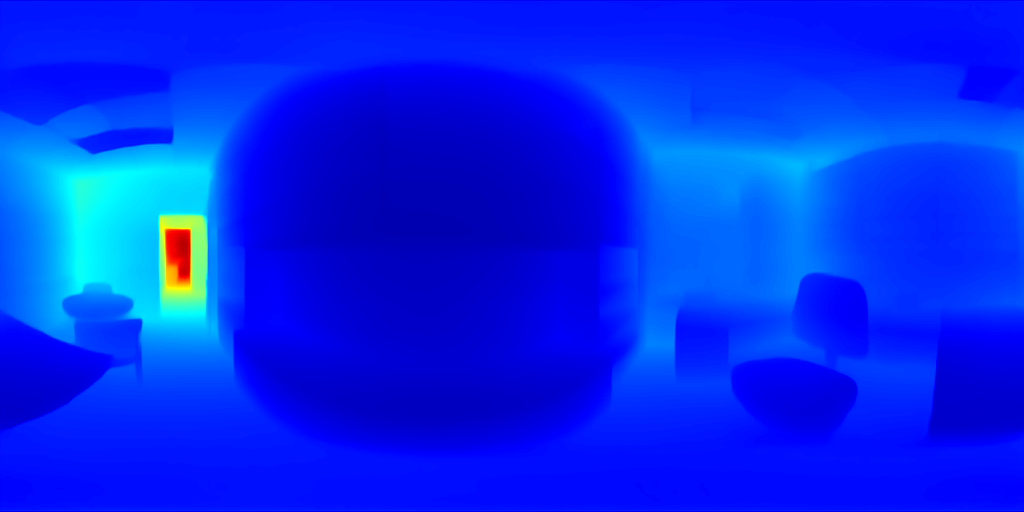}
	\end{subfigure}
	\begin{subfigure}{0.18\linewidth}
		\includegraphics[width=.98\linewidth]{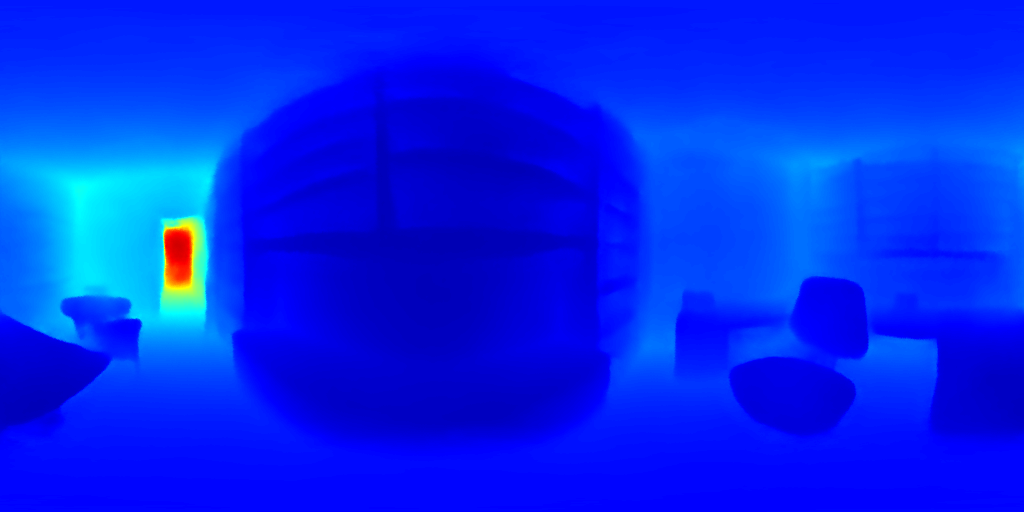}
	\end{subfigure}
	
	\vspace{1pt}
	
	\begin{subfigure}{0.18\linewidth}
		\includegraphics[width=.98\linewidth]{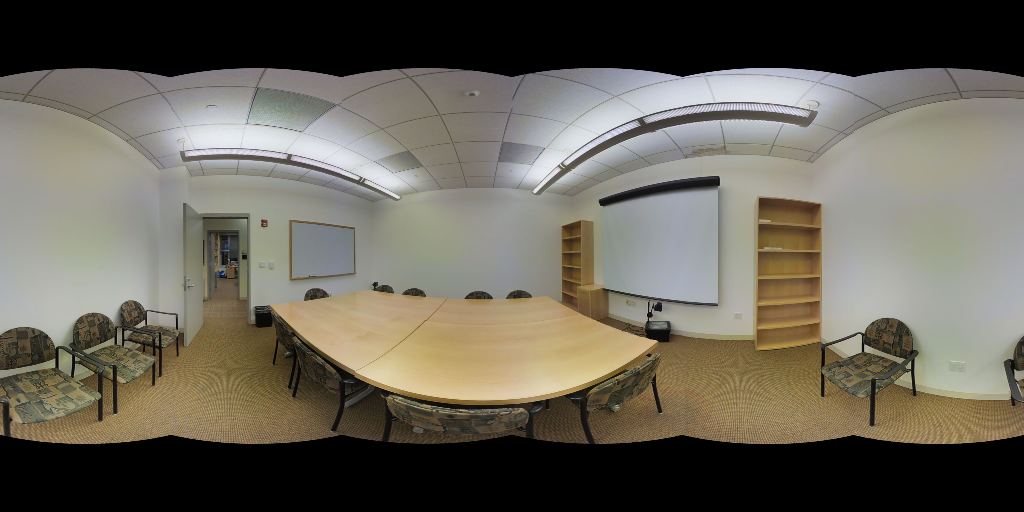}
		\caption{RGB}
	\end{subfigure}
	\begin{subfigure}{0.18\linewidth}
		\includegraphics[width=.98\linewidth]{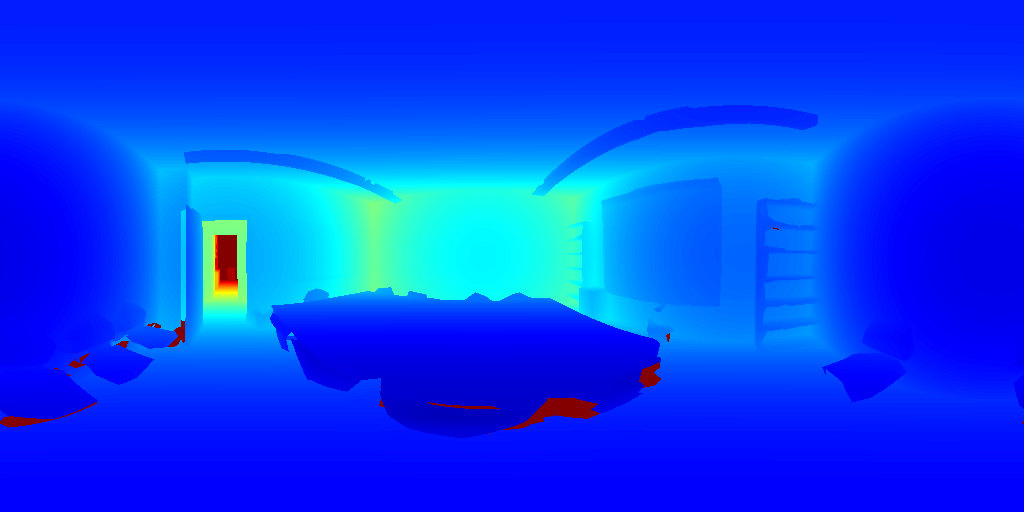}
		\caption{GT}
	\end{subfigure}
	\begin{subfigure}{0.18\linewidth}
		\includegraphics[width=.98\linewidth]{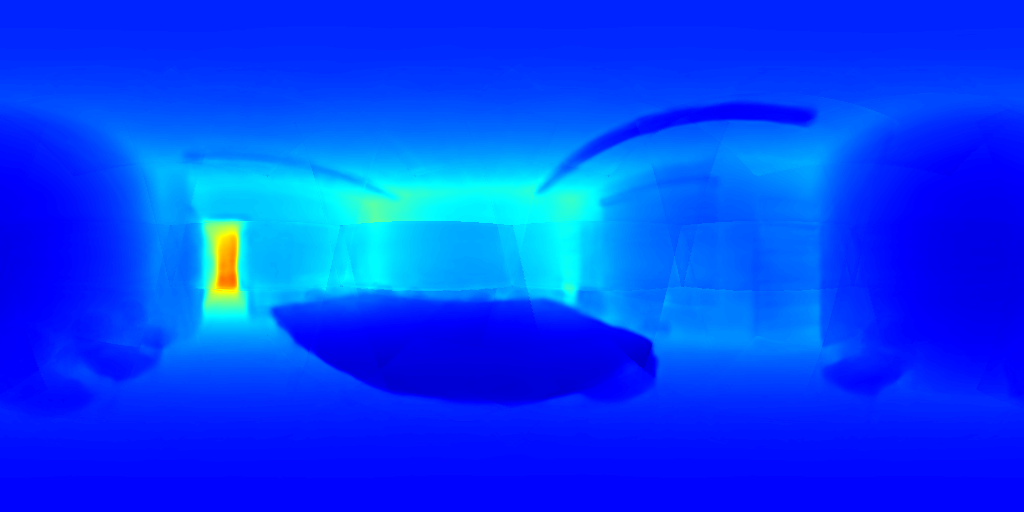}
		\caption{OmniFusion \cite{li2022omnifusion}}
	\end{subfigure}
	\begin{subfigure}{0.18\linewidth}
		\includegraphics[width=.98\linewidth]{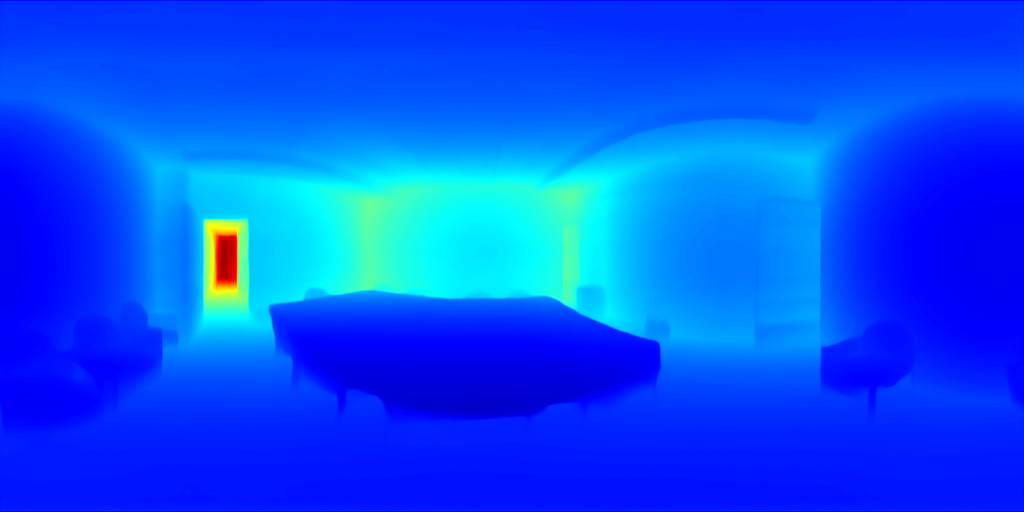}
		\caption{PanoFormer \cite{shen2022panoformer}}
	\end{subfigure}
	\begin{subfigure}{0.18\linewidth}
		\includegraphics[width=.98\linewidth]{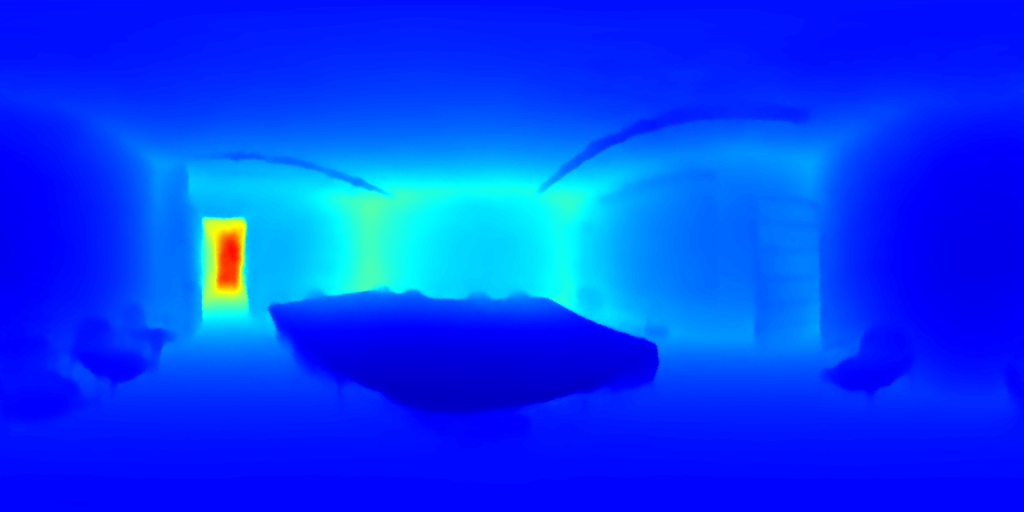}
		\caption{SphereFusion (ours)}
	\end{subfigure}

	\caption{
		\textbf{Depth Maps of Stanford2D3D.} Invalid parts of the depth map are set to red. 
	}
	\label{fig:2d3d_depth}
    \vspace{-1.0em}
\end{figure*}

\begin{figure*}[t]
	\centering
	\captionsetup[subfigure]{labelformat=empty}

	\begin{subfigure}{0.3\linewidth}
		\includegraphics[width=.98\linewidth]{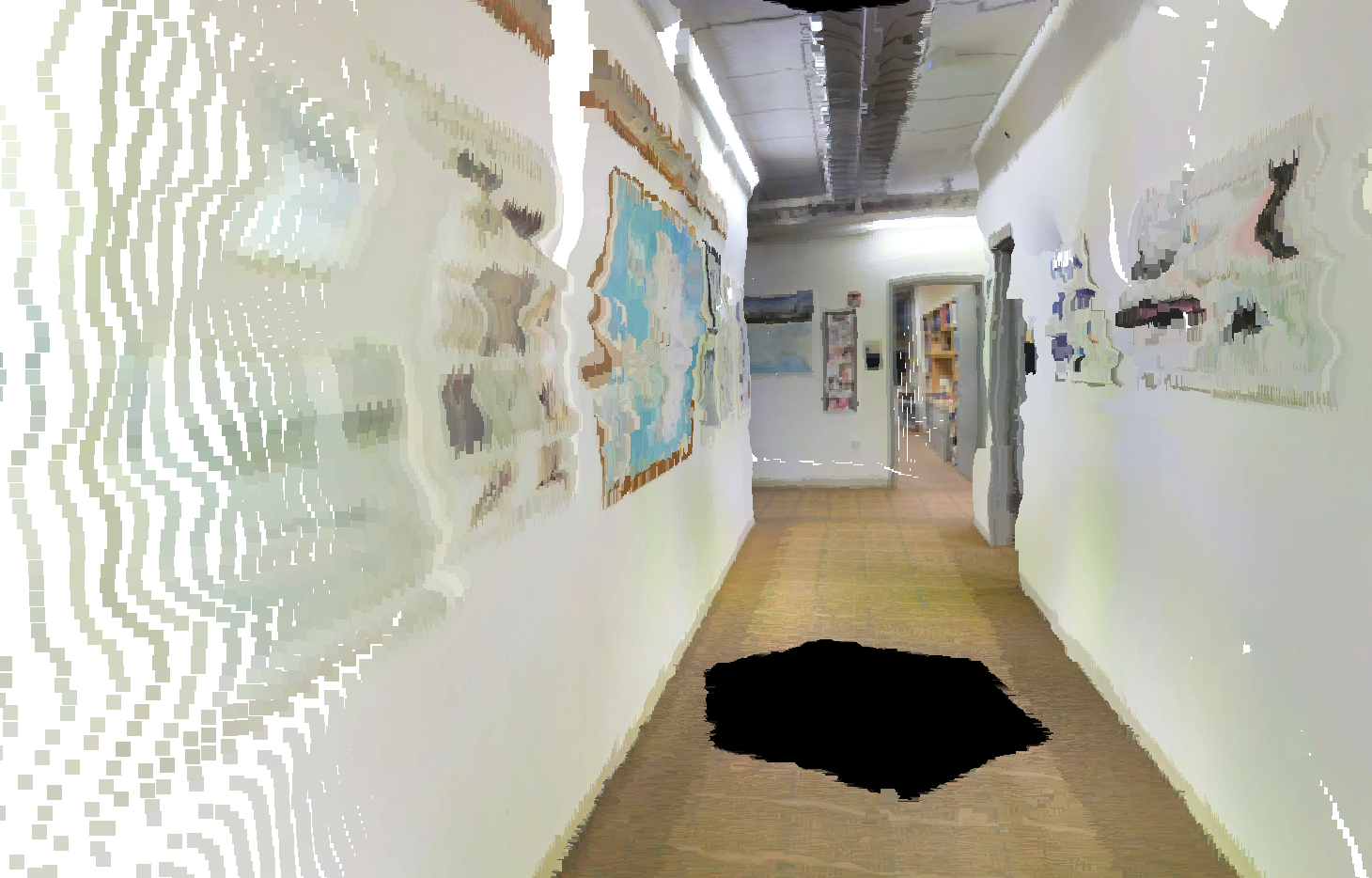}
	\end{subfigure}
	\begin{subfigure}{0.3\linewidth}
		\includegraphics[width=.98\linewidth]{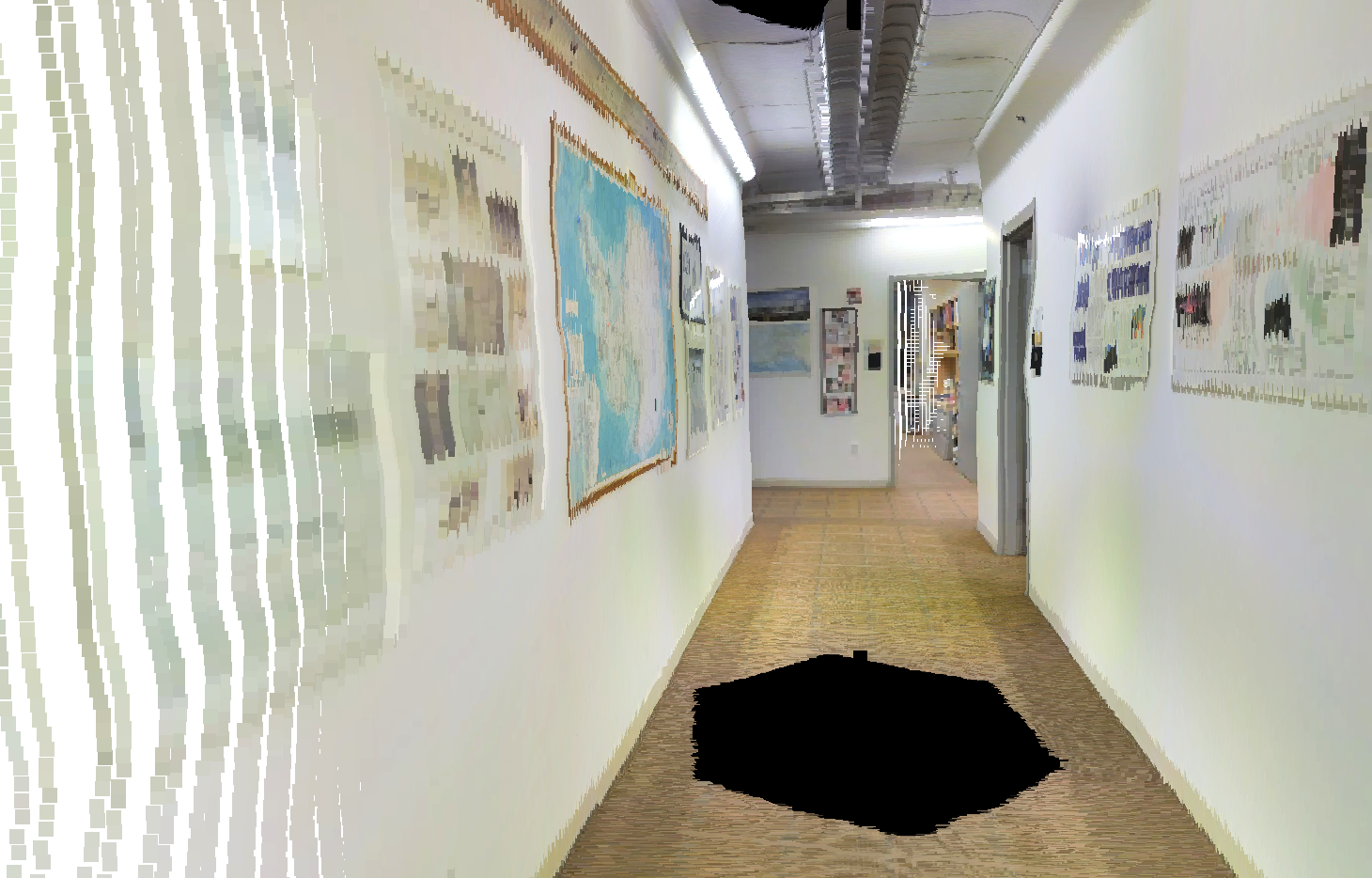}
	\end{subfigure}
	\begin{subfigure}{0.3\linewidth}
		\includegraphics[width=.98\linewidth]{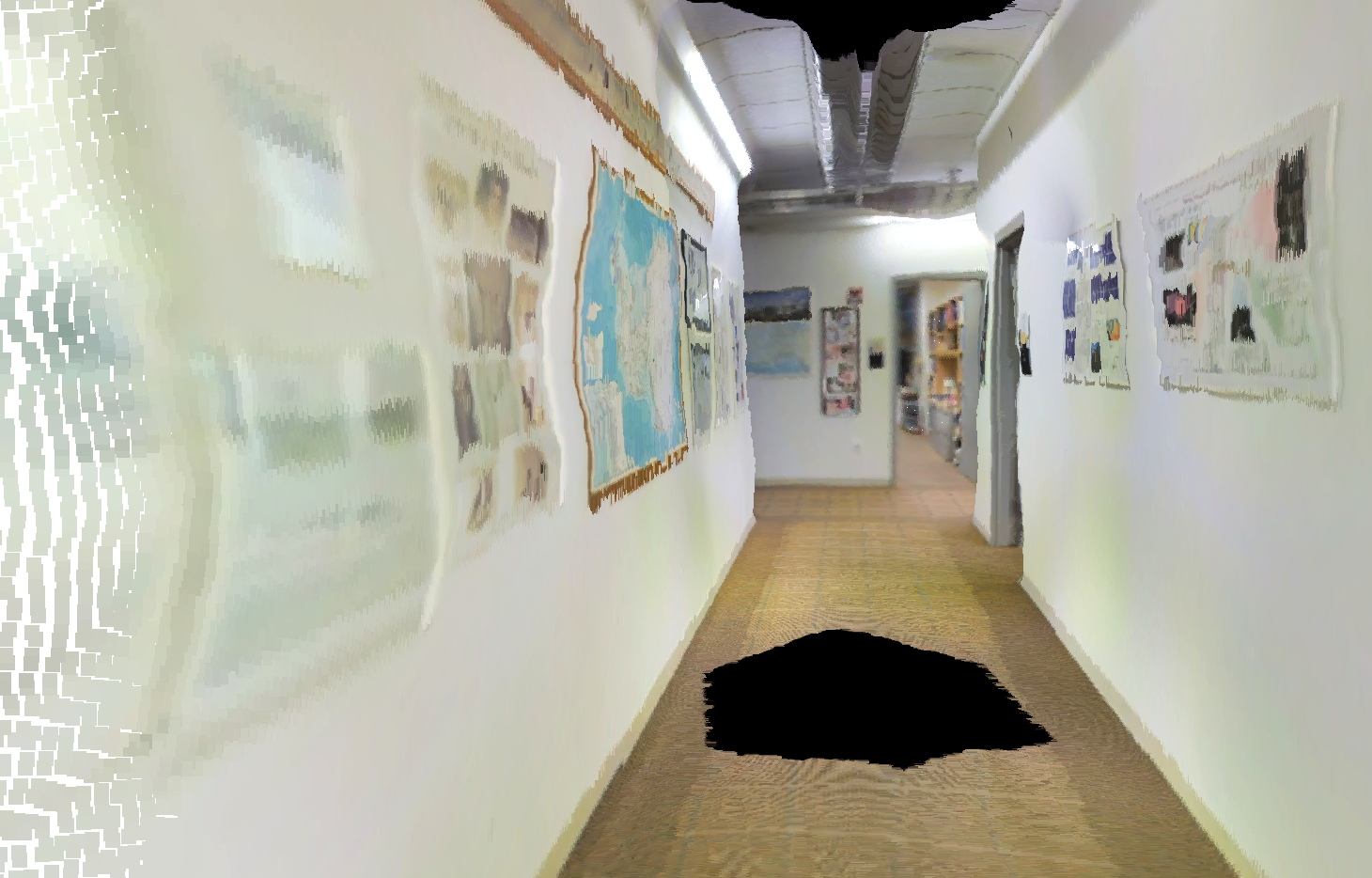}
	\end{subfigure}
	
	\vspace{1pt}

	\begin{subfigure}{0.3\linewidth}
		\includegraphics[width=.98\linewidth]{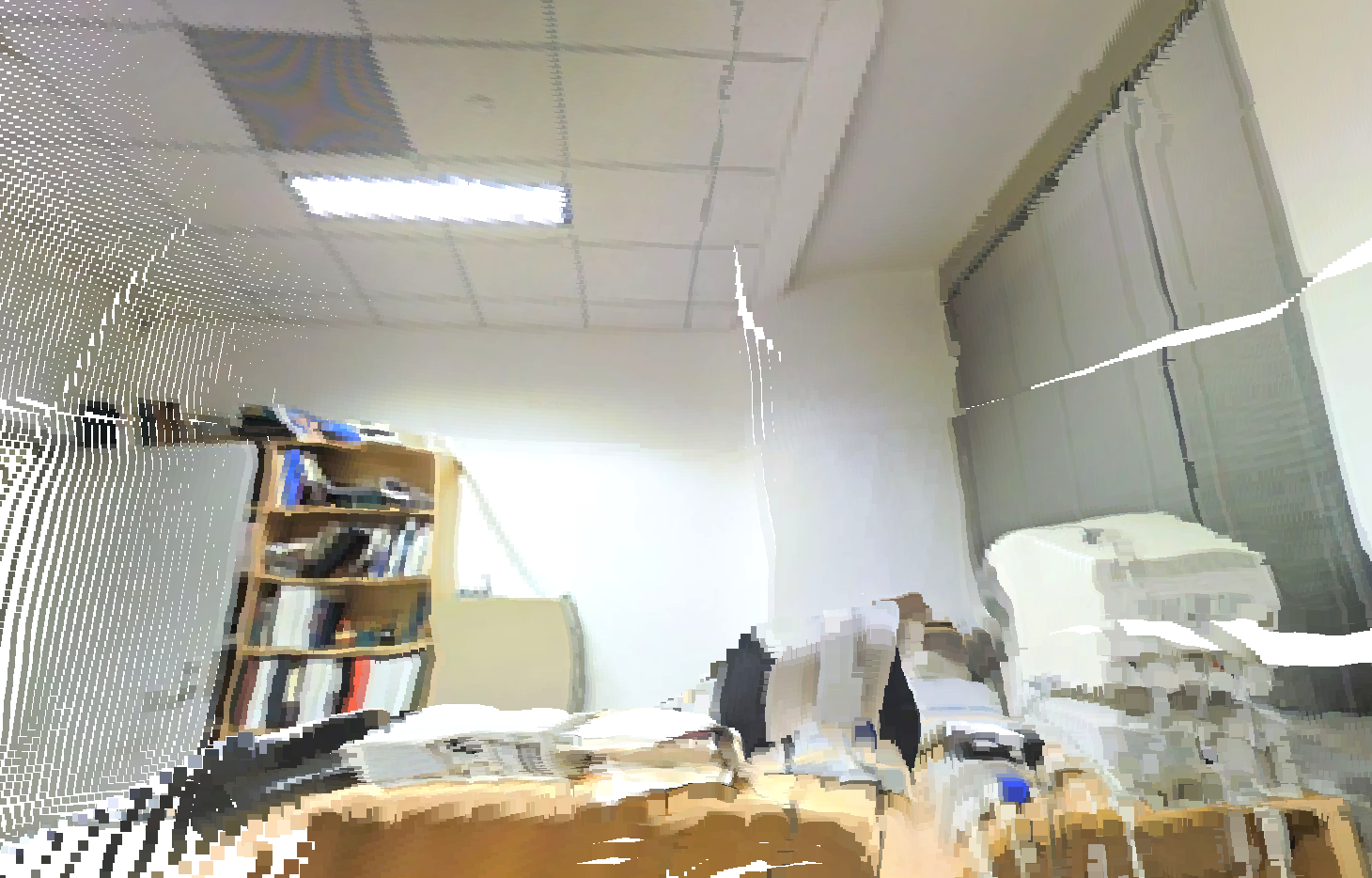}
	\end{subfigure}
	\begin{subfigure}{0.3\linewidth}
		\includegraphics[width=.98\linewidth]{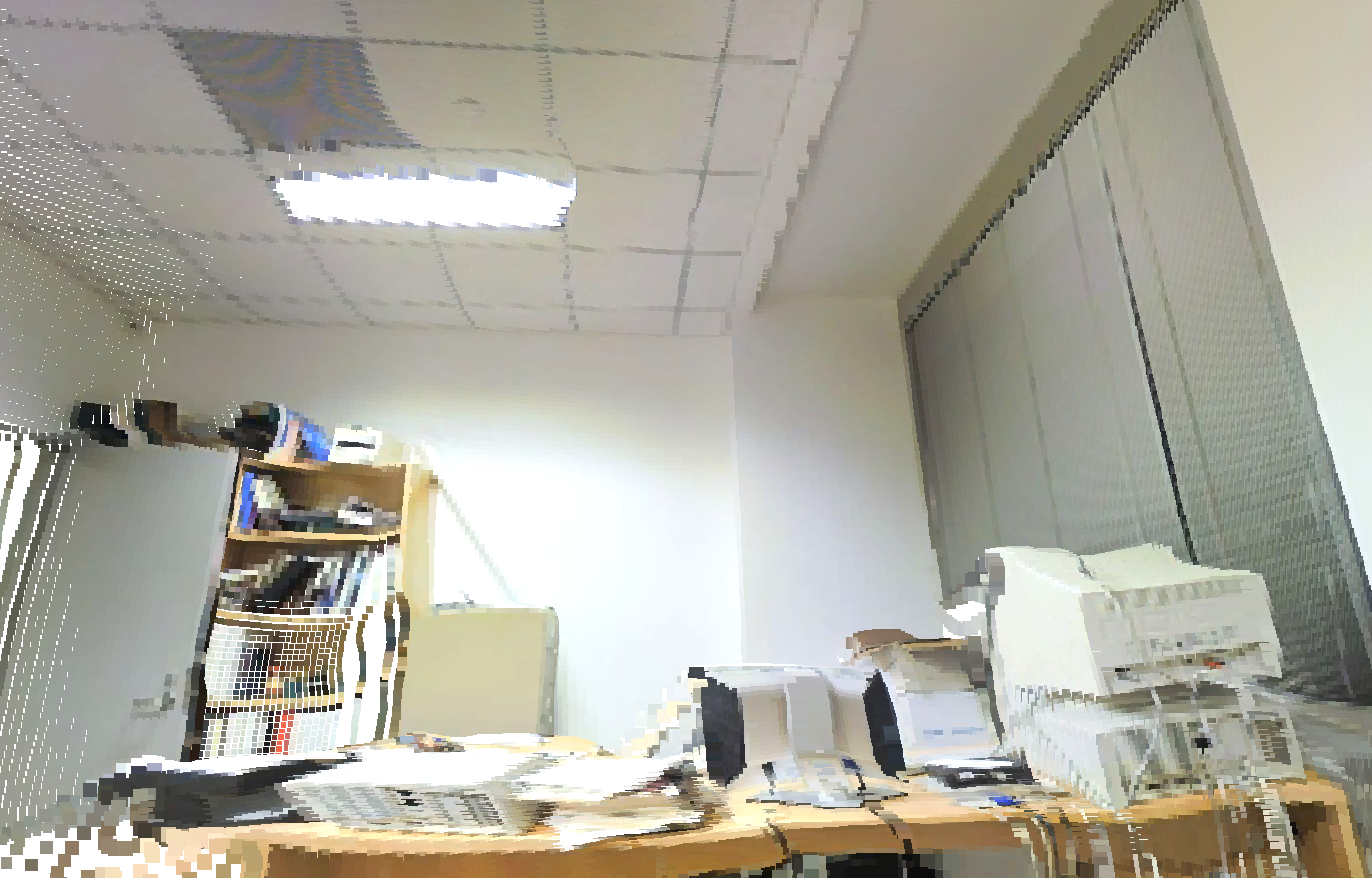}
	\end{subfigure}
	\begin{subfigure}{0.3\linewidth}
		\includegraphics[width=.98\linewidth]{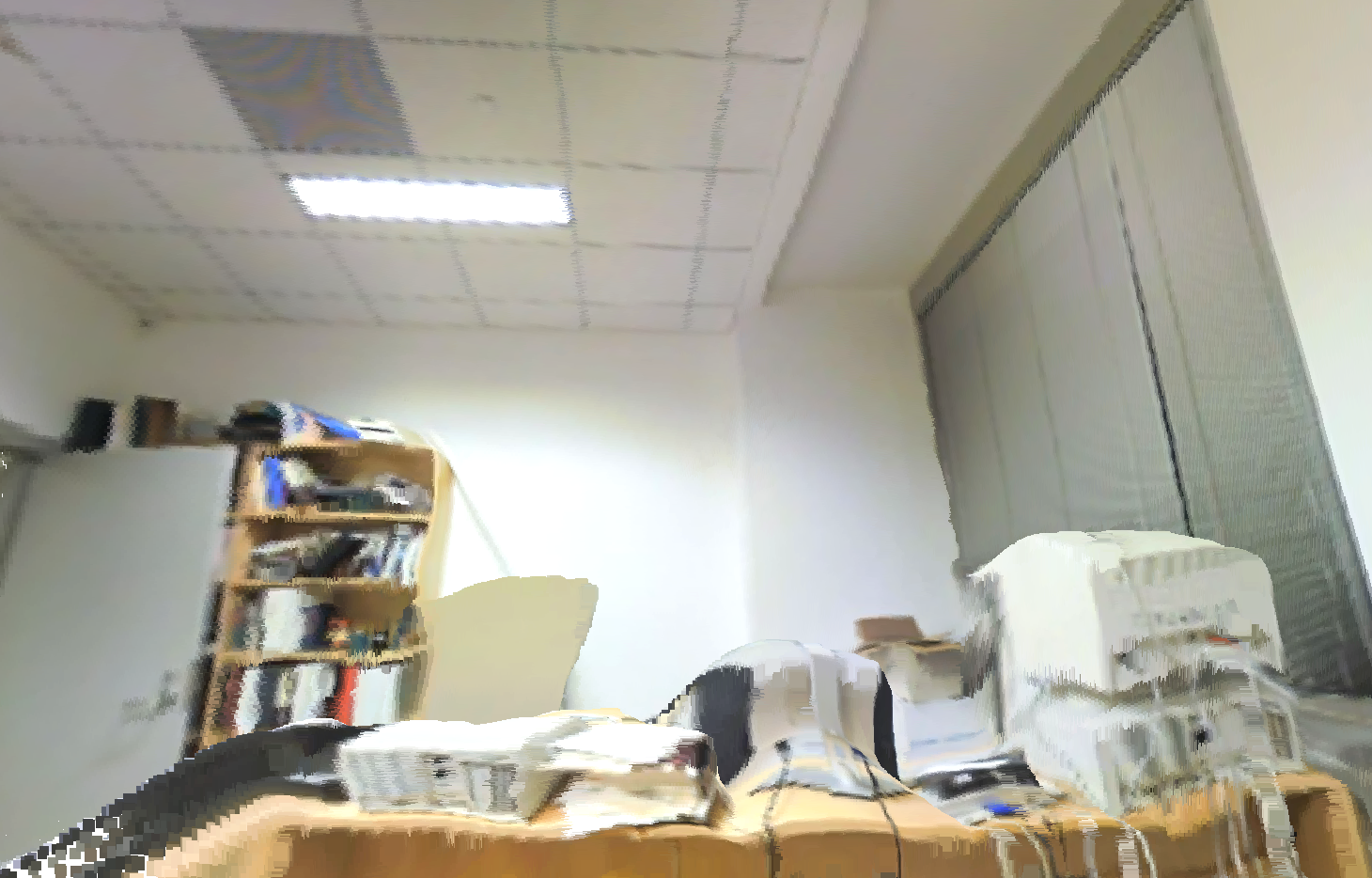}
	\end{subfigure}
	
	\vspace{1pt}
	
	\begin{subfigure}{0.3\linewidth}
		\includegraphics[width=.98\linewidth]{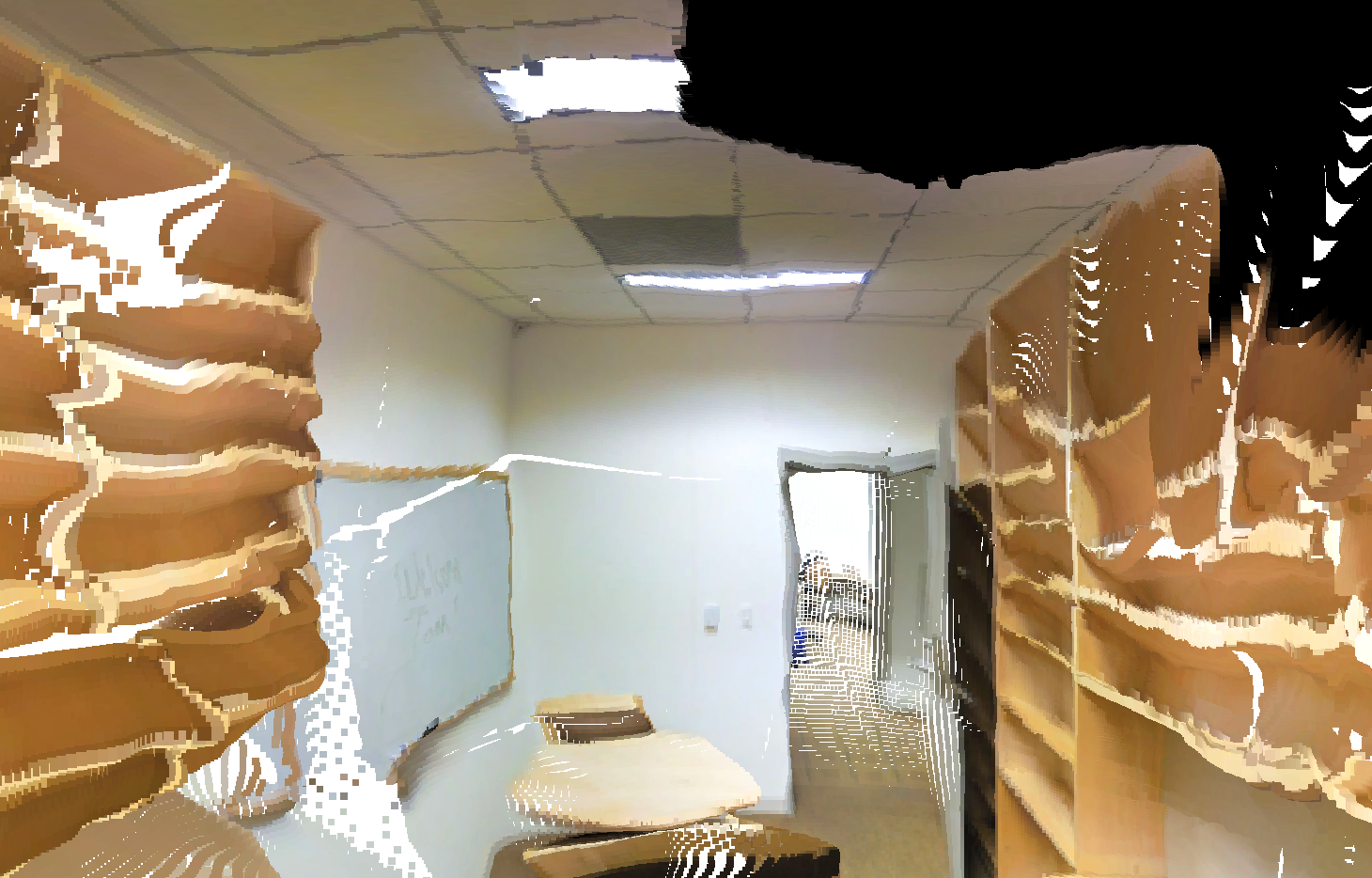}
	\end{subfigure}
	\begin{subfigure}{0.3\linewidth}
		\includegraphics[width=.98\linewidth]{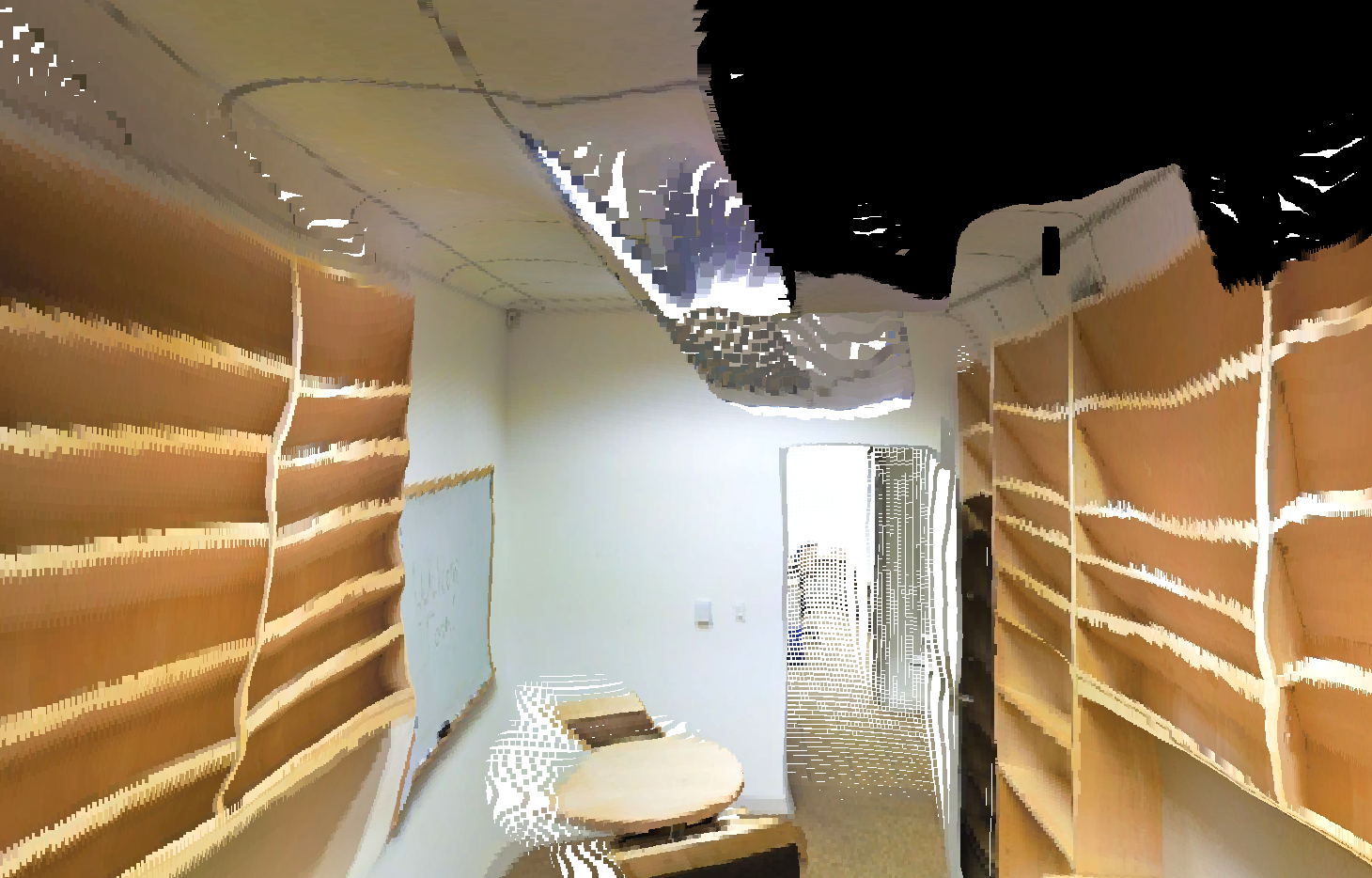}
	\end{subfigure}
	\begin{subfigure}{0.3\linewidth}
		\includegraphics[width=.98\linewidth]{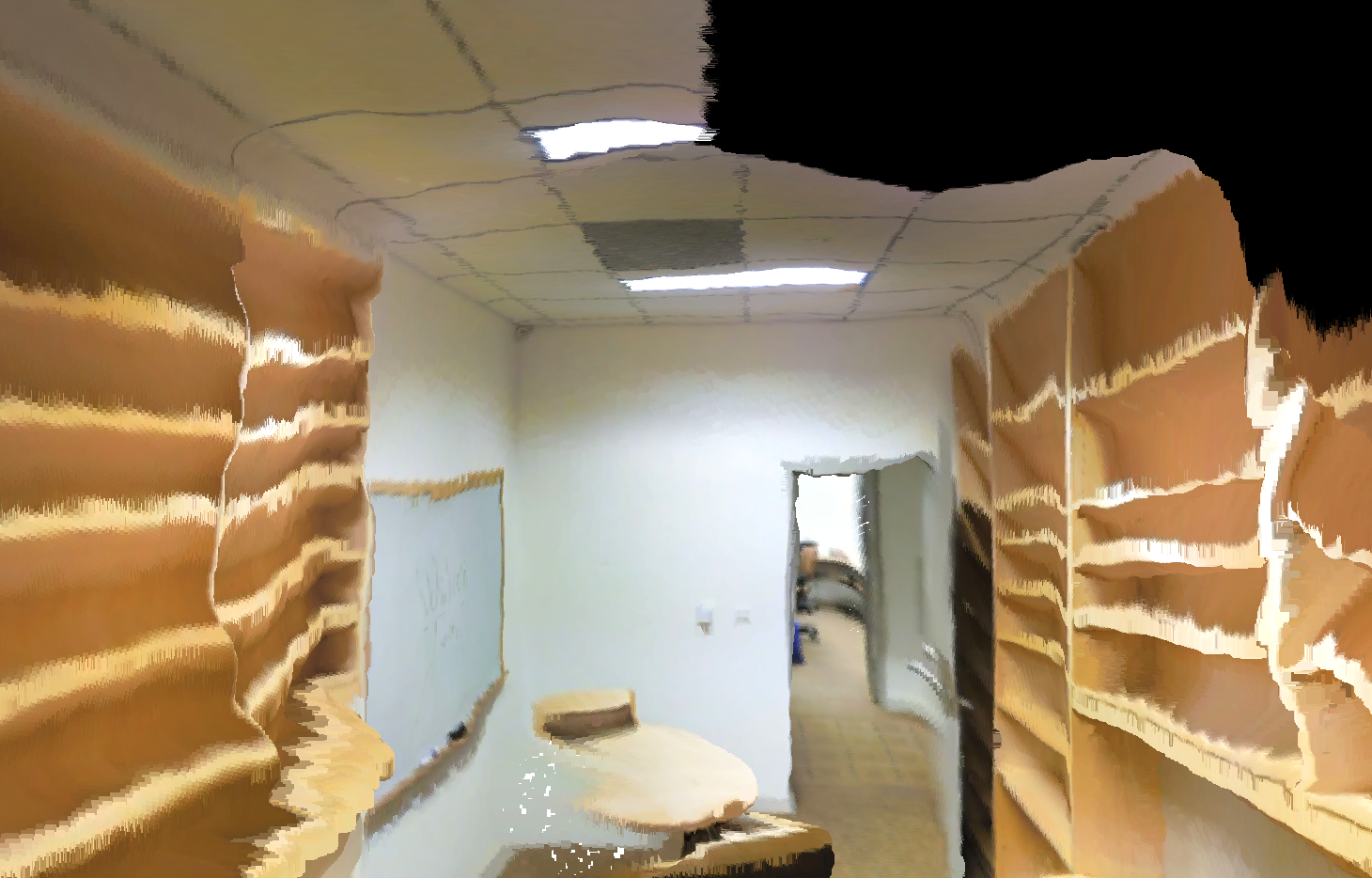}
	\end{subfigure}

	\vspace{1pt}
	
	\begin{subfigure}{0.3\linewidth}
		\includegraphics[width=.98\linewidth]{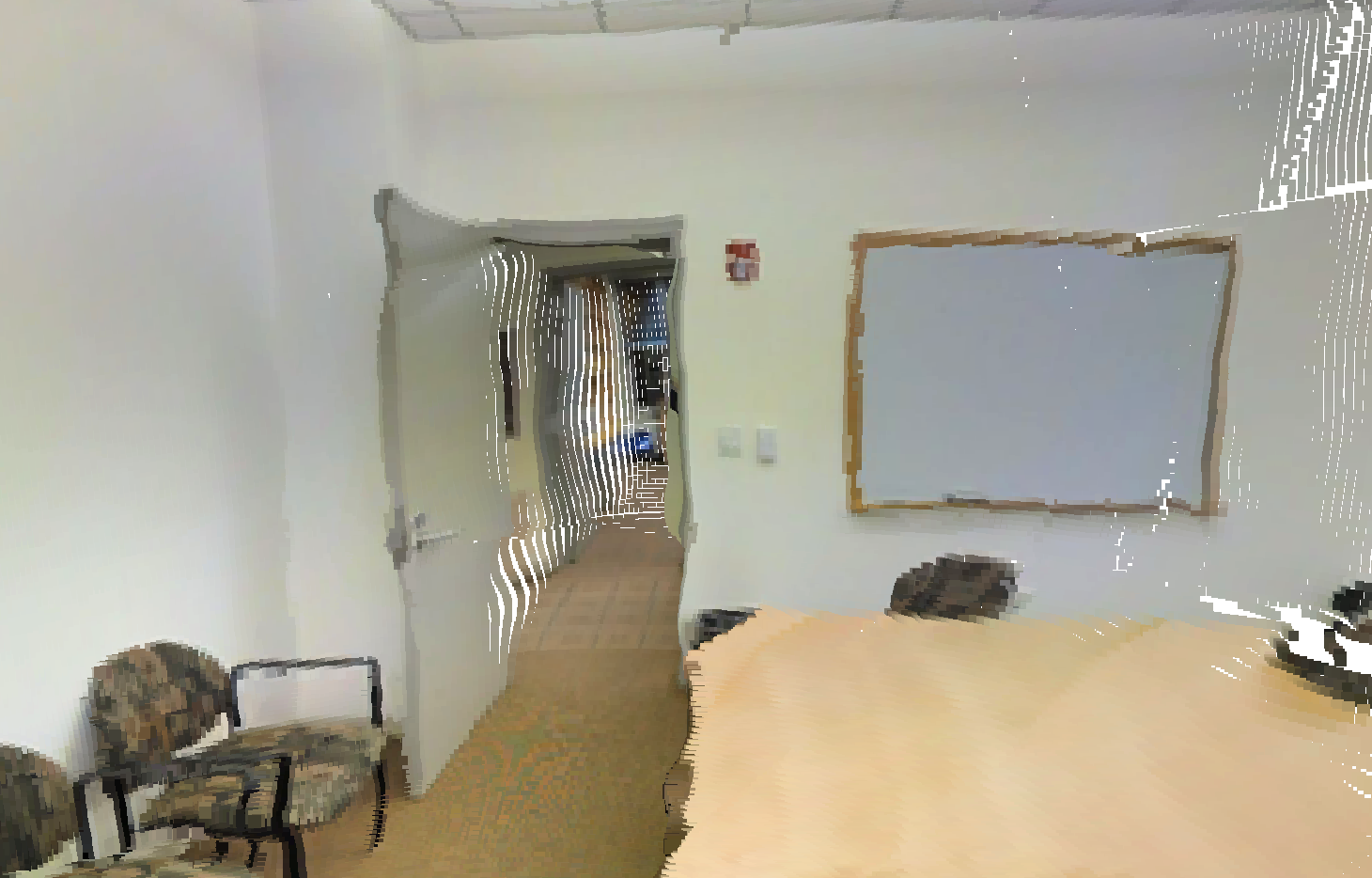}
		\caption{OmniFusion \cite{li2022omnifusion}}
	\end{subfigure}
	\begin{subfigure}{0.3\linewidth}
		\includegraphics[width=.98\linewidth]{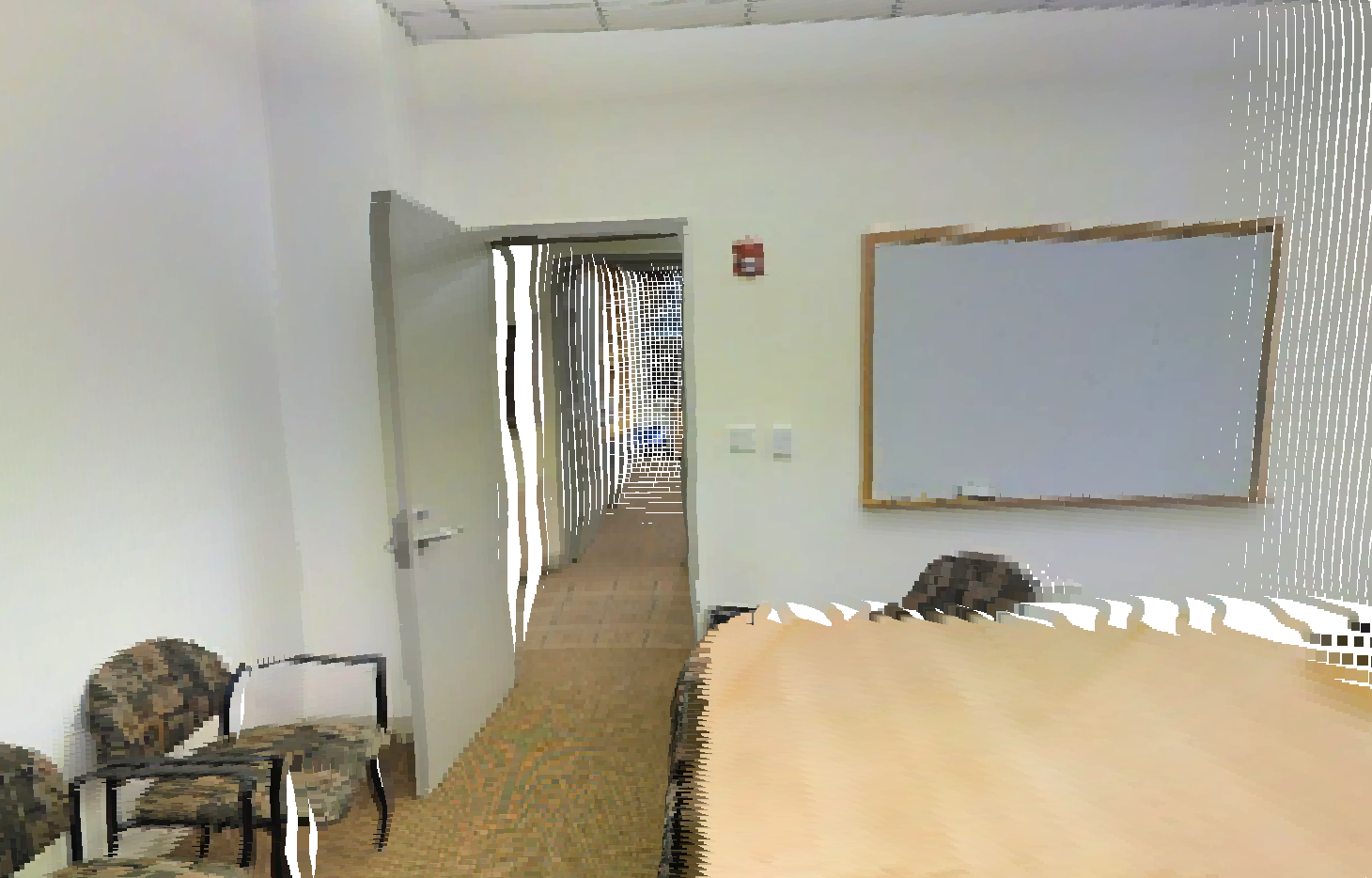}
		\caption{PanoFormer \cite{shen2022panoformer}}
	\end{subfigure}
	\begin{subfigure}{0.3\linewidth}
		\includegraphics[width=.98\linewidth]{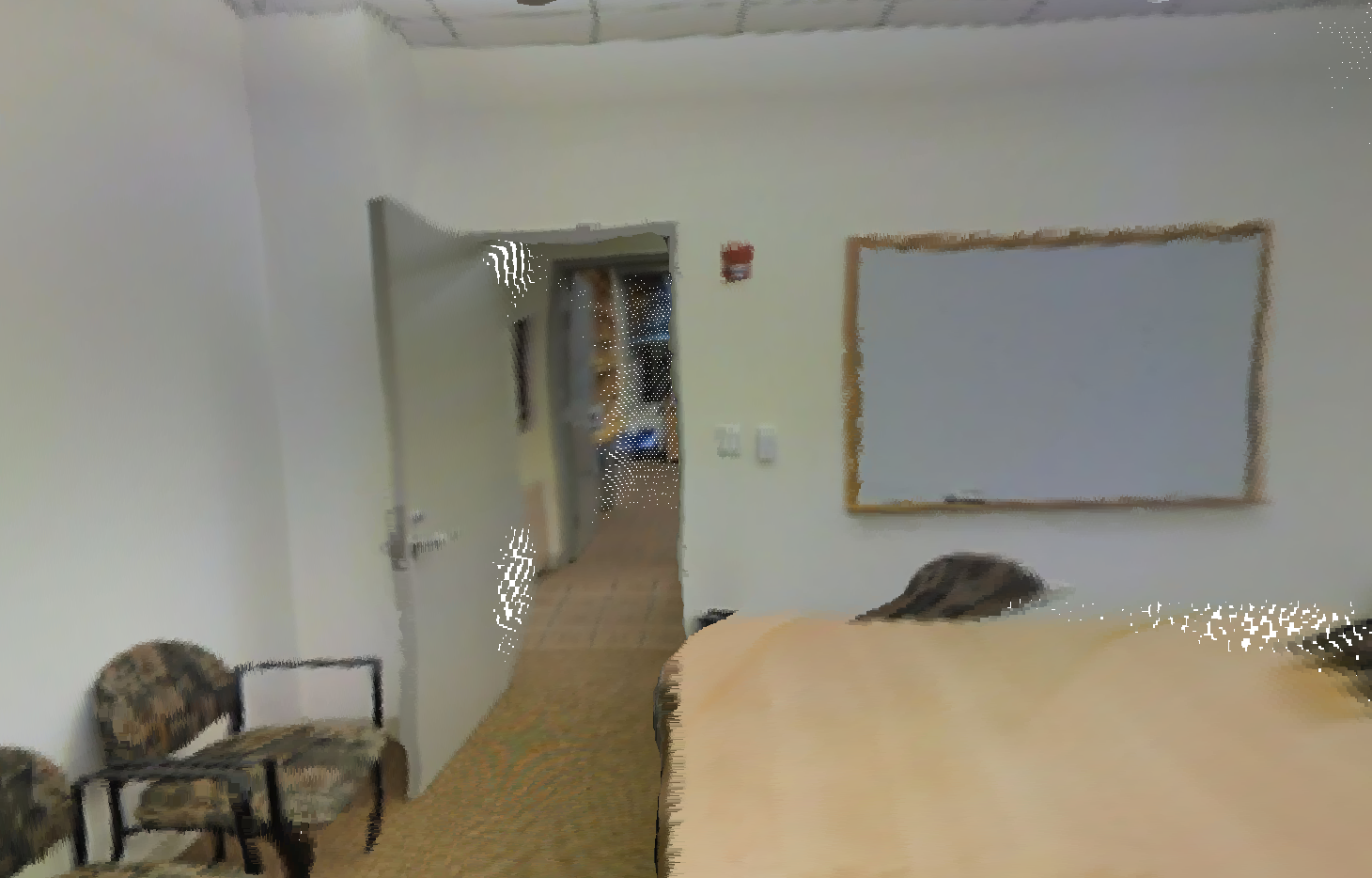}
		\caption{SphereFusion (ours)}
	\end{subfigure}

	\caption{
		\textbf{Depth Maps of Stanford2D3D.} Our method does not suffer from discontinuity.
	}
	\label{fig:2d3d_cloud}
    \vspace{-1.0em}
\end{figure*}

In this section, we add more visualization results on three datasets and compare SphereFusion (ours) with state-of-the-art methods. 
On 360D \cite{zioulis2018omnidepth}, we compare our method with SphereDepth \cite{yan2022spheredepth} and UniFuse \cite{jiang2021unifuse}. Figure \ref{fig:3d60_depth} and Figure \ref{fig:3d60_cloud} show depth maps and point clouds generated by different methods. Our method uses texture features extracted by the 2D encoder to enhance the mesh encoder in the spherical domain and combines the strengths of two encoders. Results show that our method captures more features from the scene and can reconstruct more details in the scene, such as doors, tables, and walls. 
On Matterport3D \cite{chang2017matterport3d} and Stanford2D3D \cite{armeni2017joint}, we compare our method with OmniFusion \cite{yan2022spheredepth} and PanoFormer \cite{jiang2021unifuse}. Figure \ref{fig:mat3d_depth} and Figure \ref{fig:mat3d_cloud} show depth maps generated by different methods on Matterport3D and corresponding point clouds. Figure \ref{fig:2d3d_depth} and Figure \ref{fig:2d3d_cloud} show results generated by different methods on Stanford2D3D. Compared with tangent patches, our method directly estimates the panorama depth in the spherical domain and does not need any special mechanism to fuse these patches. The results of OmniFusion have obvious patch gaps. Meanwhile, the results of PanoFormer are smoother but also lose some details. Our method reconstructs more details but suffers from imperfect ground truth \cite{jiang2021unifuse}. Depth maps and point clouds show that our method achieves competitive results with state-of-the-art methods with a lighter network and higher efficiency.

\FloatBarrier

\FloatBarrier

\end{document}